\documentclass{article}

\usepackage{PRIMEarxiv}

\usepackage[utf8]{inputenc} 
\usepackage[T1]{fontenc}    
\usepackage{url}            
\usepackage{booktabs}       
\usepackage{amsfonts}       
\usepackage{nicefrac}       
\usepackage{microtype}      
\usepackage{lipsum}
\usepackage{csquotes}
\usepackage{fancyhdr}       
\usepackage{longtable}
\usepackage{graphicx}       
\usepackage{multirow}
\usepackage{listings}
\usepackage{microtype}  
\usepackage[most]{tcolorbox}
\usepackage[T1]{fontenc}
\usepackage{caption}
\usepackage{subcaption}
\usepackage{tgpagella}
\usepackage{xspace}
\usepackage[round,semicolon]{natbib}

\definecolor{galpurple}{rgb}{0.3,0.3,0.5}
\definecolor{galpurple2}{rgb}{0.4,0.4,0.77}
\definecolor{galwhite}{rgb}{0.99,0.99,0.99}

\usepackage[colorlinks,urlcolor=galpurple2,citecolor=galpurple2,linkcolor=galpurple]{hyperref}

\usepackage[labelfont=bf]{caption}
\usepackage[version=4]{mhchem}
\graphicspath{{media/}}     

\title{Galactica: A Large Language Model for Science}

\author{
  Ross Taylor
  \And
  Marcin Kardas
  \And
  Guillem Cucurull
  \AND
  Thomas Scialom
  \And
  Anthony Hartshorn
  \And
  Elvis Saravia
  \AND
  Andrew Poulton
  \And
  Viktor Kerkez
  \And
  Robert Stojnic
  \AND \\
  Meta AI
}

\begin{document}
\maketitle

\begin{abstract}
Information overload is a major obstacle to scientific progress. The explosive growth in scientific literature and data has made it ever harder to discover useful insights in a large mass of information. Today scientific knowledge is accessed through search engines, but they are unable to organize scientific knowledge alone. In this paper we introduce Galactica: a large language model that can store, combine and reason about scientific knowledge. We train on a large scientific corpus of papers, reference material, knowledge bases and many other sources. We outperform existing models on a range of scientific tasks. On technical knowledge probes such as LaTeX equations, Galactica outperforms the latest GPT-3 by 68.2\% versus 49.0\%. Galactica also performs well on reasoning, outperforming Chinchilla on mathematical MMLU by 41.3\% to 35.7\%, and PaLM 540B on MATH with a score of 20.4\% versus 8.8\%. It also sets a new state-of-the-art on downstream tasks such as PubMedQA and MedMCQA dev of 77.6\% and 52.9\%. And despite not being trained on a general corpus, Galactica outperforms BLOOM and OPT-175B on BIG-bench. We believe these results demonstrate the potential for language models as a new interface for science. We open source the model for the benefit of the scientific community\footnote{\url{galactica.org}}.
\end{abstract}

\section{Introduction}


The original promise of computing was to solve information overload in science. In his 1945 essay "As We May Think", Vannevar Bush observed how "publication has been extended far beyond our present ability to make real use of the record"~\citep{bush1945}. He proposed computers as a solution to manage the growing mountain of information. Licklider expanded on this with the vision of a symbiotic relationship between humans and machines. Computers would take care of routine tasks such as storage and retrieval, "preparing the way for insights and decisions in scientific thinking"~\citep{licklider1960}. 

Computing has indeed revolutionized how research is conducted, but information overload remains an overwhelming problem~\citep{GrowthRateScience}. In May 2022, an average of 516 papers per day were submitted to arXiv~\citep{arxivpapers}. Beyond papers, scientific data is also growing much more quickly than our ability to process it~\citep{BigChallengesBigData}. As of August 2022, the NCBI GenBank contained $1.49 \times 10^{12}$ nucleotide bases~\citep{genbank}. Given the volume of information, it is impossible for a single person to read all the papers in a given field; and it is likewise challenging to organize data on the underlying scientific phenomena.

Search engines are the current interface for accessing scientific knowledge following the Licklider paradigm. But they do not organize knowledge directly, and instead point to secondary layers such as Wikipedia, UniProt and PubChem Compound which organize literature and data. These resources require costly human contributions, for example writing a review of literature, an encyclopedia article or annotating a protein. Given this bottleneck, researchers continue to feel overwhelmed even with powerful search tools to hand.

In this paper, we argue for a better way through large language models. Unlike search engines, language models can potentially store, combine and reason about scientific knowledge. For example, a model trained on the literature could potentially find hidden connections between different research, find hidden gems, and bring these insights to the surface. It could synthesize knowledge by generating secondary content automatically: such as literature reviews, encyclopedia articles, lecture notes and more. And lastly, it could organize different modalities: linking papers with code, protein sequences with compounds, theories with LaTeX, and more. Our ultimate vision is a single neural network for powering scientific tasks. We believe this is will be the next interface for how humans access scientific knowledge, and we get started in this paper.

\subsection{Our Contribution}

We introduce a new large language model called Galactica (GAL) for automatically organizing science. Galactica is trained on a large and curated corpus of humanity's scientific knowledge. This includes over 48 million papers, textbooks and lecture notes, millions of compounds and proteins, scientific websites, encyclopedias and more. Unlike existing language models, which rely on an uncurated crawl-based paradigm, our corpus is high-quality and highly curated. We are able to train on it for multiple epochs without overfitting, where upstream and downstream performance improves with use of repeated tokens.

Dataset design is critical to our approach, which includes curating a high-quality dataset and engineering an interface to interact with the body of knowledge. All data is processed in a common markdown format to blend knowledge between sources. We also include task-specific datasets in pre-training to facilitate composition of this knowledge into new task contexts. For the interface, we use task-specific tokens to support different types of knowledge. We process citations with a special token, that allows a researcher to predict a citation given any input context. We wrap step-by-step reasoning in a special token, that mimicks an internal working memory. And lastly, we wrap modalities such as SMILES and protein sequences in special tokens, which allows a researcher to interface with them using natural language. With this interface and the body of scientific knowledge in the model, we achieve state-of-the-art results across many scientific tasks.

On reasoning tasks, Galactica beats existing language models on benchmarks such as MMLU and MATH~\citep{MMMLU, MATH}. With our reasoning token approach, we outperform Chinchilla on mathematical MMLU with an average score of 41.3\% versus 35.7\%~\citep{Chinchilla}. Our 120B model achieves a score of 20.4\% versus PaLM 540B's 8.8\% on MATH~\citep{PaLM, Minerva}. The 30B model also beats PaLM 540B on this task with 18 times less parameters. We believe this adds another reasoning method to the deep learning toolkit, alongside the existing chain-of-thought approach that has been well explored recently~\citep{ChainOfThought, CoTBigBENCH}.

We also find Galactica performs strongly in knowledge-intensive scientific tasks. We conduct detailed knowledge probes of Galactica's knowledge of equations, chemical reactions and other scientific knowledge. Galactica significantly exceeds the performance of general language models such as the latest GPT-3 in these tasks; on LaTeX equations, it achieves a score of 68.2\% versus the latest GPT-3's 49.0\%~\citep{GPT3}. Galactica also performs well in downstream scientific tasks, and we set a new state-of-the-art on several downstream tasks such as PubMedQA (77.6\%) and MedMCQA dev (52.9\%)~\citep{PubMedQA, MedMCQA}. 

We also demonstrate new capabilities with Galactica's interface. First, the capability of predicting citations improves smoothly with scale, and we also find the model  becomes better at modelling the underlying distribution of citations: the empirical distribution function approaches the reference distribution with scale. Importantly, we find this approach outperforms tuned sparse and dense retrieval approaches for citation prediction. This, along other results, demonstrates the potential for language models to replace the Licklider paradigm, document storage and retrieval, with their context-associative power in weight memory.

In addition, Galactica can perform multi-modal tasks involving SMILES chemical formulas and protein sequences. We formulate drug discovery tasks as text prompts and show performance scales in a weakly supervised setup. We also demonstrate Galactica learns tasks such as IUPAC name prediction in a self-supervised way, and does so by attending to interpretable properties such as functional groups. Lastly, Galactica can annotate protein sequences with natural language, including predicting functional keywords.

Galactica was used to help write this paper, including recommending missing citations, topics to discuss in the introduction and related work, recommending further work, and helping write the abstract and conclusion.

\section{Related Work}

\paragraph{Large Language Models (LLMs)}LLMs have achieved breakthrough performance on NLP tasks in recent years. Models are trained with self-supervision on large, general corpuses and they perform well on hundreds of tasks~\citep{GPT3, Gopher, Chinchilla, GPTNeox, OPT, PaLM}. This includes scientific knowledge tasks such as MMLU~\citep{MMMLU}. They have the capability to learn in-context through few-shot learning~\citep{GPT3}. The capability set increases with scale, and recent work has highlighted reasoning capabilities at larger scales with a suitable prompting strategy~\citep{ChainOfThought,PaLM,StepByStep,Minerva}.

One downside of self-supervision has been the move towards uncurated data. Models may mirror misinformation, stereotypes and bias in the corpus~\citep{WomanBabysitter,MeasuringBias,MeasureMitigate,LanguageIsPower,SocietalBiases}. This is undesirable for scientific tasks which value truth. Uncurated data also means more tokens with limited transfer value for the target use-case; wasting compute budget. For example, the PaLM corpus is 50\% social media conversations, which may have limited transfer towards scientific tasks~\citep{PaLM}. The properties of scientific text also differ from general text - e.g. scientific terms and mathematics - meaning a general corpus and tokenizer may be inefficient. We explore whether a normative approach to dataset selection can work with the large model paradigm in this work.

\paragraph{Scientific Language Models}Works such as SciBERT, BioLM and others have shown the benefit of a curated, scientific corpus~\citep{SciBERT, BioLM, PubMedBERT, S2ORCBERT, PubMedBERT, BioMegatron, ScholarBERT}. The datasets and models were typically small in scale and scope,  much less than corpora for general models\footnote{One of the larger corpora S2ORC has \(<20\)bn tokens, whereas corpora for GPT-3 and PaLM have \(\geq 300\)bn tokens. ScholarBERT has a very large corpus at >200bn tokens, but the model is small at 770M capacity.}. Beyond scientific text, Transformers for protein sequences and SMILES have shown potential for learning natural representations~\citep{BioEmerge, honda2019smiles, Chemformer, ProGen2, ESMFold}. However, sequences like SMILES have descriptive limitations for representing chemical structure. We explore in this work whether a large, multi-modal scientific corpus can aid representation learning, where sequences occur alongside footprints and text in a signal-dense context.

\paragraph{Scaling Laws}The idea of "scaling laws" was put forward by \citet{ScalingLaws}, who demonstrated evidence that loss scales as a power-law with model size, dataset size, and the amount of training compute. The focus was on upstream perplexity, and work by \citet{ScalingLawsModelArch} showed that this does not always correlate with downstream performance. \citet{Chinchilla} presented new analysis taking into account the optimal amount of data, and suggested that existing language models were undertrained: "Chinchilla scaling laws". This work did not take into the account of fresh versus repeated tokens. In this work, we show that we can improve upstream and downstream performance by training on repeated tokens.

\paragraph{Language Models as Knowledge Bases}Storing information in weights is more unreliable in the sense models may blend information together, \textit{hallucination}, but it is more "pliable" in the sense it can associate information through the representation space, \textit{association}. Despite hallucination risks, there is evidence large language models can act as implicit knowledge bases with sufficient capacity~\citep{petroni2019language}. They perform well on knowledge-intensive tasks such as general knowledge (TriviaQA) and specialist knowledge (MMLU) without an external retrieval mechanism~\citep{GPT3, MMMLU}. 

The question of how to update network knowledge remains an active research question~\citep{ContinualT0, MBMES}. Likewise, the question of how to improve the reliability of generation is an active question~\citep{PostHocResearch}. Despite these limitations, today's large models will become cheaper with experience~\citep{WrightsLaw}, and so a growing proportion of scientific knowledge will enter weight memory as training and re-training costs fall. In this work we perform probes to investigate Galactica's depth of knowledge, and show that the ability to absorb scientific knowledge improves smoothly with scale.

\paragraph{Retrieval-Augmented Models}Retrieval-augmented models aim to alleviate the shortcomings of weight memory. Examples of such models include RAG, RETRO and Atlas~\citep{RAG, RETRO, izacard2022fewshot}. These models have the advantage of requiring less capacity but the disadvantage of needing supporting retrieval infrastructure. Since knowledge is often fine-grained, e.g. the sequence of a particular protein, or the characteristics of a particular exoplanet, retrieval will likely be needed in future even for larger models. In this work we focus on how far we can go with model weights alone, but we note the strong case for using retrieval augmentation for future research on this topic.

\section{Dataset}

\begin{table}[t!]
\begin{center}
\begin{tabular}{lllc}
\toprule
  Modality & Entity & Sequence &      \\ 
\midrule

   \begin{tabular}{c}
    Text
  \end{tabular} &
  \begin{tabular}{c}
    Abell 370
  \end{tabular} &
  \begin{tabular}{c}
    \verb|Abell 370 is a cluster...|
  \end{tabular} &
  \begin{tabular}{c}
    \includegraphics[height=1.1cm]{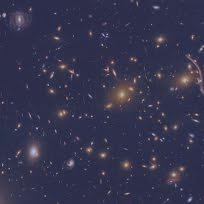}
  \end{tabular} \\

  \begin{tabular}{c}
    \LaTeX
  \end{tabular} &
  \begin{tabular}{c}
    Schwarzschild radius
  \end{tabular} &
  \begin{tabular}{c}
    \verb|r_{s} = \frac{2GM}{c^2}|
  \end{tabular} &
  \begin{tabular}{c}
    \includegraphics[height=1.1cm]{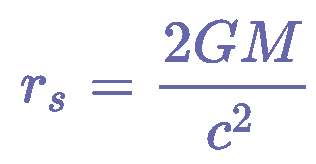}
  \end{tabular} \\

    \begin{tabular}{c}
    Code
  \end{tabular} &
  \begin{tabular}{c}
    Transformer
  \end{tabular} &
  \begin{tabular}{c}
    \verb|class Transformer(nn.Module)|
  \end{tabular} &
  \begin{tabular}{c}
    \includegraphics[height=1.1cm]{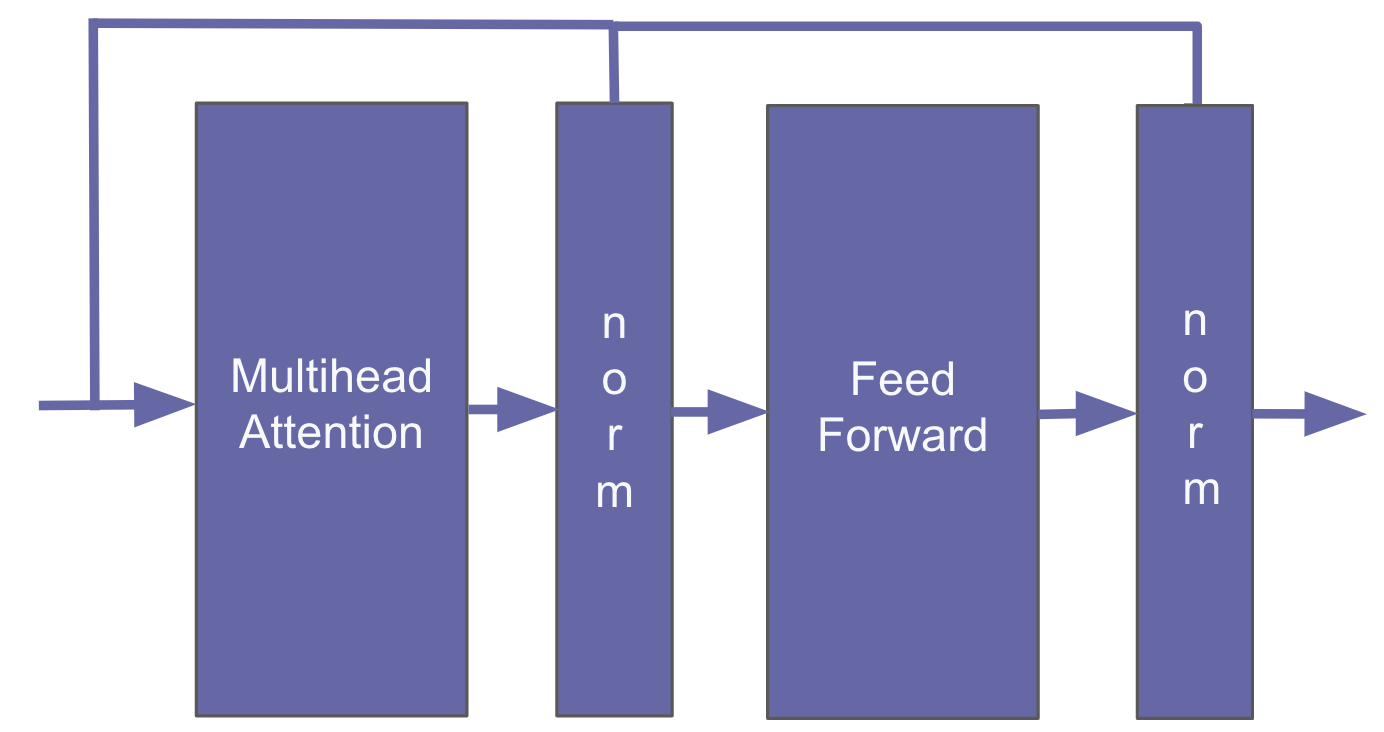}
  \end{tabular} \\

    \begin{tabular}{c}
    SMILES
  \end{tabular} &
  \begin{tabular}{c}
    Glycine
  \end{tabular} &
  \begin{tabular}{c}
    \verb|C(C(=O)O)N|
  \end{tabular} &
  \begin{tabular}{c}
    \includegraphics[height=1.1cm]{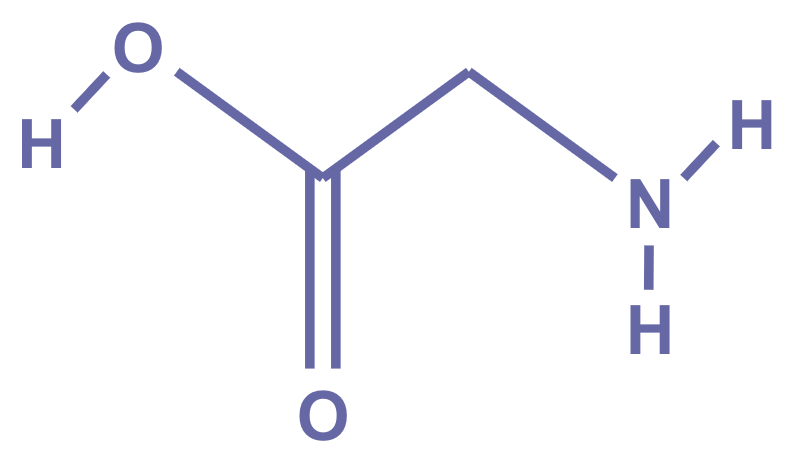}
  \end{tabular} \\
  
  \begin{tabular}{c}
    AA Sequence
  \end{tabular} &
  \begin{tabular}{c}
    Collagen $\alpha$-1(II) chain
  \end{tabular} &
  \begin{tabular}{c}
    \verb|MIRLGAPQTL..|
  \end{tabular} &
  \begin{tabular}{c}
    \includegraphics[height=1.1cm, width=2.6cm]{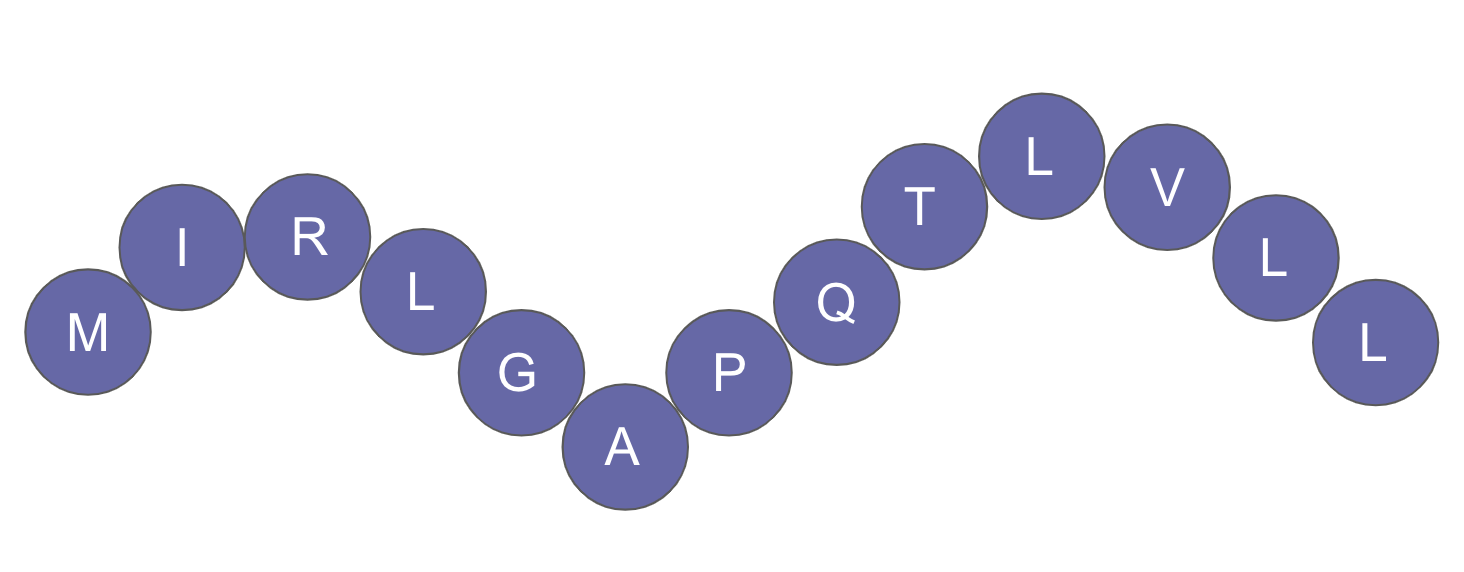}
  \end{tabular} \\
  
    \begin{tabular}{c}
    DNA Sequence
  \end{tabular} &
  \begin{tabular}{c}
    Human genome
  \end{tabular} &
  \begin{tabular}{c}
    \verb|CGGTACCCTC..|
  \end{tabular} &
  \begin{tabular}{c}
    \includegraphics[height=1.1cm]{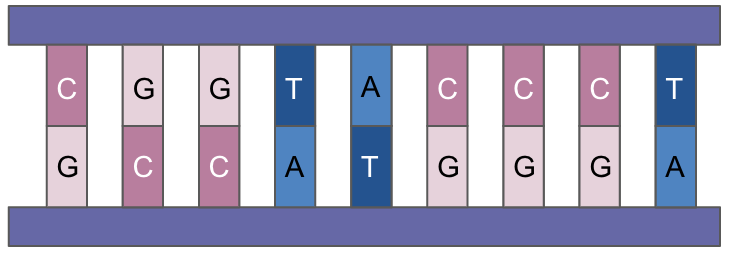}
  \end{tabular} \\

\hline
\end{tabular}
\end{center}
\caption{\textbf{Tokenizing Nature}. Galactica trains on text sequences that represent scientific phenomena.}
\label{table:naturebook-modalities}
\end{table}

\begin{table}[t!]
\begin{center}
\begin{tabular}{ lrrc } 
\toprule
  \multicolumn{4}{c}{Total dataset size = 106 billion tokens}      \\ 
  \midrule
  Data source & Documents & Tokens & Token \%     \\ 
\midrule
 Papers & 48 million & 88 billion & 83.0\% \\
 Code & 2 million & 7 billion & 6.9\%   \\ 
 Reference Material & 8 million & 7 billion & 6.5\% \\ 
 Knowledge Bases & 2 million & 2 billion & 2.0\% \\
 Filtered CommonCrawl & 0.9 million & 1 billion & 1.0\%  \\ 
 Prompts & 1.3 million & 0.4 billion & 0.3\%  \\ 
 Other & 0.02 million & 0.2 billion & 0.2\%  \\
\bottomrule
\end{tabular}
\end{center}
\caption{\textbf{The Galactica Corpus}. A full breakdown of these sources is contained in the Appendix.}
\label{table:naturebook-corpus}
\end{table}

\begin{displayquote}
“Nature is written in that great book which ever is before our eyes -- I mean the universe -- but we cannot understand it if we do not first learn the language and grasp the symbols in which it is written." \\ \\
\textit{Galileo Galilei, The Assayer}
\end{displayquote}

The idea that Nature can be understood in terms of an underlying language has a long history~\citep{Assayer, Wigner, Wheeler}. In recent years, deep learning has  been used to represent Nature, such as proteins and molecules~\citep{AlphaFold2021, MoLformer}. Amino acids are an alphabet in which the language of protein structure is written, while atoms and bonds are the language of molecules. At a higher level, we organize knowledge through natural language, and many works have trained on scientific text~\citep{SciBERT, BioLM, PubMedBERT, S2ORCBERT}. With Galactica, we train a single neural network on a large scientific corpus to learn the different languages of science.

Our corpus consists of \(106\) billion tokens from papers, reference material, encyclopedias and other scientific sources. We combine natural language sources, such as papers and textbooks, and natural sequences, such as protein sequences and chemical formulae. We process \LaTeX\  where we can capture it, and also include academic code to capture computational science. We highlight the corpus details in Table~\ref{table:naturebook-modalities} and~\ref{table:naturebook-corpus}. Full details, including dataset components and filtering logic, are contained in the Appendix.

Notably the dataset is small and curated compared to other LLM corpuses, which are larger and uncurated. This is a key question of this work: can we make a working LLM based on a curated, normative paradigm? If true, we could make more purposefully-designed LLMs by having a clear understanding of what enters the corpus, similar to expert systems which had normative standards~\citep{Jackson}.

\subsection{Tokenization}

\begin{figure}[t!]
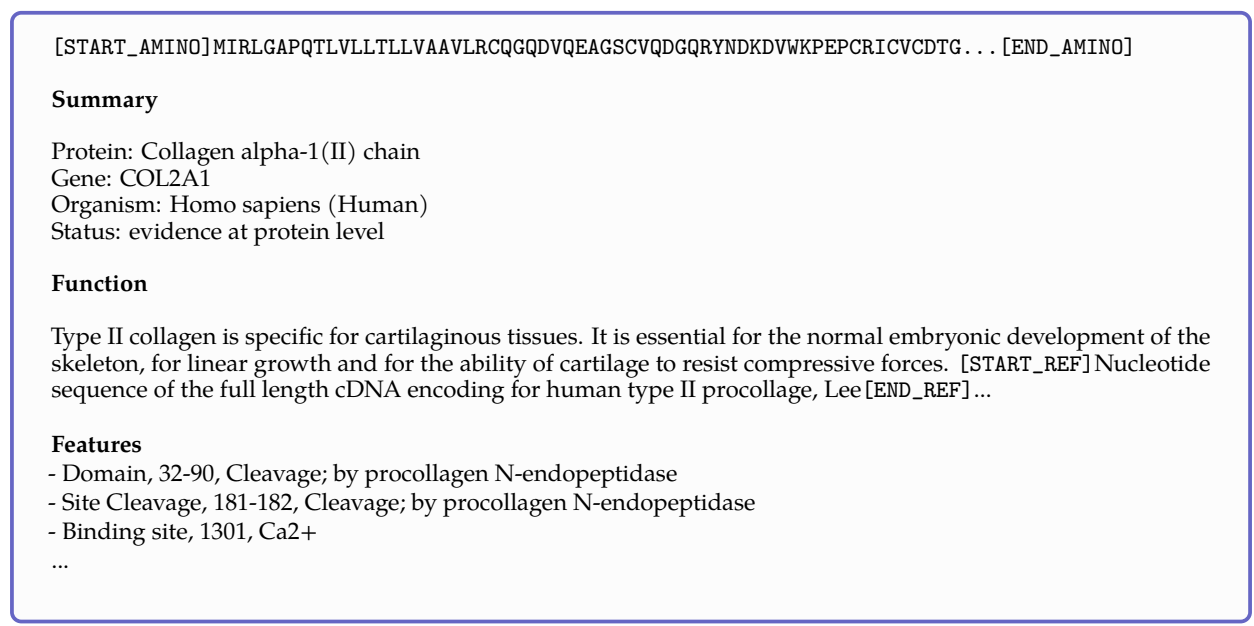

\begin{tcolorbox}[colback=galwhite,colframe=galpurple2]
\begin{small}
\verb|[START_AMINO]MIRLGAPQTLVLLTLLVAAVLRCQGQDVQEAGSCVQDGQRYNDKDVWKPEPCRICVCDTG...[END_AMINO]| \\

\textbf{Summary} \\

Protein: Collagen alpha-1(II) chain \\
Gene: COL2A1 \\
Organism: Homo sapiens (Human) \\
Status: evidence at protein level \\

\textbf{Function} \\

Type II collagen is specific for cartilaginous tissues. It is essential for the normal embryonic development of the skeleton, for linear growth and for the ability of cartilage to resist compressive forces. \verb|[START_REF]|Nucleotide sequence of the full length cDNA encoding for human type II procollage, Lee\verb|[END_REF]|... \\

\textbf{Features} \\
- Domain, 32-90, Cleavage; by procollagen N-endopeptidase \\
- Site Cleavage, 181-182, Cleavage; by procollagen N-endopeptidase \\
- Binding site, 1301, Ca2+ \\
... \\
\end{small}
\end{tcolorbox}
\caption{\textbf{Multi-Modal Data}.
A protein sequence occurs in a document context along with annotations, text and citations from UniProt. Full contents of the document are cut for clarity of exposition.
}
\label{fig:example_data}
\end{figure}

Tokenization is an important part of dataset design given the different modalities present. For example, protein sequences are written in terms of amino acid residues, where character-based tokenization is appropriate. To achieve the goal of \textit{specialized tokenization}, we utilize specialized tokens for different modalities:

\begin{enumerate}
 \item \textbf{Citations}: we wrap citations with special reference tokens \verb|[START_REF]| and \verb|[END_REF]|.
     \item \textbf{Step-by-Step Reasoning}: we wrap step-by-step reasoning with a working memory token \verb|<work>|, mimicking an internal working memory context.
  \item \textbf{Mathematics}: for mathematical content, with or without LaTeX, we split ASCII operations into individual characters. Parentheses are treated like digits. The rest of the operations allow for unsplit repetitions. Operation characters are \verb|!"#$%&'*+,-./:;<=>?\^_`|| and parentheses are \verb|()[]{}|.
  \item \textbf{Numbers}: we split digits into individual tokens. For example \verb|737612.62| -> \verb|7,3,7,6,1,2,.,6,2|.
  \item \textbf{SMILES formula}: we wrap sequences with \verb|[START_SMILES]| and \verb|[END_SMILES]| and apply character-based tokenization. Similarly we use \verb|[START_I_SMILES]| and \verb|[END_I_SMILES]| where isomeric SMILES is denoted. For example, \verb|C(C(=O)O)N| \(\rightarrow\) \verb|C,(,C,(,=,O,),O,),N|. 
  \item \textbf{Amino acid sequences}: we wrap sequences with \verb|[START_AMINO]| and \verb|[END_AMINO]| and apply character-based tokenization, treating each amino acid character as a single token. For example, \verb|MIRLGAPQTL| -> \verb|M,I,R,L,G,A,P,Q,T,L|.
  \item \textbf{DNA sequences}: we also apply a character-based tokenization, treating each nucleotide base as a token, where the start tokens are \verb|[START_DNA]| and \verb|[END_DNA]|. For example, \verb|CGGTACCCTC| -> \verb|C, G, G, T, A, C, C, C, T, C|.
 \end{enumerate}

 We cover a few of the specialized token approaches below that do not have clear parallels in the literature, in particular the working memory and citation tokens.

\subsubsection{Working Memory Token, <work>}

Transformer-based architectures lack an explicit working memory capability, which means a single-forward pass has limited efficacy. This is problematic for tasks that require multiple steps of computation. A current workaround is using a Transformer's output context as an external working memory to read from and write to. This is seen in recent work on chain-of-thought prompting~\citep{ChainOfThought, CoTBigBENCH}. In one sense this is intuitive, as humans also augment their limited working memory with scratchpads. In another sense, we would like models to refine their representations internally like humans; e.g. mental arithmetic.

\begin{figure}[t!]
\centering
\includegraphics[width=\textwidth]{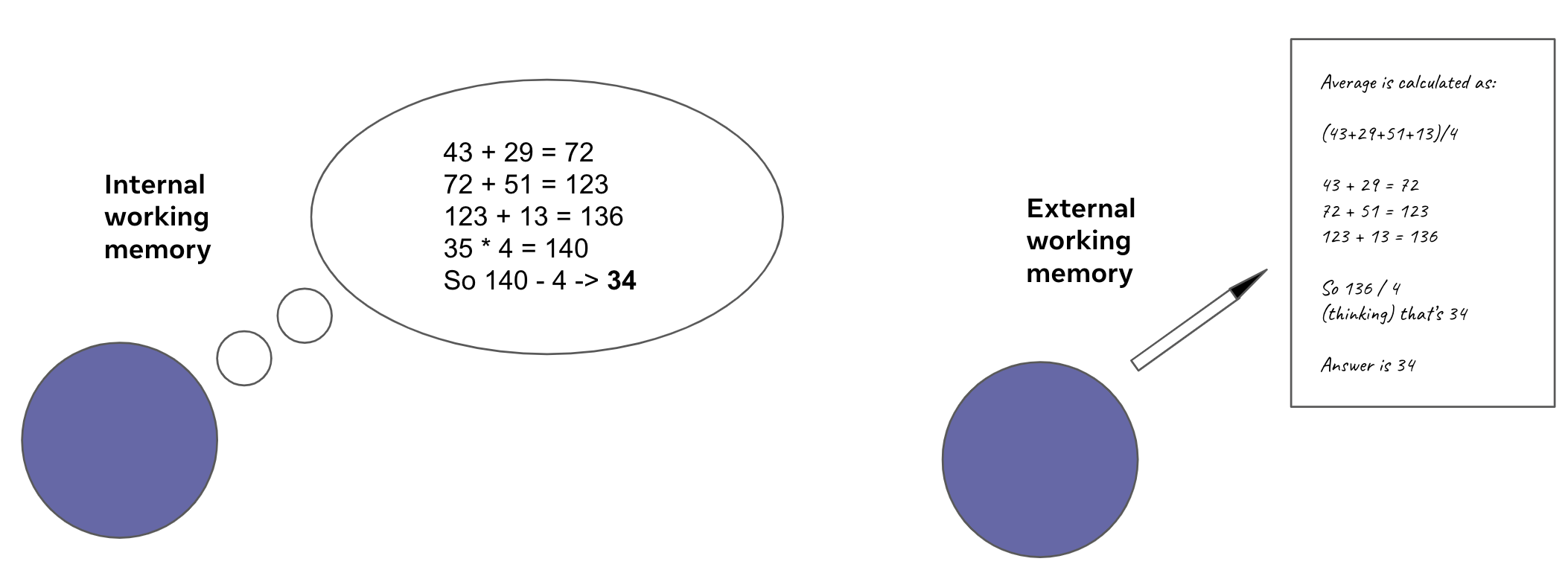}
\caption{Given a task like "What is the average of 43, 29, 51, 13?" a human can use internal or external working memory. In practice, they will use both symbiotically; meaning that working out that is written down in text is usually "missing" some steps performed internally.}
\end{figure}

There are two limitations with chain-of-thought. First, it relies on prompt discovery to find a prompt that elicits robust step-by-step reasoning; i.e. minimizes mistakes from doing too much in a single forward pass. Not only does this require finding a robust prompt that works in all cases, but it also often relies on few-shot examples which take up context space. What is worse, much of the step-by-step reasoning on the internet misses intermediate steps that a human has performed using internal memory. Humans do not write down every step they perform because it would lead to long and tedious answers. They write down the principal steps of reasoning, and do lower-level steps via internal working memory. This means there is "missing data" in written text, i.e. between written steps there are internal memory steps that are not explicitly stated.

Secondly, chain-of-thought prompting uses the neural network to perform tasks that it is arguably not best suited to doing; for example, arithmetic. Prior work has shown that accuracy on tasks like multiplication is proportional to term frequency~\citep{PretrainingFrequency}. Given that classical computers are specialized for tasks like arithmetic, one strategy is to offload these tasks from the neural network to external modules. For example, prior work has looked at the possibilities of external tool augmentation, such as calculators~\citep{LaMBDA}. However, this requires a strategy to identify where the neural network should offload; and it may not be straightforward when combined with a discovered zero-shot prompt, especially where lower-level computation steps are not explicitly stated in writing.

Our solution is a working memory token we call \verb|<work>|. We construct a few prompt datasets, see Table \ref{table:reasoning-datasets}, that wrap step-by-by-step reasoning within \verb|<work>| \verb|</work>|. Some of these datasets were generated programmatically (\textit{OneSmallStep}), by creating a problem template and sampling the variables, others were sourced online (\textit{Workout}, \textit{Khan Problems}), and others used existing datasets and transformed them into a \verb|<work>| based context (\textit{GSM8k train}). Where a computation is performed that a human could not do internally, we offload by writing and executing a Python script. An example is shown in Figure~\ref{fig:example_MATH}. Importantly, we do not have to turn this on, and the model can also predict the output from running a program. For our experiments, we did not find the need to turn Python offloading on, and leave this aspect to future work.

\begin{figure}
\begin{tcolorbox}[colback=galwhite,colframe=galpurple2]
\begin{small}
\textbf{Question:} A needle $35 \mathrm{~mm}$ long rests on a water surface at \(20^{\circ} \mathrm{C}\). What force over and above the needle's weight is required to lift the needle from contact with the water surface? \(\sigma = 0.0728 \mathrm{m}\).
\vspace{3mm}

\verb|<work>| \\

\[
    \begin{aligned}
        \sigma &= 0.0728 \mathrm{~N} / \mathrm{m} \\
        \sigma &= F/L \\
        0.0728 &= F / (2 \times 0.035) \\
        F &= 0.0728(2 \times 0.035)
    \end{aligned}
\]

\verb|calculate.py| \\
\verb|```| \\
\verb|f = 0.0728*(2*0.035)| \\ \\
\verb|with open("output.txt", "w") as file:| \\
\verb|    file.write(str(round(f, 5)))|\\
\verb|```| \\

<<run: "calculate.py"> \\

<<read: "output.txt">> \\

0.0051 \\

\verb|</work>| \\

\textbf{Answer:} \(F = 0.0051 \mathrm{~N}\)
\end{small}
\end{tcolorbox}
\caption{
\textbf{Model-Machine Symbiosis.} We show an example answer with the <work> working memory token. It performs exact steps for rearranging the equation, and when it reaches a calculation that it cannot solve reliably in a forward-pass, it writes a program, which can then be offloaded to a classical computer.
}
\label{fig:example_MATH}
\end{figure}

\begin{table}[h]
\vspace{20px}
\begin{center}
\begin{tabular}{ lrrr } 
\toprule
  Data source & Split & Prompts & Tokens     \\ 
\midrule
GSM8k~\citep{GSM8k} & \textit{train} & 7,473 & 3,518,467  \\
OneSmallStep & \textit{n/a} & 9,314 & 3,392,252  \\
Khan Problems~\citep{MATH} & \textit{n/a} & 3,835 & 1,502,644  \\
Workout & \textit{n/a} & 921 & 470,921  \\
 \midrule
\textbf{Total} & & 21,543 & 9 million \\
\bottomrule
\end{tabular}
\end{center}
\caption{\textbf{Reasoning Datasets} To train the model to use <work> we include several datasets in pre-training that incorporate this token. Full details are contained in the Appendix.}
\label{table:reasoning-datasets}
\end{table}


Longer term, an architecture change may be needed to support adaptive computation, so machines can have internal working memory on the lines of work such as adaptive computation time and PonderNet~\citep{ACT, PonderNet}. In this paper, we explore the \verb|<work>| external working memory approach as a bridge to the next step. Notably our \verb|<work>| prompt datasets are not very large or diverse, so there are likely large further gains to be made with this approach.

\clearpage

\subsubsection{Citation Token}

A distinctive properties of academic text is citations. In order to represent the implicit citation graph within the text, we process citations with global identifiers and special tokens \texttt{[START\_REF]} and \texttt{[END\_REF]} signifying when a citation is made. Figure~\ref{fig:citation-example} shows an example of citation processed text from a paper.

\begin{figure}[h]
\begin{tcolorbox}[colback=galwhite,colframe=galpurple2]
\begin{small}
Recurrent neural networks, long short-term memory \verb|[START_REF]|Long Short-Term Memory, Hochreiter\verb|[END_REF]| and gated recurrent \verb|[START_REF]|Empirical Evaluation
of Gated Recurrent Neural Networks on Sequence Modeling, Chung\verb|[END_REF]| neural networks
in particular, have been firmly established as state of the art approaches in sequence modeling and transduction problems such as language modeling and machine translation \verb|[START_REF]|Sequence to Sequence Learning with Neural
Networks, Sutskever\verb|[END_REF]|\verb|[START_REF]|Neural Machine Translation by Jointly
Learning to Align and Translate, Bahdanau\verb|[END_REF]|\verb|[START_REF]|Learning Phrase Representations Using RNN Encoder-Decoder for Statistical Machine Translation, Cho\verb|[END_REF]|.
\end{small}
\end{tcolorbox}
\caption{\textbf{Citation Processed Text}.
Example of citation processed text from \textit{Attention Is All You Need}~\citep{VaswaniSPUJGKP17}. For title-processed citations, the title can be associated with the previous context.
}
\label{fig:citation-example}
\end{figure}

We considered two type of citation identifier: (a) paper titles and (b) alphanumeric IDs. Based on ablations, we found that title based identifiers have greater citation prediction accuracy than IDs. However, we also found that paper titles are more prone to hallucination error at lower scales given the text-based nature of the identifier. We consider title processing for this paper, but we note the trade-offs between both approaches. Experiments for these ablations are contained in the Appendix.

\subsection{Prompt Pre-Training}

\begin{figure}[t!]
\centering
\includegraphics[width=0.8\textwidth]{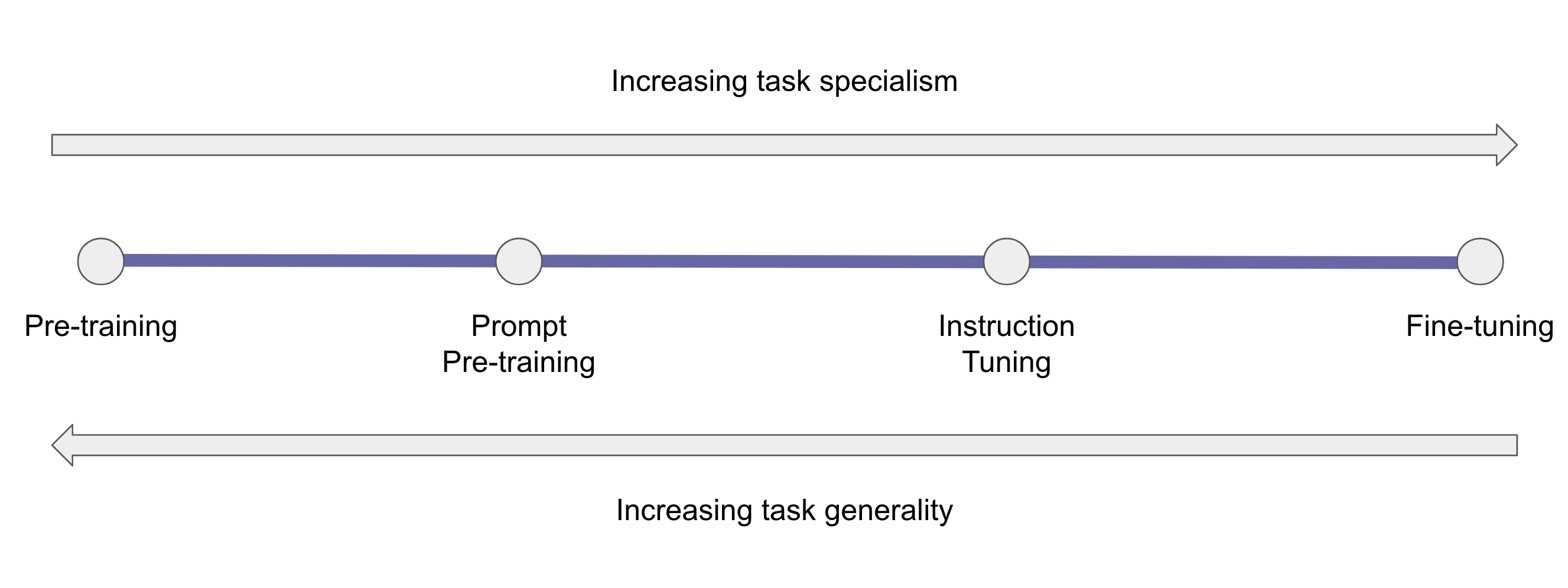}
\caption{\textbf{Prompt Pre-training}. Pre-training weighs all tokens equally as part of the self-supervised loss. This leads to a weak relative signal for tasks of interest, meaning model scale has to be large to work. Instruction tuning boosts performance \textit{post hoc}, and can generalize to unseen tasks of interest, but it risks performance in tasks that are distant from instruction set tasks. Prompt pre-training has a weaker task of interest bias than instruction tuning but less risk of degrading overall task generality.}
\end{figure}

We deviate from existing language model research in one important direction, which is our decision to include prompts in pre-training \textit{alongside} the general corpora. This is motivated by a number of observations.

First, existing work has shown the importance of training token count on performance. The Chinchilla paper derived scaling "laws" taking into account number of tokens, training a 70bn model for 1.4 trillion tokens~\citep{Chinchilla}. They obtained state-of-the-art performance on MMLU, beating much larger models such as Gopher~\citep{Gopher}. 

Separately, research such as FLAN and T0 showed prompt tuning can boost downstream performance~\citep{FLAN, T0, FLANPALM}. Their strategy involved converting tasks to text prompts, using prompt diversity in how the tasks are posed, and then fine-tuning on these prompt datasets. For FLAN and T0, this approach boosts performance, beating larger models such as GPT-3 on many tasks.

And additionally there is the UnifiedQA approach~\citep{UnifiedQA}. In this approach, a T5 model is fine-tuned on question answering datasets, and is shown to boost performance on out-of-domain question answering datasets~\citep{2020t5}. The model outperforms GPT-3 on MMLU, a model 16 times larger.

The first stream of research above focuses on total training tokens as a way to boost performance; i.e. it is \textit{token agnostic}. The second stream of research focuses on task-context tokens as a way to boost performance; i.e. it is \textit{token selective}. Since fine-tuned smaller models beat larger few-shot models on tasks like MMLU, this suggests world knowledge may be present in smaller models, but task-context knowledge may be poor given the relative number of task-context tokens seen in the general corpus. 

For this paper, we opt to augment pre-training data with more task prompts to boost performance at lower scales. This is advantageous if it obviates the need for more data scale, e.g. a  >\(1\) trillion corpus, or more model scale.  The largest 120B model we train runs on a single NVIDIA A100 node. Additionally, given that fine-tuning requires expertise, making the model work out-the-box for popular tasks like question answering and summarization is more useful for users of the model. Lastly, by including prompts alongside general data, we maximize the generality of the model while boosting performance on some tasks of interest.

The closest analog to this approach for large language models is ExT5~\citep{ExT5}. We take a similar approach by taking many machine learning training datasets, converting them to a text format, with prompt diversity, and then including them alongside general corpora in our pre-training set. A summary of prompt types is given in Table~\ref{table:prompt-breakdown}; the full details of datasets and prompts used are covered in the Appendix.

\begin{table}[h!]
\vspace{20px}
\begin{center}
\begin{tabular}{ lrr } 
\toprule
  Task & Prompts & Tokens     \\ 
\midrule
 Chemical Properties & 782,599 & 275 million  \\
 Multiple-Choice QA & 256,886 & 31 million  \\
 Extractive QA & 30,935 & 13 million  \\
 Summarization & 6,339 & 11 million  \\
 Entity Extraction & 156,007 & 9 million  \\
 Reasoning & 21,543 & 9 million  \\
 Dialog & 18,930 & 5 million  \\
 Binary QA & 36,334 & 4 million  \\
 Other & 3,559 & 1 million  \\
\midrule
\textbf{Total} & 783,599 & 358 million \\
\bottomrule
\end{tabular}
\end{center}
\caption{\textbf{Pre-training Prompts}. We include zero-shot prompts in pre-training to boost the task signal.}
\label{table:prompt-breakdown}
\end{table}

Because of prompt inclusion, it is important to distinguish between in-domain performance, where the training dataset is included in pre-training, and out-of-domain performance, where the training dataset is not included in pre-training. We mark these results clearly in the Results section of this paper. Importantly, we do not advocate for prompt pre-training as an alternative to instruction tuning. In fact, instruction tuning on Galactica is likely useful follow-up work given its potential to boost performance on several tasks of interest.

\clearpage

\section{Method}

\subsection{Architecture}

Galactica uses a Transformer architecture in a decoder-only setup~\citep{VaswaniSPUJGKP17}, with the following modifications:

\begin{itemize}
    \item \textbf{GeLU Activation} - we use GeLU activations for all model sizes~\citep{GeLU}.
    \item \textbf{Context Window} - we use a 2048 length context window for all model sizes.
    \item \textbf{No Biases} - following PaLM, we do not use biases in any of the dense kernels or layer norms~\citep{PaLM}.
    \item \textbf{Learned Positional Embeddings} - we use learned positional embeddings for the model. We experimented with ALiBi at smaller scales but did not observe large gains, so we did not use it~\citep{ALiBi}.
    \item \textbf{Vocabulary} - we construct a vocabulary of 50k tokens using BPE~\citep{BPE}. The vocabulary was generated from a randomly selected 2\% subset of the training data.
\end{itemize}

\subsection{Models}

The different model sizes we trained, along with training hyperparameters are outlined in Table~\ref{table:models-trained}.

\begin{table}[h!]
\begin{center}
\begin{tabular}{ lcccccccc } 
\toprule
  Model & $n_{params}$ & $n_{layers}$ & $d_{model}$ & $n_{heads}$ & $d_{heads}$ & Batch Size & Max LR & Warmup    \\ 
\midrule
 GAL 125M & 125M & 12 & 768 & 12 & 64 & 0.5M & $6 \times 10^{-4}$ & 375M  \\ 
 GAL 1.3B & 1.3B & 24 & 2,048 & 32 & 64 & 1.0M & $2 \times 10^{-4}$ & 375M \\
 GAL 6.7B & 6.7B & 32 & 4,096 & 32 & 128 & 2.0M & $1.2 \times 10^{-4}$ & 375M \\
 GAL 30B & 30.0B & 48 & 7,168 & 56 & 128 & 2.0M & $1 \times 10^{-4}$ & 375M \\
 GAL 120B & 120.0B & 96 & 10,240 & 80 & 128 & 2.0M & $0.7 \times 10^{-5}$ & 1.125B \\
\bottomrule
\end{tabular}
\end{center}
\caption{Details of the models trained}
\label{table:models-trained}
\end{table}

We train using AdamW with $\beta_{1}= 0.9$, $\beta_{2} = 0.95$ and weight decay of $0.1$~\citep{AdamW}. We clip the global norm of the gradient at 1.0, and we use linear decay for learning rate down to 10\% of it value. We use dropout and attention dropout of $p=0.1$. We do not use embedding dropout. We found longer warmup was important for the largest model in the early stages of training to protect against the effects of bad initialization, which can have long-memory effects on the optimizer variance state and slow down learning. This may be specific to our model and training setup, and it is not clear whether this advice generalizes.

\subsection{Libraries and Infrastructure}

We use the metaseq library\footnote{\href{https://github.com/facebookresearch/metaseq/}{https://github.com/facebookresearch/metaseq/}} for training the models, built by the NextSys team at Meta AI.

For training the largest 120B model, we use 128 NVIDIA A100 80GB nodes. For inference Galactica 120B requires a single A100 node. We choose the maximum model size to obey this constraint for downstream accessibility, and we will work to improve its accessibility for the research community in coming months.

\clearpage

\section{Results}

\begin{figure}[t!]
    \centering
    \includegraphics[width=\textwidth]{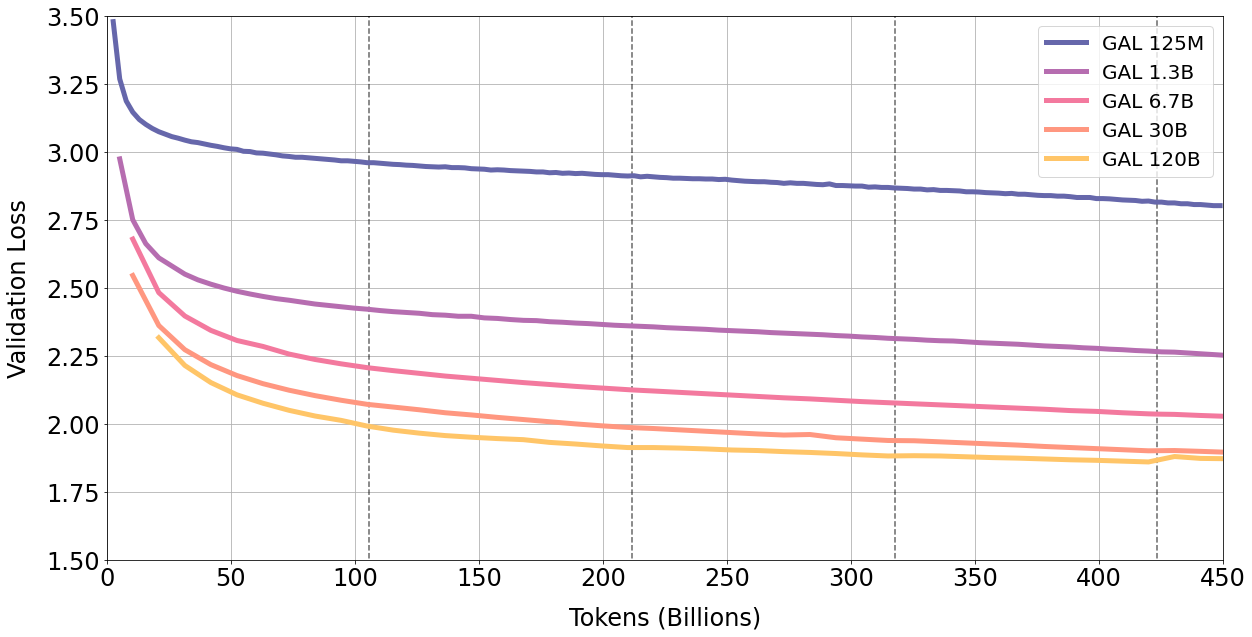}
    \caption{\textbf{Repeated Tokens and Validation Loss}. With four epochs of training, we continue to see validation loss fall for all model sizes. For the 120B model we see the first signs of overfitting at the beginning of the fifth epoch, and we early stop at this point.}
    \label{fig:validation_loss}
\end{figure}

\subsection{Repeated Tokens Considered Not Harmful}

We train the models for 450 billion tokens, or approximately 4.25 epochs. We find that performance continues to improve on validation set, in-domain and out-of-domain benchmarks with multiple repeats of the corpus.

First, from Figure~\ref{fig:validation_loss}, validation loss continues to fall with four epochs of training. The largest 120B model only begins to overfit at the start of the fifth epoch. This is unexpected as existing research suggests repeated tokens can be harmful on performance~\citep{HarmfulRepeats}. We also find the 30B and 120B exhibit a epoch-wise double descent effect of plateauing (or rising) validation loss followed by a decline. This effect becomes stronger with each epoch, and is most visible above with the 120B model towards end of training.

To investigate further, we examine the per-source breakdown of validation loss to see if there is heterogeneity in loss behaviour. We plot example curves in Figure~\ref{fig:source_validation_loss} overleaf for the 30B model. We see no signs of loss heterogeneity: loss falls for all sources. The 120B exhibits the same relative trend of declining validation loss for all sources until the beginning of fifth epoch, where all sources spike (see Appendix).

\begin{figure}[h!]
    \centering
    \includegraphics[width=\textwidth]{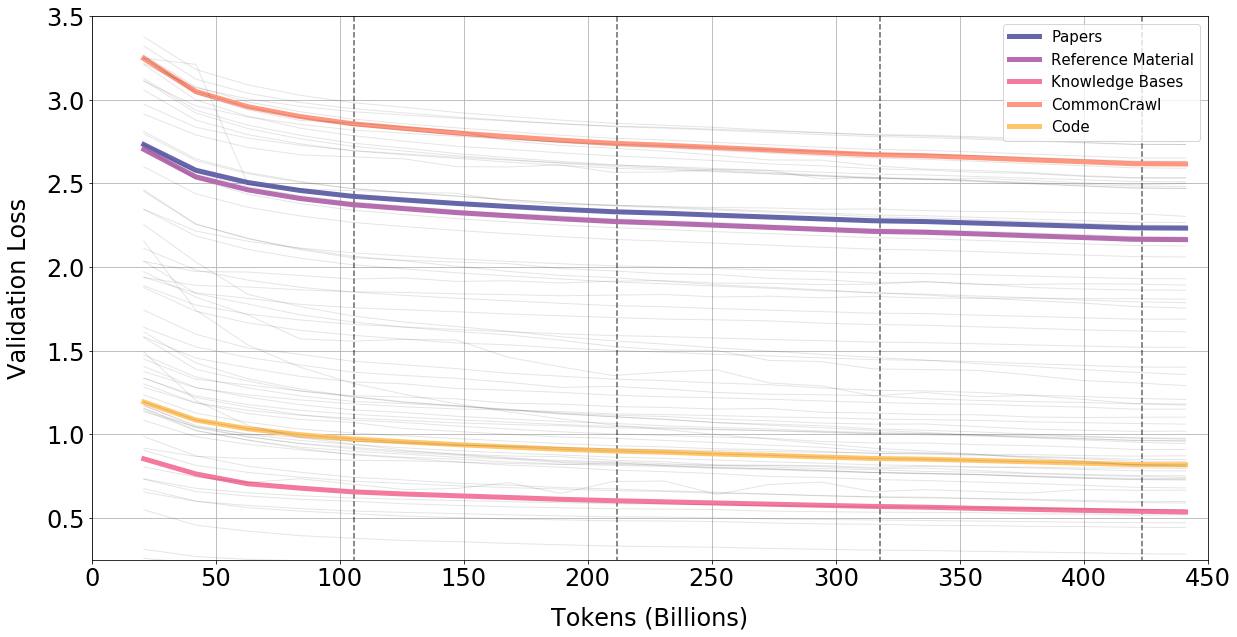}
    \caption{\textbf{Validation Loss Per Source}. Validation loss falls through training for all dataset categories. Results are shown for the 30B model above. The 120B exhibits the same relative trend of declining validation loss for all sources until the beginning of fifth epoch, where all sources spike (see Appendix).}
    \label{fig:source_validation_loss}
\end{figure}

\begin{figure}[t!]
    \centering
    \includegraphics[width=\textwidth]{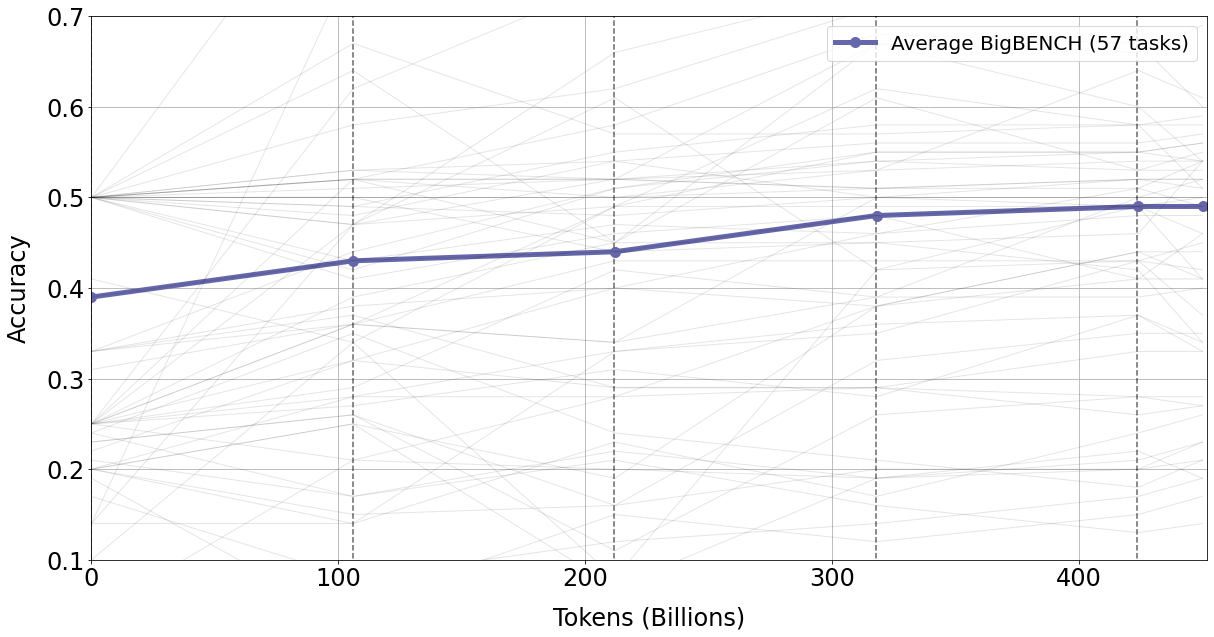}
    \caption{\textbf{BIG-bench Performance During Training}. The 57 task selection from BIG-bench contains principally non-scientific tasks. We use it as a proxy for \textit{out-of-domain} performance. For the 120B model above, we see no signs of overfitting after four repeats of the corpus.}
    \label{fig:downstream_scale}
\end{figure}

The next question to answer is whether this trend extends to downstream performance and out-of-domain generalization. For this we use a 57 task subset of \textit{BIG-bench} subset, a general corpus with principally non-scientific tasks and prompt types not included in pre-training~\citep{BIGBenchakasomanyauthorsitdoesntfitinthecontextwindow}. We plot results in Figure~\ref{fig:downstream_scale}. We see no signs of overfitting suggesting that use of repeated tokens is improving downstream performance as well as upstream performance.

We suspect that two factors could be at play, a \textit{quality factor}, the curated nature of the corpus enables more value per token to be extracted, or a \textit{modality factor}, the nature of scientific data enables more value per token to be extracted. The missing step of causation is what leads specifically from either factor towards less overfitting, and we leave this question to further work. We note the implication that the "$\text{tokens} \rightarrow \infty$" focus of current LLM projects may be overemphasised versus the importance of filtering the corpus for quality.

In the following sections, we turn to evaluating Galactica's scientific capabilities. Specifically, we focus on the high-level design goals of building an LLM that can store, combine and reason about scientific knowledge - as these are needed for building a new interface for science.

\clearpage

\subsection{Knowledge Probes}

First, we examine how well Galactica absorbs scientific knowledge. We set up several knowledge probe benchmarks, building off the LAMA approach of~\citet{petroni2019language}. These were critical metrics during model development for identifying knowledge gaps within the corpus, and informing how to iterate the corpus. They also provide insight into the relative knowledge strengths of Galactica versus general language models, and we cover these results in this section before turning to the downstream tasks.

\subsubsection{LaTeX Equations}

We construct a dataset of popular LaTeX equations from the fields of chemistry, physics, mathematics, statistics and economics. Memorisation of equations is useful to measure as it is necessary for many downstream tasks; for example, recalling an equation to use as part of an answer to a problem. Unless stated explicitly, Galactica results are reported as zero-shot. In total there are 434 equations we test for the knowledge probe.

We prompt with an equation name and generate LaTeX. An example is shown in Figure~\ref{fig:latex_ex}.

\begin{figure}[h]
\begin{tcolorbox}[colback=galwhite,colframe=galpurple2]
\begin{small}
\textbf{Prompt} \\

The formula for Bessel's differential equation is: \\

\textbf{Generated Answer}

$$ x^{2}{\frac {d^{2}y}{dx^{2}}}+x{\frac {dy}{dx}}+\left(x^{2}-\alpha ^{2}\right)y=0 $$

\end{small}
\end{tcolorbox}
\caption{\textbf{LaTeX Equations Probe}.
We prompt for the name of an equation and evaluate whether the generated LaTeX is correct. We manually evaluate given the possibility of multiple correct answers.
}
\label{fig:latex_ex}
\end{figure}

We summarize the results in Table~\ref{table:latex-perf}. Equation knowledge increases smoothly with scale. Galactica outperforms larger language models trained on general corpuses, indicating the value of a curated dataset.

\begin{table}[h!]
\begin{center}
\begin{tabular}{ lrrrrrrr } 
\toprule
  Model & Params (bn) & Chemistry & Maths & Physics & Stats & Econ & Overall \\ 
 \midrule
 OPT & 175 & 34.1\% & 4.5\% & 22.9\% & 1.0\% & 2.3\% & 8.9\% \\  
 BLOOM & 176 & 36.3\% & 36.1\% & 6.6\% & 14.1\% & 13.6\% & 21.4\% \\ 
 GPT-3 (\verb|text-davinci-002|) & ? & 61.4\% & 65.4\% & 41.9\% & 25.3\% & 31.8\% & 49.0\% \\  
\midrule
 GAL 125M & 0.1 & 0.0\% & 0.8\% & 0.0\% & 1.0\% & 0.0\% & 0.5\% \\
 GAL 1.3B & 1.3 & 31.8\% & 26.3\% & 23.8\% & 11.1\% & 4.6\% & 20.5\% \\ 
 GAL 6.7B & 6.7 & 43.2\% & 59.4\% & 36.2\% & 29.3\% & 27.3\% & 41.7\% \\
 GAL 30B & 30 & 63.6\% & 74.4\% & 35.2\% & 40.4\% & 34.1\% & 51.5\% \\  
 GAL 120B & 120 & \textbf{79.6\%} & \textbf{83.5\%} & \textbf{72.4\%} & \textbf{52.5\%} & \textbf{36.4\%} & \textbf{68.2\%} \\   
\bottomrule
\end{tabular}
\end{center}
\caption{\textbf{Results on LaTeX equations}. Results are evaluated zero-shot.}
\label{table:latex-perf}
\end{table}

\subsubsection{Domain Probes}

We also set up domain probes to track specialized knowledge for certain fields. We detail these below:

\begin{itemize}
    \item \textbf{AminoProbe}: a dataset of names, structures and  properties of the 20 common amino acids.
    \item \textbf{BioLAMA}: a dataset of biomedical factual knowledge triples.
    \item \textbf{Chemical Reactions}: a dataset of chemical reactions.
    \item \textbf{Galaxy Clusters}: a dataset of galaxy clusters with their constellation classifications.
    \item \textbf{Mineral Groups}: a dataset of minerals and their mineral group classifications.
\end{itemize}

\begin{figure}[h!]
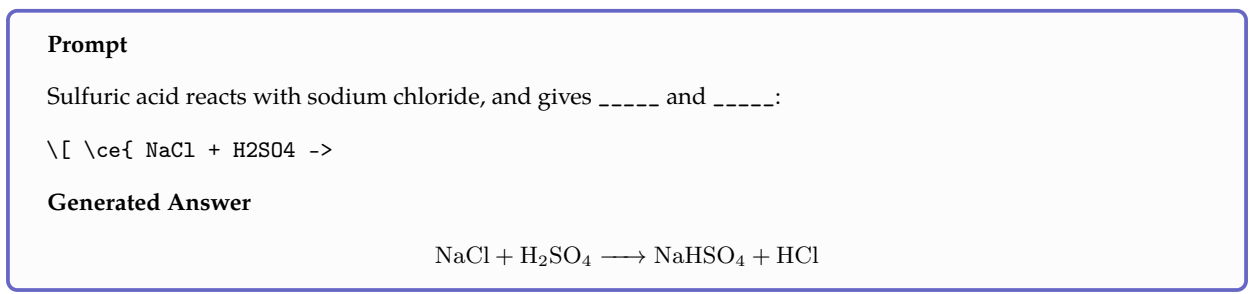

\begin{tcolorbox}[colback=galwhite,colframe=galpurple2]
\begin{small}
\textbf{Prompt} \\

Sulfuric acid reacts with sodium chloride, and gives \verb|_____| and \verb|_____|: \\

\verb|\[ \ce{ NaCl + H2SO4 ->| \\

\textbf{Generated Answer}

$$ \ce{NaCl + H2SO4 -> NaHSO4 + HCl} $$

\end{small}
\end{tcolorbox}
\caption{\textbf{Chemical Reactions}.
We prompt based on a description and reactants, and evaluate whether the generated products are correct.
}
\label{fig:chemical_reactions}
\end{figure}

In each case, we construct a prompt to test the knowledge. For example, for \textbf{Chemical Reactions}, we ask Galactica to predict the products of the reaction in the chemical equation LaTeX. We mask out products in the description so the model is inferring based on the reactants only. An example is shown in Figure~\ref{fig:chemical_reactions}.

We report results for these knowledge probes in Table~\ref{table:domain_probes}.

\begin{table}[h!]
\begin{center}
\begin{tabular}{ lrrrrrr } 
\toprule
  Model & Params (bn) & Amino & BioLAMA & Reactions & Clusters & Minerals    \\ 
 \midrule
  OPT & 175 & 12.0\% & 7.1\% & 12.7\% & 21.7\% & 1.6\% \\ 
  BLOOM & 176 & 14.0\% & \textbf{9.7\%} & 22.4\% & 15.0\% & 10.3\% \\   
 GPT-3 (\verb|text-davinci-002|) & ? & 14.0\% & 8.4\% & 35.1\% & 20.8\% & 18.3\% \\  
 \midrule
 GAL 125M & 0.1 & 12.0\% & 3.1\% & 0.3\% & 6.7\% & 0.0\% \\
 GAL 1.3B & 1.3 & 16.0\% & 7.2\% & 14.4\% & 14.2\% & 10.3\% \\
 GAL 6.7B & 6.7 & 17.0\% & 7.9\% & 26.4\% & 17.5\% & 8.7\% \\
 GAL 30B & 30 & 21.0\% & 6.9\% & 36.5\% & 20.0\% & 17.5\% \\  
 GAL 120B & 120 & \textbf{21.0\%} & 8.0\% & \textbf{43.1\%} & \textbf{24.2\%} & \textbf{29.4\%} \\ 
\bottomrule
\end{tabular}
\end{center}
\caption{\textbf{Results on Domain Probes}. Results are evaluated zero-shot.}
\label{table:domain_probes}
\end{table}

We also observe steady scaling behaviour in these knowledge probes, with the exception of BioLAMA which we suspect reflects zero-shot prompt difficulty for all LLMs. Notably fine-grained factual knowledge, such as "\texttt{ConstellationOf(GalaxyCluster)}" type-queries seems to scale smoothly with the size of the model.

\clearpage

\subsubsection{Reasoning}

We now turn to reasoning capabilities with the \verb|<work>| token. We start by evaluating on the \textbf{MMLU} mathematics benchmarks, which we report in Table~\ref{table:mmlu-maths-perf}~\citep{MMMLU}. Galactica performs strongly compared to larger base models, and use of the \verb|<work>| token appears to boost performance over Chinchilla, even for the smaller 30B Galactica model.

\begin{table}[h!]
\begin{center}
\begin{tabular}{ lrrrrrrr } 
\toprule
  \multicolumn{8}{c}{Mathematics MMLU}      \\ 
\midrule
  Model & Params (bn) & A.Algebra & Elem & HS & College & F. Logic & Average \\ 
   \midrule
  BLOOM (5-shot) & 176 & 25.0\% & 26.7\% & 27.0\% & 25.0\% & 26.2\% & 26.4\% \\
 OPT (5-shot) & 175 & 21.0\% & 25.7\% & 24.4\% & 33.0\% & 29.4\% & 26.7\% \\
 Gopher (5-shot) & 280 & 25.0\% & 33.6\% & 23.7\% & 37.0\% & 35.7\% & 30.6\% \\
  Chinchilla (5-shot) & 70 & 31.0\% & 41.5\% & 31.9\% & 32.0\% & 33.3\% & 35.7\% \\
 \midrule
 GAL 1.3B & 1.3 & 28.0\% & 27.2\% & 26.7\% & 30.0\% & 24.6\% & 27.1\% \\
 GAL 6.7B & 6.7 & 28.0\% & 28.9\% & 26.7\% & 36.0\% & 31.0\% & 29.2\% \\
 GAL 30B & 30 & 30.0\% & 30.2\% & 26.3\% & 36.0\% & 31.7\% & 29.9\% \\
 GAL 120B & 120 & 33.0\% & 38.1\% & 32.6\% & 43.0\% & 32.5\% & 35.8\% \\
 \midrule
 GAL 1.3B \verb|<work>| & 1.3 & 22.0\% & 24.6\% & 18.9\% & 25.0\% & 31.0\% & 24.6\% \\
 GAL 6.7B \verb|<work>| & 6.7 & \textbf{33.3\%} & 30.7\% & 25.2\% & 26.0\% & 33.3\% & 28.0\% \\
 GAL 30B \verb|<work>| & 30 & 33.0\% & 41.5\% & 33.3\% & 39.0\% & 37.3\% & 37.1\%\\
GAL 120B \verb|<work>| & 120 & 27.0\% & \textbf{54.2\%} & \textbf{37.0\%} & \textbf{44.0\%} & \textbf{40.5\%} & \textbf{41.3\%} \\
   \bottomrule
\end{tabular}
\end{center}
\caption{\textbf{Results on Mathematics MMLU}. Galactica is evaluated without few-shot examples. With the <work> token we see large gains in performance. Results are on MMLU test.}
\label{table:mmlu-maths-perf}
\end{table}

We also evaluate on the MATH dataset to further probe the reasoning capabilities of Galactica~\citep{MATH}. We compare the \verb|<work>| token prompt directly with the Minerva 5-shot chain-of-thought prompt \verb|mCoT| for comparability. We report results in Table~\ref{table:math-benchmark-results}.

\begin{table}[h!]
\begin{center}
\begin{tabular}{ lrrrrrrrr } 
\toprule
  \multicolumn{9}{c}{MATH Results}      \\ 
\midrule
  Model & Alg & CProb & Geom & I.Alg & N.Theory & Prealg & Precalc & Average \\ 
  \midrule
  \multicolumn{9}{c}{Base Models}      \\ 
   \midrule
GPT-3 175B (8-shot) & 6.0\% & 4.7\% & 3.1\% & 4.4\% & 4.4\% & 7.7\% & 4.0\% & 5.2\% \\
PaLM 540B (5-shot) \verb|mCoT| & 9.7\% & 8.4\% & 7.3\% & 3.5\% & 6.0\% & 19.2\% & 4.4\% & 8.8\% \\
 GAL 30B \verb|<work>| & 15.8\% & 6.3\% & 5.8\% & 4.9\% & 2.4\% & 19.4\% & 8.2\% & 11.4\%\\
 GAL 30B (5-shot) \verb|mCoT| & 17.9\% & 6.8\% & 7.9\% & 7.0\% & 5.7\% & 17.9\% & 7.9\% & 12.7\% \\
 GAL 120B \verb|<work>| & 23.1\% & 10.1\% & 9.8\% & 8.6\% & 6.5\% & 23.8\% & 11.7\% & 16.6\% \\
GAL 120B (5-shot) \verb|mCoT| & 29.0\% & 13.9\% & 12.3\% & 9.6\% & 11.7\% & 27.2\% & 12.8\% & 20.4\% \\
 \midrule
  \multicolumn{9}{c}{Fine-tuned LaTeX Models}      \\ 
 \midrule
Minerva 540B (5-shot) \verb|mCoT| & 51.3\% & 28.0\% & 26.8\% & 13.7\% & 21.2\% & 55.0\% & 18.0\% & 33.6\% \\
   \bottomrule
\end{tabular}
\end{center}
\caption{\textbf{Results on MATH}. With both the chain-of-thought and \texttt{<work>} token prompts, Galactica exceeds PaLM's performance with 18 times less capacity.}
\label{table:math-benchmark-results}
\end{table}

We see that Galactica outperforms the base PaLM model by a significant margin, with both chain-of-thought and \verb|<work>| prompts. Galactica 30B outperforms PaLM 540B on both prompts: an 18 times smaller model. This suggests Galactica may be a better base model for fine-tuning towards mathematical tasks.

We report Minerva results for completeness, which is a 540B PaLM fine-tuned towards LaTeX specifically. Minerva outperforms base Galactica, but the performance differences are non-uniform; which points towards different mathematical data biases. For a direct comparison to Minerva, the model is freely available for those who want to finetune Galactica towards LaTeX specifically as follow-up work.

\clearpage

\subsection{Downstream Scientific NLP}

We now evaluate on downstream scientific tasks to see how well Galactica can compose its knowledge in different task contexts. We focus on knowledge-intensive scientific tasks and report full results in Table~\ref{table:full-qa}. For this we use the MMLU benchmark as well as some other popular scientific QA benchmarks. We include the MMLU results earlier without <work> to test for knowledge association specifically. Full MMLU results, including social sciences and other fields, are reported in the Appendix. We also perform data leakage analysis on these benchmarks for more confidence; results are in the Appendix.

From Table~\ref{table:full-qa}, Galactica can compose its knowledge into the question-answering task, and performance is strong; significantly outperforming the other open language models, and outperforming a larger model (Gopher 280B) in the majority of tasks. Performance against Chinchilla is more variable, and Chinchilla appears to be  stronger in a subset of tasks: in particular, high-school subjects and less-mathematical, more memorization intensive tasks. In contrast, Galactica tends to perform better in mathematical and graduate-level tasks.

Our working hypothesis is that the Galactica corpus is biased towards graduate scientific knowledge, given it consists mostly of papers, which explains lagging performance in high-school subjects. While we do pick up some high-school level content through encyclopedias, textbooks and the filtered CommonCrawl, this amounts to a small quantity of tokens (a few billion). We leave the question of how to capture more of this base scientific knowledge in a curated way to future work.

On remaining tasks, we achieve state-of-the-art results over fine-tuned models at the time of writing. On PubMedQA, we achieve a score of 77.6\% which outperforms the state-of-the-art of 72.2\%~\citep{BioLinkBERT}. On MedMCQA dev we achieve score of 52.9\% versus the state-of-the-art of 41.0\%~\citep{PubMedBERT}. For BioASQ and MedQA-USMLE, performance is close to the state-of-the-art performance of fine-tuned models (94.8\% and 44.6\%)~\citep{BioLinkBERT}.

\begin{table}[h]
\begin{center}
\begin{tabular}{ llc|ccccc } 
\toprule
  Dataset & Domain & GAL & OPT & BLOOM & GPT-3 & Gopher & Chinchilla   \\ 
\midrule
Abstract Algebra & \textit{out-of-domain} & \textbf{33.3\%} & 21.0\% & 25.0\% & - & 25.0\% & 31.0\% \\
ARC Challenge & \textit{in-domain} & \textbf{67.9\%} & 31.1\% & 32.9\% & 51.4\% & - & - \\
ARC Easy & \textit{in-domain} & \textbf{83.8\%} & 37.4\% & 40.7\% & 68.8\% & - & - \\
Astronomy & \textit{out-of-domain} & 65.1\% & 23.0\% & 25.7\% & - & 65.8\% & \textbf{73.0\%} \\
BioASQ & \textit{in-domain} & \textbf{94.3\%} & 81.4\% & 91.4\% & - & - & - \\
Biology (College) & \textit{out-of-domain} & 68.8\% & 30.6\% & 28.5\% & - & 70.8\% & \textbf{79.9\%}  \\
Biology (High-School) & \textit{out-of-domain} & 69.4\% & 27.7\% & 29.4\% & - & 71.3\% & \textbf{80.3\%}   \\
Chemistry (College) & \textit{out-of-domain} & 46.0\% & 30.0\% & 19.0\% & - & 45.0\% & \textbf{51.0\%}  \\
Chemistry (High-School) & \textit{out-of-domain} & 47.8\% & 21.7\% & 23.2\% & - & 47.8\% & \textbf{58.1\%}   \\
Comp. Science (College) & \textit{out-of-domain} & 49.0\% & 17.0\% & 6.0\% & - & 49.0\% & \textbf{51.0\%}  \\
Comp. Science (High-School) & \textit{out-of-domain} & \textbf{70.0\%} & 30.0\% & 25.0\% & - & 54.0\% & 58.0\%   \\
Econometrics & \textit{out-of-domain} & 42.1\% & 21.0\% & 23.7\% & - & \textbf{43.0\%} & 38.6\%   \\
Electrical Engineering & \textit{out-of-domain} & \textbf{62.8\%} & 36.6\% & 32.4\% & - & 60.0\% & 62.1\%   \\
Elementary Mathematics & \textit{out-of-domain} & 38.1\% & 25.7\% & 27.6\% & - & 33.6\% & \textbf{41.5\%} \\
Formal Logic & \textit{out-of-domain} & 32.5\% & 29.4\% & 26.2\% & - & \textbf{35.7\%} & 33.3\% \\
Machine Learning & \textit{out-of-domain} & 38.4\% & 28.6\% & 25.0\% & - & 41.1\% & 41.1\% \\
Mathematics (College) & \textit{out-of-domain} & \textbf{43.0\%} & 33.0\% & 25.0\% & - & 37.0\% & 32.0\% \\
Mathematics (High-School) & \textit{out-of-domain} & \textbf{32.6\%} & 24.4\% & 27.0\% & - & 23.7\% & 31.9\% \\
Medical Genetics & \textit{out-of-domain} & \textbf{70.0\%} & 35.0\% & 36.0\% & - & 69.0\% & 69.0\% \\
Physics (College) & \textit{out-of-domain} & 42.2\% & 21.6\% & 18.6\% & - & 34.3\% & \textbf{46.1\%} \\
Physics (High-School) & \textit{out-of-domain} & 33.8\% & 29.8\% & 25.2\% & - & 33.8\% & \textbf{36.4\%} \\
MedQA-USMLE & \textit{out-of-domain} & 44.4\% & 22.8\% & 23.3\% & - & - & -  \\
MedMCQA Dev & \textit{in-domain} & \textbf{52.9\%} & 29.6\% & 32.5\% & - & - & -  \\
PubMedQA & \textit{in-domain} & \textbf{77.6\%} & 70.2\% & 73.6\% & - & - & -  \\
Statistics (High-School) & \textit{out-of-domain} & 41.2\% & 43.5\% & 19.4\% & - & 50.0\% & \textbf{58.8\%} \\
\bottomrule
\end{tabular}
\end{center}
\caption{\textbf{Question Answering Results}. Galactica is evaluated without few-shot examples. Other LLMs are evaluated 5-shot, except for 0-shot results for GPT-3 on ARC results and OPT and BLOOM on PubMedQA and BioASQ. For abstract algebra and medical genetics, we obtained best results with 30B, so we report these scores; the 120B scores for these were 27.0\% and 68.0\% respectively. Rest of results are for 120B.}
\label{table:full-qa}
\end{table}

\clearpage

\subsection{Citation Prediction}

 In this section we evaluate Galactica's capability to predict citations given an input context, which is an important test of Galactica's capability to organize the scientific literature. We find that both accuracy and the quality of distributional approximation improves with scale.

{\subsubsection{Citation Accuracy}

We construct three datasets to evaluate the model's capability to cite:

\begin{itemize}
    \item \textbf{PWC Citations}: a dataset with 644 pairs of machine learning concepts and papers that introduced them. Concepts consist of methods (e.g. \textit{ResNet}) and datasets (e.g. \textit{ImageNet}) from \textit{Papers with Code}\footnote{\href{https://paperswithcode.com}{https://paperswithcode.com}}. 
    \item \textbf{Extended Citations}: a dataset with 110 pairs of non-machine learning concepts and papers that introduced them. Examples of concepts include \textit{Kozac sequence} and \textit{Breit-Wigner distribution}.
    \item \textbf{Contextual Citations}: a dataset with 1,869 pairs of references and contexts from our arXiv validation set. The dataset is constructed by sampling 1,000 random references and collecting their contexts.
\end{itemize}

For the \textbf{PWC Citations} and \textbf{Extended Citations} datasets, the citation prediction task is framed as a text generation task. The model is given a prompt like "In this paper we use ResNet method \texttt{[START\_REF]}" in order to generate a prediction for the \textit{ResNet} concept. For \textbf{Contextual Citations}, we prompt after the input context for the citation, where the context ends with \verb|[START_REF]|. 

We compare Galactica to sparse and dense retrieval-based approaches on this task. 

For the sparse baseline, we use ElasticSearch to create an index of all the references, including their titles, abstracts, and short snippets of text with the contexts they appear in. Then, given a text query, we retrieve the top references ordered by the sum of matching scores across all selected fields.

For dense retriever baselines, we evaluate two different Contriever models \citep{Contriever}. The first is the pre-trained model released by \citet{Contriever}. The second model we use is fine-tuned on a random subset of 10 million context/paper pairs from our corpus, trained to retrieve the right paper given a context before a citation. The setup for dense retrieval is: (1) each reference is encoded by the model using its title and abstract, (2) a text query is encoded by the same model, (3) the references that match the query re returned. Retrieval is performed using a FAISS index~\citep{FAISS}.

The results can be seen in Table \ref{table:citation-pred}.

\begin{table}[h!]
\begin{center}
\begin{tabular}{ lrrrr } 
\toprule
  Model & Params (bn) & PWC Citations & Extended Citations & Contextual Citations \\ 
\midrule
 GAL 125M & 0.1 & 7.0\% & 6.4\% & 7.1\% \\
 GAL 1.3B & 1.3 & 18.5\% & 45.5\% & 15.9\% \\
 GAL 6.7B & 6.7 & 32.0\% & 60.0\% & 23.0\% \\
 GAL 30B & 30 & 44.7\% & 66.4\% & 31.5\% \\
 GAL 120B & 120 & \textbf{51.9\%} & \textbf{69.1\%} & \textbf{36.6\%} \\
 \midrule
 Sparse Retriever & n/a & 30.9\% & 17.3\% & 5.3\% \\
 Dense Retriever (base) & n/a & 16.4\% & 8.8\% & 1.6\% \\
 Dense Retriever (fine-tuned) & n/a & 27.6\% & 11.8\% & 8.2\% \\
\bottomrule
\end{tabular}
\end{center}
\caption{\textbf{Citation Prediction Accuracy}. Performance of different model sizes on citation prediction.}
\label{table:citation-pred}
\end{table}

The performance on all evaluation sets increases smoothly with scale. At larger scales, Galactica outperforms the retrieval-based approaches as its context-associative power improves. This is an important result as current approaches for navigating the literature use these existing retrieval approaches. As the power of language models improves, we suspect they will become a valuable new tool for exploring the literature.

\subsubsection{Citation Distributional Analysis}

\begin{figure*}[t!]
    \centering
    \begin{subfigure}[t]{0.35\textwidth}
        \centering
        \includegraphics[width=\textwidth]{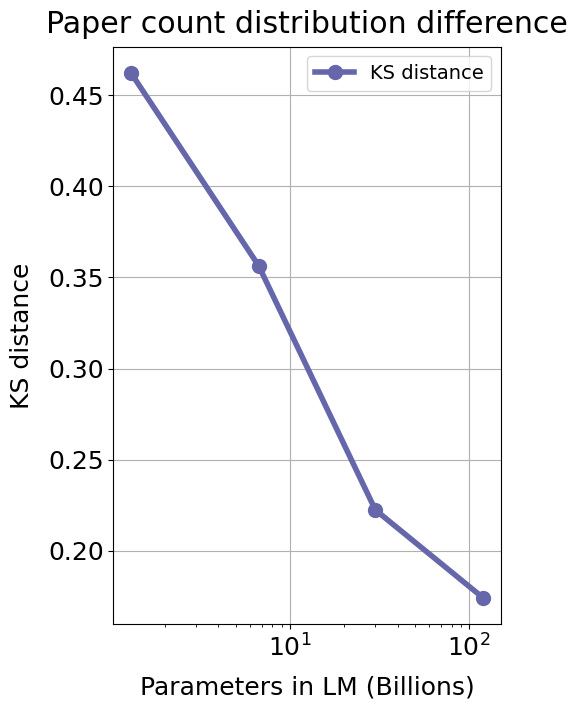}
        \caption{\textbf{Kolmogorov-Smirnov Distance}}
        \label{fig:citation-ks}
    \end{subfigure}%
    ~ \hspace{0.01\textwidth}
    \begin{subfigure}[t]{0.64\textwidth}
        \centering
        \includegraphics[width=\textwidth]{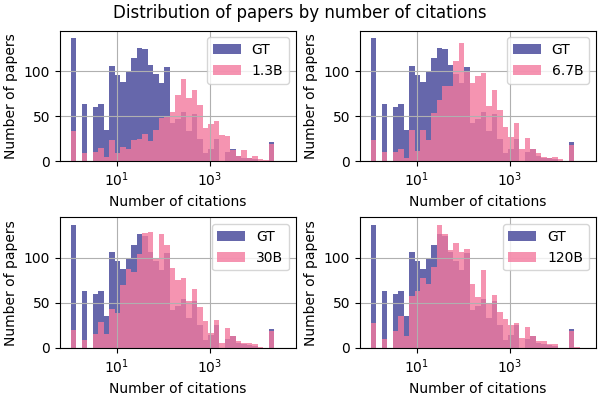}
        \caption{\textbf{Histogram Overlap}}
        \label{fig:citation-dists}
    \end{subfigure}
    \caption{\textbf{Distributional Comparison of Citations}. Galactica's citation distribution approaches the ground truth with scale. This is seen through a declining KS distance with scale, and increasing histogram overlap.}
    \label{fig:dists-comparison}
\end{figure*}

We now turn to look at how well Galactica can model the empirical citation distribution. For this analysis we use the \textbf{Contextual Citations} dataset, where prompts are extracted from a paper by taking the context before a citation as the prompt. An example prompt with a model prediction is shown overleaf in Figure \ref{fig:in_context_pred}.

We use the in-context citation data to analyse the distributional difference between predicted and ground truth paper counts. This allows us to assess the model bias towards predicting more popular papers. Specifically, for each context there is a ground truth and predicted reference. We count the number of times each reference appears in our corpus. We then compare the distribution of reference counts between the ground truth references and the predicted references using the Kolmogorov-Smirnov distance~\citep{Massey_1951}.

The comparison between the citation count distributions for different model sizes can be seen in Figure \ref{fig:dists-comparison}. Figure \ref{fig:citation-ks} shows the decrease in the Kolmogorov-Smirnov distance between the distribution of ground truth paper citations and the distribution of predicted papers citations. Figure \ref{fig:citation-dists} shows how the distribution of paper counts for the predicted papers gets closer to the ground truth as the model size grows. At smaller scales the model is more prone to predicting more popular papers. As the model grows in size this bias towards predicting popular papers diminishes.

\begin{figure}[t!]
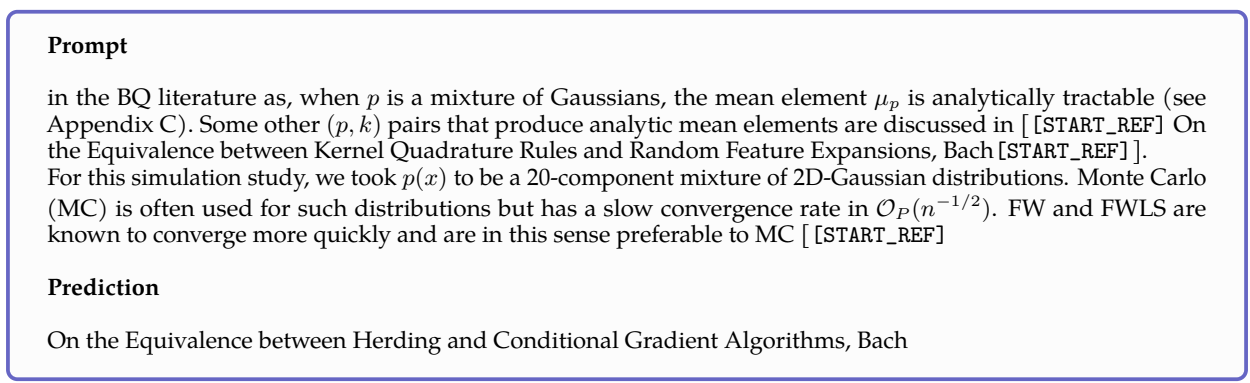

\begin{tcolorbox}[colback=galwhite,colframe=galpurple2]
\begin{small}
\textbf{Prompt} \\

in the BQ literature as, when \(p\) is a mixture of Gaussians, the mean element \(\mu_{p}\) is analytically tractable (see Appendix C). Some other \((p,k)\) pairs that produce analytic mean elements are discussed in [\texttt{[START\_REF]} On the Equivalence between Kernel Quadrature Rules and Random Feature Expansions, Bach\texttt{[START\_REF]}].

For this simulation study, we took \(p(x)\) to be a 20-component mixture of 2D-Gaussian distributions.
Monte Carlo (MC) is often used for such distributions but has a slow convergence rate in \(\mathcal{O}_{P}(n^{-1/2})\).
FW and FWLS are known to converge more quickly and are in this sense preferable to MC [\texttt{[START\_REF]} \\

\textbf{Prediction} \\

On the Equivalence between Herding and Conditional Gradient Algorithms, Bach

\end{small}
\end{tcolorbox}
\caption{\textbf{Citation Prompt}.
An example prompt predicting a citation in-context; from \citet{briol2015frank}.
}
\label{fig:in_context_pred}
\end{figure}

\subsection{General Capabilities}

We have studied Galactica's scientific capabilities. It is perhaps not surprising that a specialist scientific model outperforms general models on scientific tasks, but what would be more surprising was if it outperformed general models on general NLP tasks. In this section, we show surprising evidence that it does just that.

We evaluate on 57 BIG-bench tasks in Table~\ref{table:bigbench-summary}~\citep{BIGBenchakasomanyauthorsitdoesntfitinthecontextwindow}. The tasks are primarily non-scientific and test general language capability, for example anachronisms, figure of speech and metaphor boolean. We always evaluate with 5-shots, and we use the default prompt style from BIG-Bench. Importantly, we do not include this prompt style in pre-training; so the evaluation between Galactica and the other models is comparable 5-shot. Full details and results are in the Appendix. We summarize average scores in Table~\ref{table:bigbench-summary}:

\begin{table}[h!]
\begin{center}
\begin{tabular}{ lrrr } 
\toprule
  Model & Params (bn) & Accuracy & Accuracy \\
    &  & \textit{weighted} & \textit{unweighted} \\
\midrule
OPT 30B & 30 & 39.6\% & 38.0\%  \\ 
BLOOM 176B & 176 & 42.6\% & 42.2\% \\ 
OPT 175B & 175 & 43.4\% & 42.6\% \\ 
GAL 30B & 30 & 46.6\% & 42.7\% \\
GAL 120B & 120 & \textbf{48.7\%} & \textbf{45.3\%} \\
\bottomrule
\end{tabular}
\end{center}
\caption{\textbf{BIG-bench 57 Task Results}. Galactica outperforms general open models at smaller scales.}
\label{table:bigbench-summary}
\end{table}

Both the 30B and 120B Galactica models outperform the larger OPT and BLOOM general models. This is a surprising result given we designed Galactica to trade-off generality for performance in scientific tasks. 

We suspect this result reflects the higher-quality of the Galactica corpus, stemming from the fact it is curated and also primarily academic text. Previous open LLM efforts likely overfocused on scale goals and underfocused on data filtering. Another implication is that the focus on tokens $\rightarrow \infty$ from Chinchilla needs to be complemented with strong data quality procedures~\citep{Chinchilla}. With this paper, we took an opposite approach by focusing on high-quality tokens and repeated epochs of training. However, the Chinchilla insight stands: and there is much more scientific text that we have not exploited in this work.

\subsection{Chemical Understanding}

We now turn to Galactica's capability to interface with different scientific modalities. We start by looking at Galactica's chemical capabilities. Chemical properties exhibit complex correlations which means the chemical space is very large. Better organization of chemical information through language models could aid chemical design and discovery. We explore how Galactica can provide a new interface for these tasks in this section.

For this work, we only include a small subset of available compounds from PubChem Compound in pre-training. Specifically, we take a random subset ($2$ million) of total compounds ($110$ million). This is to ensure the model is not overly biased towards learning natural sequences over natural language. This is a constraint we can relax in future work, enabling for much larger corpus. Here we focus on the first step of investigating whether a single model can learn effectively in the multi-modal setting.

We find that a language model can learn chemical tasks such as IUPAC naming in a self-supervised way, and in addition, we can pose drug discovery tasks as natural language prompts and achieve reasonable results.

\subsubsection{IUPAC Name Prediction}

SMILES is a line notation which represents chemical structure as a sequence of characters~\citep{SMILES}. In the Galactica corpus, the SMILES formula occurs alongside information in the document, such as IUPAC names, molecular weight and XLogP. In the context of self-supervised learning, this means a language model is performing implicit multi-task learning: the model is predicting the next SMILES token, but can also use SMILES to predict other entities in the document. 

As an initial test, we set up a \textbf{IUPAC Name Prediction} task, where the task is to name a compound according to the IUPAC nomenclature given a SMILES formula input. The IUPAC nomenclature is a method of naming organic compounds that has a ruleset based on naming the longest chain of carbons connected by single bonds~\citep{IUPACNaming}. There is a large set of rules and the procedure is algorithmically complex, meaning it is hard to automate. As a result, it is missing from standard cheminformatics toolkits.

Previous works such as STOUT and Struct2IUPAC have explored the possiblity of using RNNs and Transformers for this task~\citep{STOUT,Struct2IUPAC}. We explore in this section whether Galactica can translate a SMILES specification to its IUPAC name in the self-supervised setting. We design a prompt based on the PubChem  structure, with the SMILES as the only input, and the output to predict the IUPAC name.

To evaluate, we use our compound validation set of 17,052 compounds, and prompt with the SMILES formula and predict the IUPAC name. To calculate accuracy, we use OPSIN to convert the generated IUPAC name to SMILES, canonicalize it and compare with the canonicalized SMILES target~\citep{OPSIN}. 

Results are shown in Table~\ref{table:iupac-naming}.

\begin{table}[h!]
\begin{center}
\begin{tabular}{ lrrr } 
\toprule
  Model & Params (bn) & Accuracy & Invalid Names \\ 
\midrule
 GAL 125M & 0.1 & 0.0\% & 32.8\% \\
 GAL 1.3B & 1.3 & 2.5\% & 12.0\% \\ 
 GAL 6.7B & 6.7 & 10.7\% & 12.3\% \\
 GAL 30B & 30 & 15.4\% & 9.7\% \\  
 GAL 120B & 120 & \textbf{39.2\%} & \textbf{9.2\%} \\ 
\bottomrule
\end{tabular}
\end{center}
\caption{\textbf{Results on IUPAC Naming}. Performance improves smoothly with scale.}
\label{table:iupac-naming}
\end{table}

Accuracy increases smoothly with scale. Given we restricted the corpus to 2 million molecules, it is likely much better performance is achievable through training or fine-tuning on more molecules. The model is freely available for those who want to perform this follow-up work.

The more immediate question is what is actually being learnt: is Galactica inferring names from the fundamental molecular structure? To answer this, we visualize the average atomic attention at each stage of a prediction in Figure~\ref{fig:iupac-naming-viz} overleaf. Encouragingly, the results are interpretable in terms of the underlying chemistry, and Galactica attends to the correct group when predicting a name, e.g. for "amino" it attends primarily to the $-\ce{NH_{2}}$ substituent.

\clearpage

\begin{figure}[h]
    \centering

    \textbf{Task: Convert the SMILES to IUPAC Name} \\
    \vspace{8pt}
    Example: \verb|CC(C)(C)C(=O)N(CC1=NC(=CS1)C(=O)OC)C2CCCCC2| \\
    \vspace{16pt}

\begin{tabular}{ccc} 
 Atomic Attention & Predicted So Far & Token Predicted \\
 \midrule
 \begin{tabular}{c}
    \includegraphics[width=0.25\textwidth]{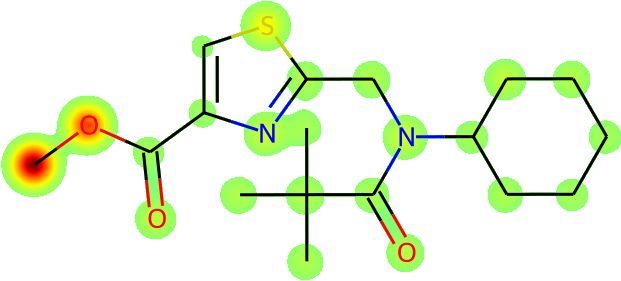}
  \end{tabular}
  & \begin{tabular}{c}
  \small
    -
  \end{tabular} &   \begin{tabular}{c}
    \verb|methyl|
  \end{tabular} \\ 

 \begin{tabular}{c}
    \includegraphics[width=0.25\textwidth]{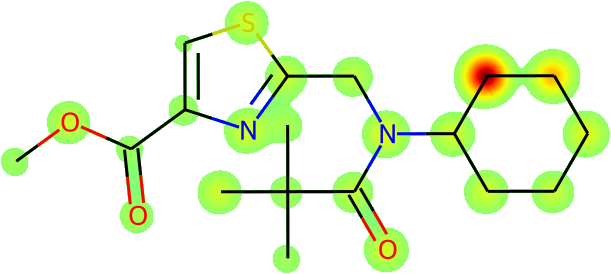}
  \end{tabular}
  & \begin{tabular}{c}
  \small
    methyl 2-[[cyclohexyl
  \end{tabular} &   \begin{tabular}{c}
    \verb|cyclohexyl|
  \end{tabular} \\ 

   \begin{tabular}{c}
    \includegraphics[width=0.25\textwidth]{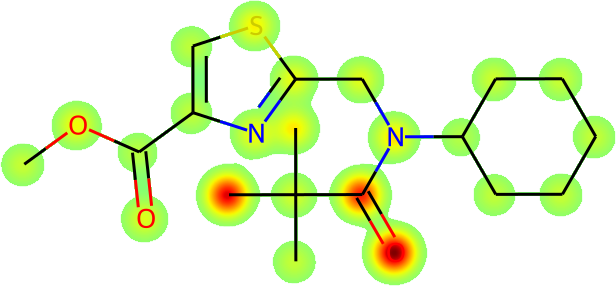}
  \end{tabular}
  & \begin{tabular}{c}
  \small
    methyl 2-[[cyclohexyl-(2,2-
  \end{tabular} &   \begin{tabular}{c}
    \verb|dimethyl|
  \end{tabular} \\

     \begin{tabular}{c}
    \includegraphics[width=0.25\textwidth]{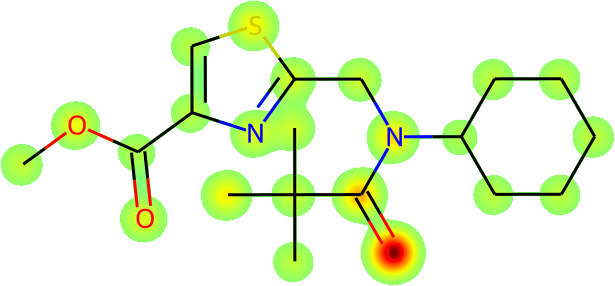}
  \end{tabular}
  & \begin{tabular}{c}
  \small
    methyl 2-[[cyclohexyl-(2,2-dimethyl
  \end{tabular} &   \begin{tabular}{c}
    \verb|prop|
  \end{tabular} \\

   \begin{tabular}{c}
    \includegraphics[width=0.25\textwidth]{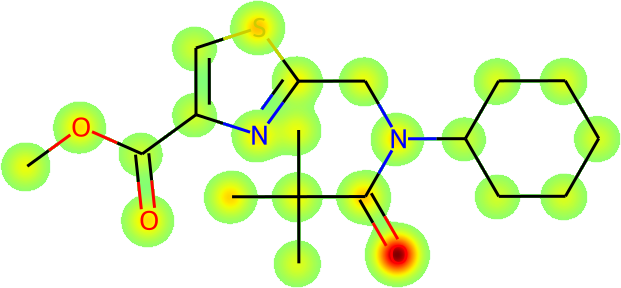}
  \end{tabular}
  & \begin{tabular}{c}
  \small
    methyl 2-[[cyclohexyl-(2,2-dimethylprop
  \end{tabular} &   \begin{tabular}{c}
    \verb|anoyl|
  \end{tabular} \\

   \begin{tabular}{c}
    \includegraphics[width=0.25\textwidth]{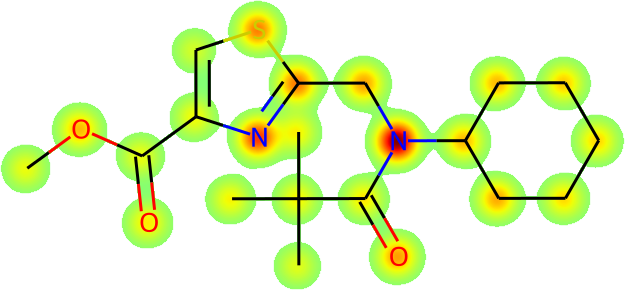}
  \end{tabular}
  & \begin{tabular}{c}
  \small
    methyl 2-[[cyclohexyl-(2,2-dimethylpropanoyl)
  \end{tabular} &   \begin{tabular}{c}
    \verb|amino|
  \end{tabular} \\ 

   \begin{tabular}{c}
    \includegraphics[width=0.25\textwidth]{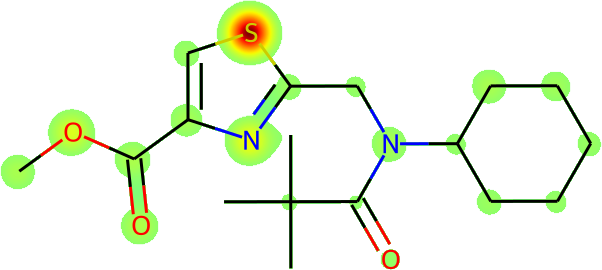}
  \end{tabular}
  & \begin{tabular}{c}
  \small
    methyl 2-[[cyclohexyl-(2,2-dimethylpropanoyl)]amino] \\
    \small methyl]
  \end{tabular} &   \begin{tabular}{c}
    \verb|th|
  \end{tabular} \\ 

   \begin{tabular}{c}
    \includegraphics[width=0.25\textwidth]{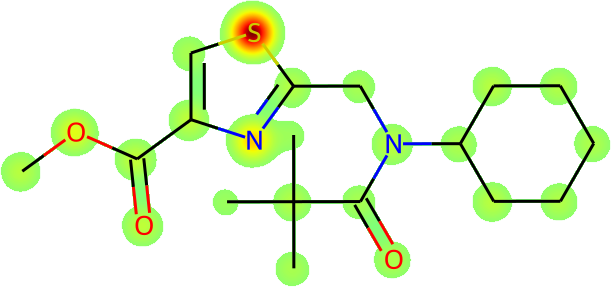}
  \end{tabular}
  & \begin{tabular}{c}
  \small
    methyl 2-[[cyclohexyl-(2,2-dimethylpropanoyl)]amino] \\
    \small methyl]th
  \end{tabular} &   \begin{tabular}{c}
    \verb|iazole|
  \end{tabular} \\

   \begin{tabular}{c}
    \includegraphics[width=0.25\textwidth]{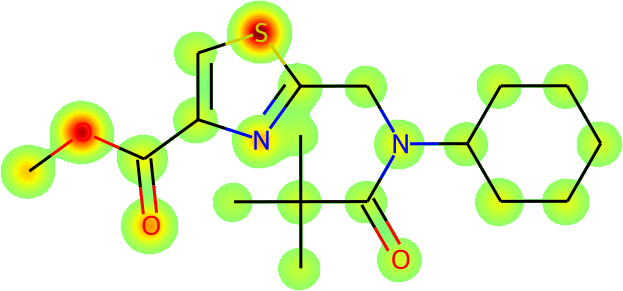}
  \end{tabular}
  & \begin{tabular}{c}
  \small
    methyl 2-[[cyclohexyl-(2,2-dimethylpropanoyl)]amino] \\ \small methyl]thiazole-4-
  \end{tabular} &   \begin{tabular}{c}
    \verb|carboxylate|
  \end{tabular} \\ 

   \hline
\end{tabular}

    \caption{\textbf{Attending to Functional Groups}. Galactica uses its knowledge of chemistry to help with the IUPAC Naming task. At each stage of prediction, it attends to the part of the molecular graph associated with the group name, e.g. for "amino" it attends to the nitrogen atom; for thiazole, the sulphur atom.}
    \label{fig:iupac-naming-viz}
\end{figure}

\clearpage

\subsubsection{MoleculeNet}

We now explore whether we can pose traditional drug discovery tasks in a natural language format, combining the different modalities involved. Humans organize knowledge via natural language, and so learning an interface between natural language and scientific modalities like SMILES could be a new tool for navigating the chemical space. We use MoleculeNet classification benchmarks to answer this question, which are summarized in Table~\ref{table:moleculenet}~\citep{MoleculeNet}.

\begin{table}[h!]
\begin{center}
\begin{tabular}{ lllll } 
\toprule
  Category & Dataset & Type & Other modalities \\ 
\midrule
 \multirow{ 3}{*}{Biophysics} & HIV & Classification & n/a \\
& BACE C & Classification & n/a \\
\midrule
 \multirow{ 4}{*}{Physiology} & BBBP & Classification & n/a \\
& Tox21 & Classification & protein sequences \\
& SIDER & Classification & n/a \\
& ClinTox & Classification & n/a \\
\bottomrule
\end{tabular}
\end{center}
\caption{\textbf{MoleculeNet datasets used for evaluation}. We convert training sets to text format and include in pre-training. We evaluate using the splits suggested by the DeepChem library~\citep{Ramsundaretal}.}
\label{table:moleculenet}
\end{table}

To evaluate, we include the training sets in pre-training by converting to a text format. We use prompt randomization (varying how the question is posed). For example, for BBBP the training prompt has forms like in Figure~\ref{fig:bbbp_prompt} below. These examples occur alongside the other corpuses in training, and each example is seen just over $4$ times. This is not comparable to \textit{direct} fine-tuning or supervision due to the presence of other data in pre-training, so it might be considered a form of weak supervision instead.

\begin{figure}[h]
\begin{tcolorbox}[colback=galwhite,colframe=galpurple2]
\begin{small}
Here is a SMILES formula: \\

\verb|[START_I_SMILES]O=C(O)CCCC1=CC=C(N(CCCl)CCCl)C=C1[END_I_SMILES]| \\

\textbf{Question:}  Will the chemical compound penetrate the blood-brain barrier? \\

\textbf{Answer:} No
\end{small}
\end{tcolorbox}
\caption{\textbf{BBBP Prompt}.
We include the SMILES and pose the classification problem in natural language.
}
\label{fig:bbbp_prompt}
\end{figure}

For some MoleculeNet datasets, other modalities are implicitly present. For example, in the Tox21 dataset, bioassays concern particular receptors such as the androgen receptor (AR). As an experiment, we decided to frame the task in a text format with the protein sequence and the SMILES as part of the prompt. We show an example for Tox21 in Figure~\ref{fig:tox21_prompt}.

\begin{figure}[h]
\begin{tcolorbox}[colback=galwhite,colframe=galpurple2]
\begin{small}
Here is a sequence for a protein: \\

\verb|[START_AMINO]MEEPQSDPSVEPPLSQETFSDLWKLLPE...[END_AMINO]| \\
 
And here is an isomeric SMILES for a compound: \\

\verb|[START_I_SMILES]CC(O)(P(=O)(O)O)P(=O)(O)O[END_I_SMILES]| \\

\textbf{Question:} Will the the chemical compound be active against this protein? \\

\textbf{Answer:} No
\end{small}
\end{tcolorbox}
\caption{\textbf{Tox21 Prompt}.
We include the protein sequence and the SMILES formula and pose the classification problem in natural language.
}
\label{fig:tox21_prompt}
\end{figure}

We make sure to Kekulize the SMILES to be consistent with PubChem representations. For evaluation, we use the recommended splits from the DeepChem library~\citep{Ramsundaretal}. 

We present results in Table~\ref{table:moleculenet-classification}. Performance scales with model size. The scaling is slower than tasks like QA, and the base model lags a specialist model with explicit 3D information and 10 times more molecules ~\citep{UniMol}. We suspect the weak supervision setup is harder for this task, and fine-tuning and/or more molecule data is required to get sufficient task signal. The model is available for work on this. 

\begin{table}[h!]
\begin{center}
\begin{tabular}{ lccrrrrrrrr } 
\toprule
  \multicolumn{10}{c}{MoleculeNet Classification}      \\ 
\midrule
  Model & Modality & Molecules & BACE & BBBP & ClinTox & HIV & SIDER & Tox21 & Av.   \\ 
 \midrule
 GAL 125M & SMILES & 2M & 0.561 & 0.393 & 0.518 & 0.702 & 0.559 & 0.543 & 0.581 \\
 GAL 1.3B & SMILES & 2M & 0.576 & 0.604 & 0.589 & 0.724 & 0.540 & 0.606 & 0.619 \\
 GAL 6.7B & SMILES & 2M & 0.584 & 0.535 & 0.784 & 0.722 & 0.559 & 0.639 & 0.640 \\
 GAL 30B & SMILES & 2M & 0.727 & 0.596 & 0.822 & 0.759 & 0.613 & 0.685 & 0.687 \\
GAL 120B & SMILES & 2M & 0.617 & 0.661 & 0.826 & 0.745 & 0.632 & 0.689 & 0.690  \\
\midrule
\textit{Uni-Mol} & 3D & 20M & 0.857 & 0.729 & 0.919 & 0.808 & 0.659 & 0.796 & 0.770  \\
\bottomrule
\end{tabular}
\end{center}
\caption{\textbf{Results on MoleculeNet Classification}. Results are scored by ROC-AUC.}
\label{table:moleculenet-classification}
\end{table}

For our purposes, the implication for future work is that we can learn drug discovery tasks via natural language prompts. If we can learn these relationships automatically in a signal-dense document context (e.g. online chemical databases), this might reduce the reliance on supervised datasets to perform these tasks.

As a final check, we can average Galactica's attention heads across layers, and visualize whereabouts the model looks in the SMILES sequence to make a prediction (atomic attention). We show an example in Figure~\ref{fig:moleculenet-viz} for some Tox21 predictions.

\begin{figure}
     \centering
     \textbf{Positive Examples} \\
    \begin{subfigure}[b]{0.29\textwidth}
         \centering
         \includegraphics[width=\textwidth]{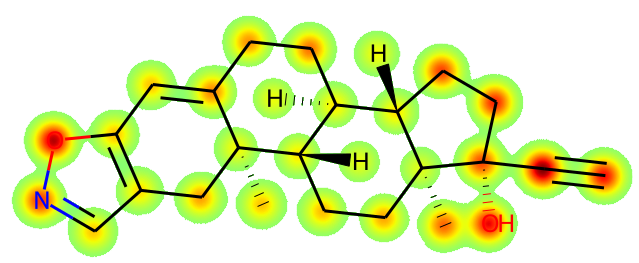}
         \caption{Danazol (28417) on NR-AR}
         \label{fig:three sin x}
     \end{subfigure}
     \hfill
     \begin{subfigure}[b]{0.3\textwidth}
         \centering
         \includegraphics[width=\textwidth]{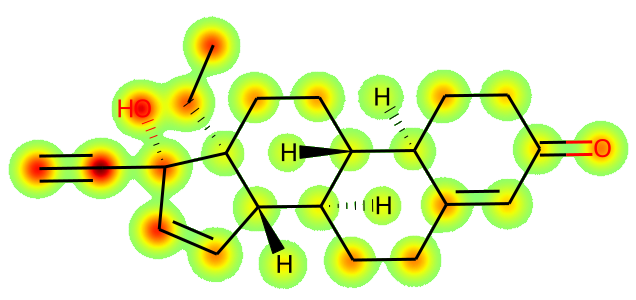}
         \tiny
         \caption{Gestodene (3033968) on NR-AR}
         \label{fig:gestodene}
     \end{subfigure}
     \hfill
     \begin{subfigure}[b]{0.32\textwidth}
         \centering
         \includegraphics[width=\textwidth]{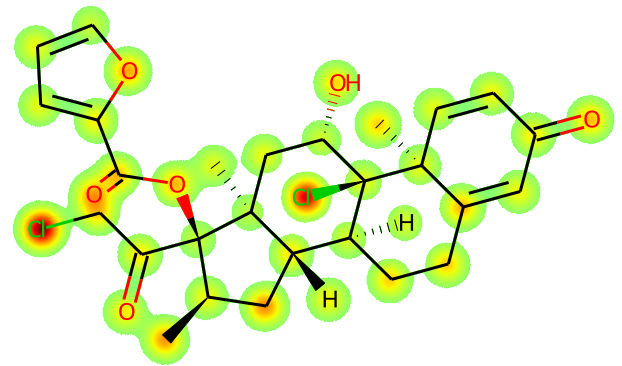}
         \caption{Mometasone f. (441336) on NR-AR}
         \label{fig:five over x}
     \end{subfigure}

     \par\bigskip
     
    \textbf{Negative Examples}
    \par\bigskip

     \begin{subfigure}[b]{0.32\textwidth}
         \centering
         \includegraphics[width=\textwidth]{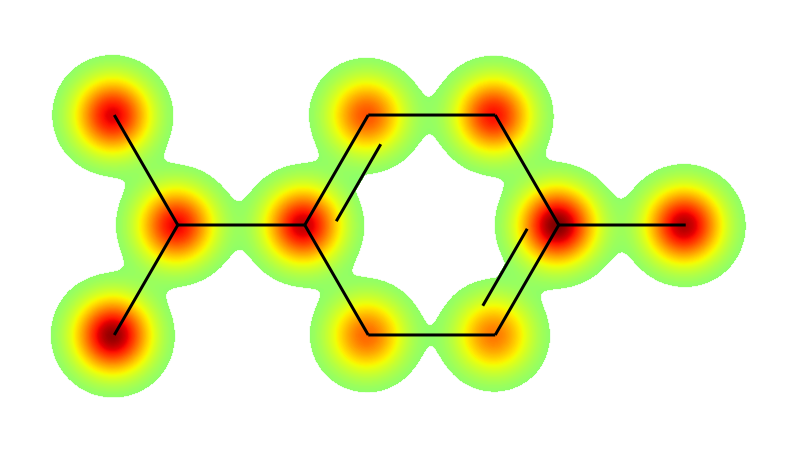}
         \tiny
         \caption{$\gamma$-Terpinene (7461) on NR-PPAR-$\gamma$}
         \label{fig:gestodene}
     \end{subfigure}
     \hfill
     \begin{subfigure}[b]{0.29\textwidth}
         \centering
         \includegraphics[width=\textwidth]{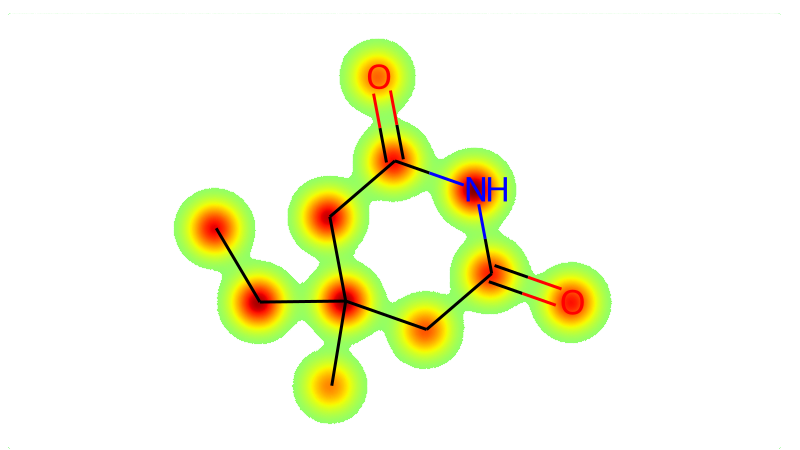}
         \caption{Bemegride (2310) on NR-AR}
         \label{fig:three sin x}
     \end{subfigure}
     \hfill
     \begin{subfigure}[b]{0.3\textwidth}
         \centering
         \includegraphics[width=\textwidth]{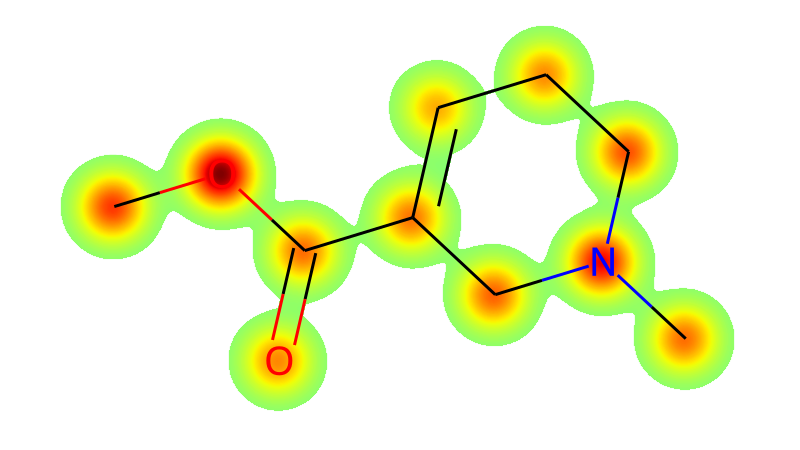}
         \caption{Arecoline (2230) on NR-PPAR-$\gamma$}
         \label{fig:five over x}
     \end{subfigure}

     \par\bigskip

\caption{\textbf{Attention Visualization on Tox21}. The top three molecules are highest confidence positive examples for the 30B model; the bottom three are the highest confidence negatives. We match attention weights from the SMILES with the canonical atom ordering. Danazol and gestodene are known to possess high affinities for the androgen receptor (AR)~\citep{andrology}.}
\label{fig:moleculenet-viz}
\end{figure}

\clearpage

\subsection{Biological Understanding}

In this section we examine Galactica's capability to interface with biological modalities. Language models could potentially play a role in automatic organisation of this data, for example annotating newly sequenced proteins with functional information. We explore the potential of this interface in this section.

For protein sequences from UniProt, we include a small subset of available sequences in pre-training. Specifically,  we take reviewed Swiss-Prot proteins; a high-quality subset ($0.5$ million) of total ($227$ million). This is to ensure the model is not overly biased towards learning natural sequences over natural language. As with molecule data, this is a constraint we can relax in future work, enabling for much larger corpus. Here we focus on the first step of investigating whether a single model can learn effectively in the multi-modal setting.

We find that a language model can learn an implicit measure of sequence similarity that it can use for tasks such as functional annotation and descriptions.

\subsubsection{Sequence Validation Perplexity}

\begin{figure}[t!]
\centering
\includegraphics[width=\textwidth]{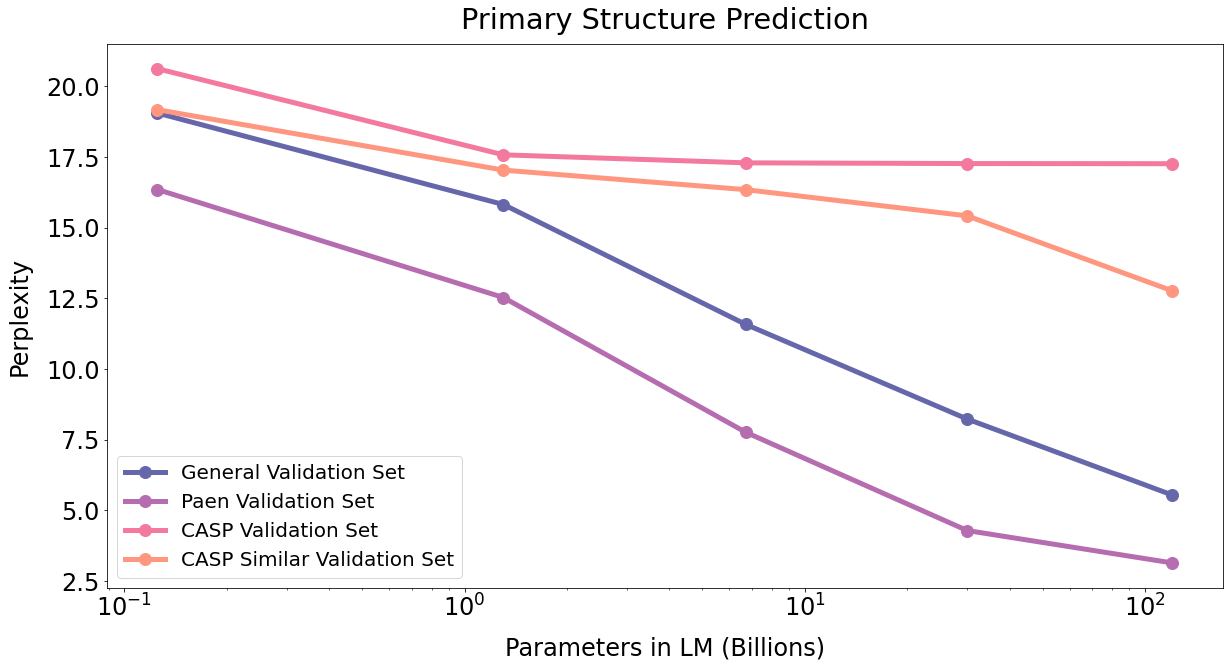}
\caption{\textbf{Primary Structure Prediction}. For three of the validation sets we observe smooth scaling, reflecting the potential for high sequence similarity with sequences in the training set; for example, orthologs in the case of the Paen validation set. The CASP set with sequence similarity constraints levels off, suggesting the gains from the 550k proteins in training quickly saturates for more out-of-domain sequences.}
\label{fig:protein_val_graph}
\end{figure}

While Galactica does not explicitly model the 3D structure of a protein, the information needed for a specific conformation is contained in the linear amino acid sequence, which in turn determine function. As a first step, we test upstream performance through evaluating protein sequence perplexity. Constructing a good validation set is important and data leakage is a problem for works in this field. We construct four holdout sets to obtain more confidence about what is being learnt and what generalizes.

First, we conduct BLAST on the sequences in the training set and remove all sequences with a sequence identity \(\geq 50\%\) with 51 CASP14 target sequences. These are the same test sequences used in ESMFold~\citep{ESMFold}. In total we remove 167 sequences from the training set using this approach. We call this this holdout set \textbf{CASPSimilarSeq}. We call the 51 CASP14 target sequences \textbf{CASPSeq}.

Secondly, we conduct organism-level holdout, and remove all sequences from the Paenungulata clade of organisms, including elephants, elephant shrews, manatees and aadvarks. This allows us to test whether Galactica can annotate sequeces for organisms it has never seen before. In total we remove 109 sequences from the training set using this approach. We call this holdout set \textbf{PaenSeq}. Note that this does not enforce any sequence similarity constraints, and there may be very similar sequences in the training set.

Lastly, we conduct a randomized test split, consisting of 5456 sequences. There is no sequence identity constraint applied, so memorization may be more at play, but it still provides a signal about the breadth of sequence knowledge absorbed by the model. We call this holdout set \textbf{UniProtSeq}.

We evaluate perplexity for all holdout sets in Table~\ref{table:protein-val} and plot in Figure~\ref{fig:protein_val_graph}. For three of the validation sets we observe smooth scaling, reflecting the potential for high sequence similarity with sequences in the training set; for example, orthologs in the case of the Paen validation set. Interestingly, the CASP set with sequence similarity constraints levels off, suggesting the gains from the 550k proteins in training quickly saturates. 

\begin{table}[h!]
\begin{center}
\begin{tabular}{ lrrrrr } 
\toprule
  \multicolumn{6}{c}{Protein Sequence Validation Perplexity}      \\ 
\midrule
  Model & Param (bn) & CASPSeq & CASPSimSeq & PaenSeq & UniProtSeq \\ 
\midrule
 GAL 125M & 0.1 & 20.62 & 19.18 & 16.35 & 19.05 \\
 GAL 1.3B & 1.3 & 17.58 & 17.04 & 12.53 & 15.82 \\
 GAL 6.7B & 6.7 & 17.29 & 16.35 & 7.76 & 11.58 \\
 GAL 30B & 30 & 17.27 & 15.42 & 4.28 & 8.23 \\
 GAL 120B & 120 & \textbf{17.26} & \textbf{12.77} & \textbf{3.14} & \textbf{5.54} \\
\bottomrule
\end{tabular}
\end{center}
\caption{\textbf{Protein Validation Perplexity}. Validation sets with higher potential sequence similarity with the training set have lower perplexity than the restricted sets (CASP validation sets).}
\label{table:protein-val}
\end{table}

To investigate further, we example validation perplexity on the \textbf{CASPSeq} set during training of the 120B model, and we plot results in Figure~\ref{fig:casp_intermed} below.

\begin{figure}[h]
\centering
\includegraphics[width=\textwidth]{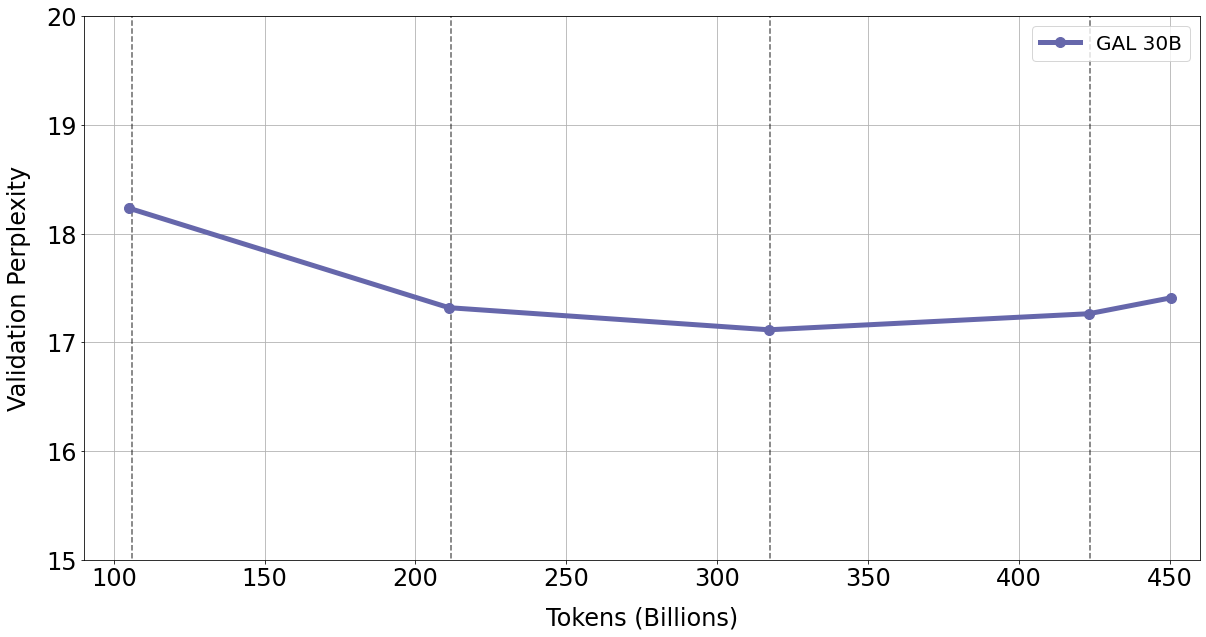}
\caption{\textbf{CASPSeq Validation During Training}. Overfitting occurs before the end of training, but the effect is not drastic, and repeating the protein sequences three times does not damage performance on this task. The final 120B model is the second-last point, reflecting the early stopping we applied (see earlier Sections)}
\label{fig:casp_intermed}
\end{figure}

We observe falling validation perplexity up until the start of the fourth epoch, at which point the model overfits for this particular dataset. This may suggest Galactica is getting worse at more "out-of-domain" proteins that differ significantly from the test set. For future work, less repetition is probably desirable; and more generally, increasing the diversity of proteins in the training dataset is likely to be beneficial.

\clearpage

\subsubsection{Functional Keyword Prediction}

We now look at specific translation capabilities from protein sequence toward natural language, which may be useful for tasks such as protein annotation. As a first test, we look at UniProt keywords that Galactica can infer from the sequence. An example of these is shown in Figure~\ref{fig:protein_keywords_example} overleaf.

\begin{figure}[t!]
\centering
\includegraphics[width=\textwidth]{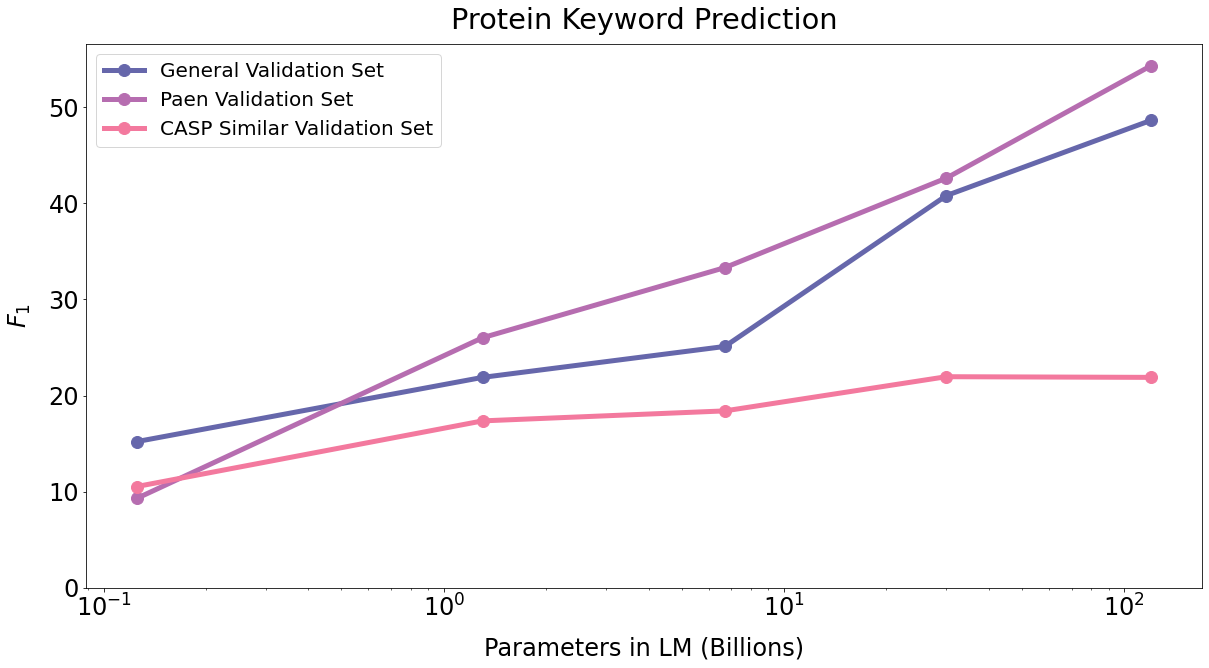}
\caption{\textbf{Protein Keyword Prediction}. This test's Galactica's capability to predict protein keywords, e.g. "cytoplasm", from the sequence alone. For the Paen and General datasets, this capability improves smoothly with scale. It scales more slowly and begins to saturate for the CASPSimSeq set, reflecting the lower sequence similarity with sequences in the training set.}
\end{figure}

\begin{figure}[h]
\begin{tcolorbox}[colback=galwhite,colframe=galpurple2]
\begin{small}
\textbf{\#\# Sequence} \\

Here is the sequence: \\

\verb|[START_AMINO]MQKSPLERASVISKLFFSWPGPILRKGYRQHLKLSDIYQIPSVDSADNLSEKLERE...[END_AMINO]| \\

\textbf{\#\#\# Ground-Truth Keywords} \\

ATP-binding, Cell membrane, Chloride, Chloride channel, Endoplasmic reticulum, Endosome, Glycoprotein, Ion channel, Ion transport, Isomerase, Isopeptide bond, Lipoprotein, Membrane, Nucleotide-binding, Nucleus, Palmitate, Phosphoprotein, Reference proteome, Repeat, Transmembrane, Transmembrane helix, Transport, Ubl conjugation \\

\textbf{\#\#\# Galactica 30B Predicted Keywords} \\

ATP-binding, Cell membrane, Chloride, Chloride channel, Endoplasmic reticulum, Endosome, Glycoprotein, Ion channel, Ion transport, Isomerase, Isopeptide bond, Lipoprotein, Membrane, Nucleotide-binding, Nucleus, Palmitate, Phosphoprotein, Reference proteome, Repeat, Transmembrane, Transmembrane helix, Transport, Ubl conjugation
\end{small}
\end{tcolorbox}
\caption{\textbf{Protein Keyword Prediction}. Example shown is Q108U0 from the PaenSeq holdout, a cystic fibrosis transmembrane conductance regulator from the African elephant. The closest protein by sequence similarity in the training set is the Q2QLA3 protein, a cystic fibrosis transmembrane conductance regular from a horse, with 91.8\% sequence similarity.}
\label{fig:protein_keywords_example}
\end{figure}

We report results in Table~\ref{table:protein-keyword}. $F_{1}$ score increases across the holdout sets with scale, suggesting that Galactica can learn keywords by inferring from the sequence. However, we see saturation for the CASPSimSeq, suggesting this capability depends on how similar the sequences are to those in the training set. This is reflected in the example in Figure~\ref{fig:protein_keywords_example}, where Galactica uses its knowledge of a similar proteins from different organisms, with a maximum sequence similarity of 91.8\% in the training set, to help annotate.

\begin{table}[h!]
\begin{center}
\begin{tabular}{ lrrrr } 
\toprule
  \multicolumn{5}{c}{Protein Keyword Prediction}      \\ 
\midrule
  Model & Param (bn) & CASPSimSeq & PaenSeq & UniProtSeq \\ 
\midrule
 GAL 125M & 0.1 & 10.5\% & 9.3\% & 15.2\% \\
 GAL 1.3B & 1.3 & 17.4\% & 26.0\% & 21.9\% \\
 GAL 6.7B & 6.7 & 18.4\% & 33.3\% & 25.1\% \\
 GAL 30B & 30 & \textbf{22.0\%} & 42.6\% & 40.8\% \\
 GAL 120B & 120 & 21.9\% & \textbf{54.5\%} & \textbf{48.7\%} \\
\bottomrule
\end{tabular}
\end{center}
\caption{\textbf{Protein Keyword Prediction}. Metric shown is $F_{1}$ score. Performance increases with scale across the holdout sets. Note we do not include CASPSeq as these do not have UniProt keywords we can test against.}
\label{table:protein-keyword}
\end{table}

We attempted to visualize attention in the protein sequence, but we did not observe anything with biological intepretation (e.g. attention to domains). Our working hypothesis is that Galactica has learnt an implicit measure of sequence similarity that it uses to associate predicted keywords, but that this is not directly interpretable from where it attends to. This differs from our chemistry analysis where results were interpretable in terms of attention to the underlying atomic structure.

\subsubsection{Protein Function Description}

As the next test, we look at generating free-form descriptions of protein function from the sequence. We look at the UniProt function descriptions and compare to Galactica generated descriptions.

We report results in Table~\ref{table:protein-function}. ROUGE-L score increases smoothly across all the holdout sets. We show an example overleaf in Figure~\ref{fig:protein_function_example} from PaenSeq. The protein is a Cytochrome b protein from a rock hyrax (Q7Y8J5). The closest sequence by similarity in the training set is a Cytochrome b protein from a pygmy hippopotamus (O03363) with 83\% sequence similarity. In this case we get a perfect prediction from the description.

\begin{table}[h!]
\begin{center}
\begin{tabular}{ lrrrr } 
\toprule
  \multicolumn{5}{c}{Protein Function Prediction}      \\ 
\midrule
  Model & Param (bn) & CASPSimSeq & PaenSeq & UniProtSeq \\ 
\midrule
 GAL 125M & 0.1 & 0.062 & 0.073 & 0.061 \\
 GAL 1.3B & 1.3 & 0.069 & 0.084 & 0.079 \\
 GAL 6.7B & 6.7 & 0.109 & 0.137 & 0.111\\
 GAL 30B & 30 & 0.137 & 0.196 & 0.186 \\
 GAL 120B & 120 & \textbf{0.252} & \textbf{0.272} & \textbf{0.252} \\
\bottomrule
\end{tabular}
\end{center}
\caption{\textbf{Protein Function Prediction}. Metric shown is ROUGE-L. Performance increases with scale.}
\label{table:protein-function}
\end{table}

\begin{figure}[h]
\begin{tcolorbox}[colback=galwhite,colframe=galpurple2]
\begin{small}
This is the sequence: \\

\verb|[START_AMINO]MTNIRKNHPLLKTINDAFIDLPTPSNISTWWNFGSLLGACLIIQVLTGLFLAMHYTSDT...[END_AMINO]| \\

\textbf{\#\#\# Ground-Truth Description} \\

Component of the ubiquinol-cytochrome c reductase complex (complex III or cytochrome b-c1 complex) that is part of the mitochondrial respiratory chain. The b-c1 complex mediates electron transfer from ubiquinol to cytochrome c. Contributes to the generation of a proton gradient across the mitochondrial membrane that is then used for ATP synthesis. \\

\textbf{\#\#\# Galactica 120B Predicted Description} \\

Component of the ubiquinol-cytochrome c reductase complex (complex III or cytochrome b-c1 complex) that is part of the mitochondrial respiratory chain. The b-c1 complex mediates electron transfer from ubiquinol to cytochrome c. Contributes to the generation of a proton gradient across the mitochondrial membrane that is then used for ATP synthesis.
\end{small}
\end{tcolorbox}
\caption{\textbf{Protein Description Prediction}. Example shown is Q7Y8J5 from the PaenSeq holdout, a Cytochrome b protein from a rock hyrax. The closest protein by sequence similarity in the training set is the O03363 protein, a Cytochrome b protein from a pygmy hippopotamus, with 83\% sequence similarity.}
\label{fig:protein_function_example}
\end{figure}

As with the keyword prediction task, Galactica appears to be learning based on matching sequences with similar ones it has seen in training, and using this to form a description. This suggests language models for protein sequences could serve as useful alternatives to existing search methods such as BLAST and MMseqs2~\citep{BLAST, MMseq2}.

\section{Toxicity and Bias}

In this section we study the toxicity and bias of the Galactica model. We evaluate on benchmarks related to stereotypes, toxicity, and misinformation. We compare results to other language models. We find Galactica is significantly less biased and toxic than existing language models.

\subsection{Bias and Stereotypes}

For the following evaluations, we investigate Galactica’s ability to detect (and generate) harmful stereotypes and hate speech, using four widely used benchmarks. 

\subsubsection{CrowS-Pairs}

\begin{table}[h!]
\begin{center}
\begin{tabular}{ lrrrr } 
\toprule
  \multicolumn{4}{c}{CrowS-Pairs}      \\ 
\midrule
Bias type & \verb|text-davinci-002| & OPT 175B & Galactica 120B \\
\midrule
Race                & 64.7          & 68.6    & \textbf{59.9}        \\
Socioeconomic       & 73.8          & 76.2    & \textbf{65.7}        \\
Gender              & 62.6          & 65.7    & \textbf{51.9}        \\
Disability          & 76.7          & 76.7    & \textbf{66.7}        \\
Nationality         & 61.6          & 62.9    & \textbf{51.6}        \\
Sexual-orientation  & \textbf{76.2} & 78.6    & 77.4       \\
Physical-appearance & 74.6          & 76.2    & \textbf{58.7}        \\
Religion            & 73.3          & 68.6    & \textbf{67.6}        \\
Age                 & \textbf{64.4}          & 67.8    & 69.0        \\
Overall             & 67.2          & 69.5    & \textbf{60.5}       \\
\bottomrule
\end{tabular}
\end{center}
\caption{\textbf{CrowS-Pairs Results}. Galactica demonstrates significantly lower stereotypical bias in all categories with the exception of sexual orientation and age.}
\label{table:crows}
\end{table}

CrowS-Pairs is a collection of 1,508 crowd-sourced pairs of sentences, one which is "more" stereotyping and one which is "less" stereotyping, and covers nine characteristics ~\citep{crows}. These characteristics are race, religion, socioeconomic status, age, disability, nationality, sexual orientation, physical appearance, and gender. A language model’s preference for stereotypical content is measured by computing the proportion of examples in which the "more" stereotypical sentence is preferred (as determined by log likelihood). Higher scores indicate a more harmfully biased model, whereas an ideal model with no bias would score 50\%.

We report results for Galactica and other language models in Table \ref{table:crows}. Galactica exhibits significantly lower stereotypical biases in most categories, with the exception of sexual orientation and age, when compared to the latest GPT-3 (\verb|text-davinci-002|) and OPT 175B. Galactica attains a better overall score of 60.5\% compared to the other models. Language models such as OPT use the Pushshift.io Reddit corpus as a primary data source, which likely leads the model to learn more discriminatory associations~\citep{OPT}. Galactica is trained on a scientific corpus where the incidence rate for stereotypes and discriminatory text is likely to be lower.

\subsubsection{StereoSet}

\begin{table}[h!]
\begin{center}
\begin{tabular}{ lcrrr } 
\toprule
  \multicolumn{5}{c}{StereoSet}      \\ 
\midrule
Category & & \verb|text-davinci-002| & OPT 175B & Galactica 120B \\
\midrule

    & LMS (\(\uparrow\)) & 78.4 & 74.1 & 75.2 \\ Prof. & SS (\(\downarrow\)) & 63.4 & 62.6 & 57.2 \\  & ICAT (\(\uparrow\)) & 57.5 & 55.4 & \textbf{64.3} \\ 
    \hline  
    & LMS (\(\uparrow\)) & 75.6 & 74.0 & 74.6 \\ Gend. & SS (\(\downarrow\)) & 66.5 & 63.6 & 59.1 \\  & ICAT (\(\uparrow\)) & 50.6 & 53.8 & \textbf{61.0} \\ 
    \hline  
    & LMS (\(\uparrow\)) & 80.8 & 84.0 & 81.4 \\ Reli. & SS (\(\downarrow\)) & 59.0 & 59.0 & 55.1 \\  & ICAT (\(\uparrow\)) & 66.3 & 68.9 & \textbf{73.1} \\ 
    \hline  & LMS (\(\uparrow\)) & 77.0 & 74.9 & 74.5 \\ Race & SS (\(\downarrow\)) & 57.4 & 56.8 & 54.8 \\  & ICAT (\(\uparrow\)) & 65.7 & 64.8 & \textbf{67.3} \\ 
    \hline  & LMS (\(\uparrow\)) & 77.6 & 74.8 & 75.0 \\ Overall & SS (\(\downarrow\)) & 60.8 & 59.9 & 56.2 \\  & ICAT (\(\uparrow\)) & 60.8 & 60.0 & \textbf{65.6} \\ 

\bottomrule
\end{tabular}
\end{center}
\caption{\textbf{StereoSet Results}. Galactica outperforms all models across all categories on the ICAT score.}
\label{table:stereoset}
\end{table}

StereoSet aims to measure stereotypical biases across profession, religion, gender, and race~\citep{stereoset}. The benchmark contains two tasks: an intrasentence task and an intersentence task, with around 2,100 examples each in the development set.

\begin{itemize}
    \item \textbf{Intrasentence Task}: the stereotype and associated context are in the same sentence.
    \item \textbf{Intersentence Task}: the context and stereotype are in different (consecutive) sentences.
\end{itemize}

Alongside stereo- and anti-stereotypical variants of sentences, each example in StereoSet contains an unrelated sentence. This sentence is included for measuring a Language Modelling Score (LMS) and a Stereotype Score (SS). These two metrics are combined to form the Idealized Context Association Test score (ICAT), which is a balanced measure of bias detection and language modeling. An ideal, unbiased language model would score an LMS of 100, an SS of 50, and an ICAT of 100.

We report results in Table~\ref{table:stereoset}. Galactica outperforms other models on all categories for the overall ICAT score.

\subsubsection{Toxicity}

\begin{figure}[t!]
    \centering
    \includegraphics[width=1.0\textwidth]{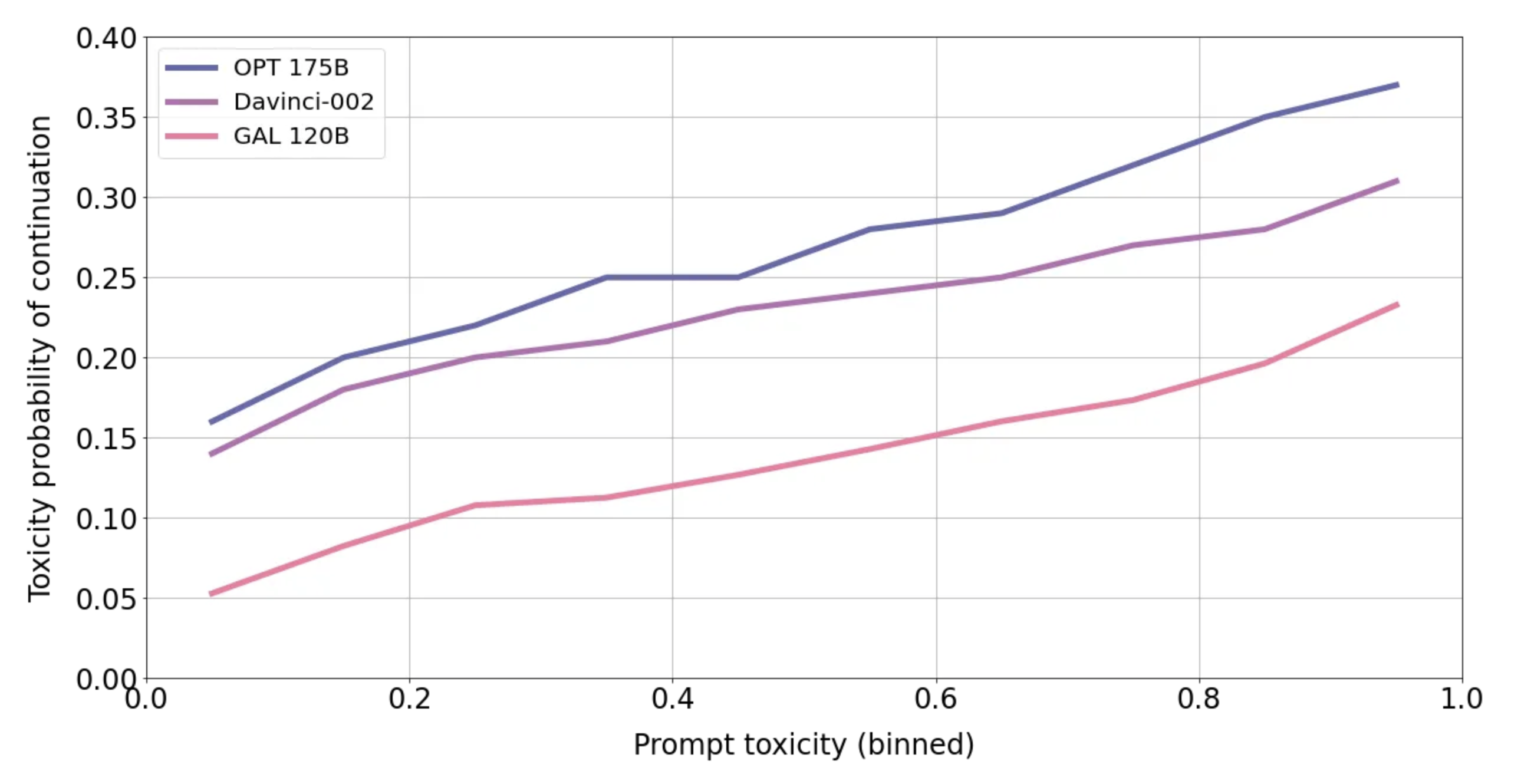}
    \caption{\textbf{Toxicity rate on RealToxicityPrompts}. Galactica exhibits much lower toxicity continuation rates, even as we increase the original prompt toxicity.}
    \label{fig:tox}
\end{figure}

To measure toxicity we use the RealToxicityPrompts (RTP) benchmark introduced in \citet{RealToxicityPrompts}. We follow the same setup of \cite{OPT} and sample 25 generations of 20 tokens using nucleus sampling \textit{(p=0.9)} for each of 5000 randomly sampled prompts from RTP. We use the prompts to produce sequences (i.e, continuations) which are then scored by a toxicity classifier provided by Perspective API\footnote{\url{https://github.com/conversationai/perspectiveapi}}.

Figure \ref{fig:tox} plots the results. The chart shows the mean toxicity probability of continuations (y-axis), stratified across bucketed toxicities of the original prompts (x-axis). Galactica exhibits substantially lower toxicity rates than the other models.

\subsection{TruthfulQA}

TruthfulQA is a benchmark that measures answer truthfulness of language model generations~\citep{TruthfulQA}. It comprises 817 questions that span health, law, finance and other categories. We compare to other published language models. We report results in Table~\ref{table:truthqa}. Galactica exceeds the performance of other language models on this benchmark. However, absolute performance is still low. Given the curated nature of our corpus, this suggests that data alone does not cause language models to struggle at this task.

\begin{table}[h]
    \begin{center}
    \begin{tabular}{lrr}
    \toprule
      \multicolumn{3}{c}{TruthfulQA}      \\ 
    \midrule
    Model        & MC1 (Acc) & MC1 (Std)   \\ 
    \midrule
    OPT 175B     & 21\%     & 0.13       \\ 
    BLOOM 176B   & 19\%     & 0.07      \\ 
    GAL 125M & 19\%     & 0.11       \\
    GAL 1.3B & 19\%     & 0.15       \\
    GAL 6.7B & 19\%    & 0.03       \\
    GAL 30B  & 24\%     & 0.05      \\
    GAL 120B  & 26\%     & 0.02      \\
     \bottomrule
    \end{tabular}
    \end{center}
       \caption{\textbf{TruthfulQA Results}. Galactica exhibits superior performance to other language models, and performance increases with scale. but slowly and at low levels.}
\label{table:truthqa}
\end{table}

\section{Limitations and Future Work}

\subsection{Limitations}

We cover some of the limitations with work in this section.

\paragraph{Corpus Limitations} Our corpus has several limitations, both external and internally imposed. The main external constraint is our restriction to use open-access resources, and much of scientific knowledge like papers and textbooks are not open access. With access to these closed sources of knowledge, performance is likely to be considerably higher. We also use self-imposed constraints, like restricting the number of molecules and proteins for this work; without these constraints, we are likely to see considerable performance gains due to much larger corpuses for these modalities.

\paragraph{Corpus Effects vs Prompt Effects} In several benchmarks, we show performance gains over existing language models, but we do not specifically disentangle the effects of the prompts we included in pre-training versus the core scientific corpus. In future work, we likely need to disentangle these effects in order to see whether general language capabilities are possible with a scientific corpus alone without prompt boosting.

\paragraph{Citation Bias} While we demonstrate that the model approaches the true citation distribution with scale, some bias towards popular papers still remains with the 120B scale model, so the model likely requires augmentation before being used in a production environment.

\paragraph{Prompt Pre-Training vs Instruction Tuning} We opted for the former in this paper, but ideally we would need to explore what the latter could achieve, along the lines of the recent work of \citet{FLANPALM}. A limitation of this work is that we do not perform this direct comparison through ablations, making clear the trade-offs between approaches.

\paragraph{General Knowledge} While Galactica absorbs broad societal knowledge through sources such as Wikipedia - e.g. 120B knows Kota Kinabalu is the capital of Malaysia’s Sabah state - we would not advise using it for tasks that require this type of knowledge as this is not the intended use-case.

\paragraph{Text as a Modality} While we have shown text-based Transformers are surprisingly powerful with text representations of scientific phenomena, we caution against the interpretation that text is all you need. For example, in chemistry, geometry is a fundamental language that determines meaning, yet Galactica has no notion of geometry; e.g. 3D co-ordinates of atoms.

\subsection{Future Work}

For development of the base model, we highlight several directions that may be worth pursuing.

\paragraph{New Objective Function} It is likely further gains can be obtained with mixture-of-denoising training as U-PaLM has recently shown ~\citep{UPALM, FLANPALM}. We suspect this might be beneficial for the scientific modalities such as protein sequences, where the left-to-right LM objective is quite limiting.

\paragraph{Larger Context Window} We use a maximum context window length of $2048$ tokens in this work. Extending this is likely to be beneficial for understanding in long-form scientific documents, such as textbooks and also documents with longer modality sequences (e.g. long protein sequences).

\paragraph{Extending to Images} We cannot capture scientific knowledge adequately without capturing images. This is a natural follow-up project, although it likely requires some architectural modification to make it work well. Existing work such as \citet{Flamingo} has shown how to extend LLMs with this modality.

\paragraph{More <work> examples} We feel \verb|<work>| could be a general-purpose reasoning token and we would like to invest more in this direction, including increasing prompt diversity and exploring performance on more benchmarks. 

\paragraph{Verification} Even as language models become more accurate with scale, we need assurances that their generations are correct and factual. Developing this layer is critical for production applications of language models in general beyond scientific applications.

\paragraph{Continual Learning} Should we re-train from scratch to incorporate new scientific knowledge or train from older checkpoints? This is an open question, and further research is needed to find the best procedure for incorporating new knowledge into the model.

\paragraph{Retrieval Augmentation} While we have shown how large language models can absorb large bodies of scientific knowledge, retrieval has a place for fine-grained types of knowledge, and we believe this is a strong direction to pursue to complement the flexible weight memory of the Transformer.

\section{Discussion and Conclusion}

For over half a century, the dominant way of accessing scientific knowledge has been through a store-and-retrieve paradigm. The limitation of this approach is the reasoning, combining and organization of information still relies on human effort. This has led to a significant knowledge throughput bottleneck. In this work we explored how language models might disrupt this paradigm and bring about a new interface for humanity to interface with knowledge.

We showed that language models are surprisingly strong absorbers of technical knowledge, such as LaTeX equations and chemical reactions, and these capabilities tend to scale smoothly with model size. The context-associative power of language models likely confers significant advantages over search engines in the long-run. We demonstrated this for citation prediction, where a language model outperforms tuned sparse and dense retrieval pipelines for this task. Language models will likely provide a valuable new tool for exploring the literature and the body of scientific knowledge in coming years.

We also demonstrated that language models can compose a curated knowledge base to perform well in knowledge-intensive question answering tasks. This includes composing knowledge in a step-by-step reasoning manner. We showed that with a working memory token approach, we can achieve strong performance over existing methods on mathematical MMLU and MATH benchmarks. We suspect tasks like MATH are in principle solvable with language model approaches. The current bottleneck is the availability of high quality step-by-step datasets. However, language models will not perform these tasks like humans until they have an architectural change that supports adaptive computation.

We also performed initial investigations on the potential of LLMs to act as a bridge between scientific modalities and natural language. We showed Galactica could learn tasks like IUPAC naming through self-supervision. We also showed that it is possible to formulate drug discovery tasks like MoleculeNet in a natural language prompt and achieve strong results without direct fine-tuning. Lastly, we showed the potential for tasks such as automatic protein annotation. In all, increasing the number (and size) of datasets that bridge between natural language and natural sequences is likely to boost performance further.

Taken together, we feel there is a strong potential for language models to take on knowledge tasks that are currently human specialisms. We open source the models so others can build on our work, and we look forward to seeing how the open machine learning community will extend it.

\section*{Acknowledgments}
Thanks to to Susan Zhang, Stephen Roller, Naman Goyal and others for their support in using metaseq. We build on the open LLM training foundation they made possible with the OPT project~\citep{OPT}.

Thanks to Iliyan Zarov, Lukas Blecher, Jian Xiang Kuan and Mikhail Pershin for their contributions to the project.

Thanks to Faisal Azhar and Joe Spisak for their valuable support in delivering this project.

Thanks to Antonine Bordes, Laurens van der Maaten and Joelle Pineau for leadership support, and belief in this project. Additional thanks to Laurens for his valuable feedback on the paper.

Thanks to Geeta Chauhan, Hamid Shojanazeri and Eric Han for help with faster inference.

Thanks to numerous others for comments and advice over the past year: Patrick Lewis, Pontus Stenetorp, Timo Schick, Sebastian Riedel, Soumith Chintala.

Thanks to the open source creators whose libraries, datasets and other tools we utilized. Your efforts accelerated our efforts; and we open source our model to accelerate yours.

Thanks to the GPU nodes that didn't die on us when training the 120B model.

\bibliographystyle{plainnat}  
\bibliography{references}

\begin{thebibliography}{119}
\providecommand{\natexlab}[1]{#1}
\providecommand{\url}[1]{\texttt{#1}}
\expandafter\ifx\csname urlstyle\endcsname\relax
  \providecommand{\doi}[1]{doi: #1}\else
  \providecommand{\doi}{doi: \begingroup \urlstyle{rm}\Url}\fi

\bibitem[Alayrac et~al.(2022)Alayrac, Donahue, Luc, Miech, Barr, Hasson, Lenc,
  Mensch, Millican, Reynolds, Ring, Rutherford, Cabi, Han, Gong, Samangooei,
  Monteiro, Menick, Borgeaud, Brock, Nematzadeh, Sharifzadeh, Binkowski,
  Barreira, Vinyals, Zisserman, and Simonyan]{Flamingo}
Jean-Baptiste Alayrac, Jeff Donahue, Pauline Luc, Antoine Miech, Iain Barr,
  Yana Hasson, Karel Lenc, Arthur Mensch, Katie Millican, Malcolm Reynolds,
  Roman Ring, Eliza Rutherford, Serkan Cabi, Tengda Han, Zhitao Gong, Sina
  Samangooei, Marianne Monteiro, Jacob Menick, Sebastian Borgeaud, Andrew
  Brock, Aida Nematzadeh, Sahand Sharifzadeh, Mikolaj Binkowski, Ricardo
  Barreira, Oriol Vinyals, Andrew Zisserman, and Karen Simonyan.
\newblock Flamingo: a visual language model for few-shot learning, 2022.
\newblock URL \url{https://arxiv.org/abs/2204.14198}.

\bibitem[Altschul et~al.(1990)Altschul, Gish, Miller, Myers, and Lipman]{BLAST}
S~F Altschul, W~Gish, W~Miller, E~W Myers, and D~J Lipman.
\newblock Basic local alignment search tool.
\newblock \emph{J. Mol. Biol.}, 215\penalty0 (3):\penalty0 403--410, October
  1990.

\bibitem[Aribandi et~al.(2021)Aribandi, Tay, Schuster, Rao, Zheng, Mehta,
  Zhuang, Tran, Bahri, Ni, Gupta, Hui, Ruder, and Metzler]{ExT5}
Vamsi Aribandi, Yi~Tay, Tal Schuster, Jinfeng Rao, Huaixiu~Steven Zheng,
  Sanket~Vaibhav Mehta, Honglei Zhuang, Vinh~Q. Tran, Dara Bahri, Jianmo Ni,
  Jai Gupta, Kai Hui, Sebastian Ruder, and Donald Metzler.
\newblock Ext5: Towards extreme multi-task scaling for transfer learning, 2021.
\newblock URL \url{https://arxiv.org/abs/2111.10952}.

\bibitem[arXiv(2022)]{arxivpapers}
arXiv.
\newblock {arXiv Monthly Submissions}, 2022.
\newblock URL \url{https://arxiv.org/stats/monthly_submissions}.

\bibitem[Banino et~al.(2021)Banino, Balaguer, and Blundell]{PonderNet}
Andrea Banino, Jan Balaguer, and Charles Blundell.
\newblock Pondernet: Learning to ponder.
\newblock \emph{CoRR}, abs/2107.05407, 2021.
\newblock URL \url{https://arxiv.org/abs/2107.05407}.

\bibitem[Beltagy et~al.(2019)Beltagy, Cohan, and Lo]{SciBERT}
Iz~Beltagy, Arman Cohan, and Kyle Lo.
\newblock Scibert: Pretrained contextualized embeddings for scientific text.
\newblock \emph{CoRR}, abs/1903.10676, 2019.
\newblock URL \url{http://arxiv.org/abs/1903.10676}.

\bibitem[Black et~al.(2022)Black, Biderman, Hallahan, Anthony, Gao, Golding,
  He, Leahy, McDonell, Phang, Pieler, Prashanth, Purohit, Reynolds, Tow, Wang,
  and Weinbach]{GPTNeox}
Sid Black, Stella Biderman, Eric Hallahan, Quentin Anthony, Leo Gao, Laurence
  Golding, Horace He, Connor Leahy, Kyle McDonell, Jason Phang, Michael Pieler,
  USVSN~Sai Prashanth, Shivanshu Purohit, Laria Reynolds, Jonathan Tow, Ben
  Wang, and Samuel Weinbach.
\newblock Gpt-neox-20b: An open-source autoregressive language model, 2022.
\newblock URL \url{https://arxiv.org/abs/2204.06745}.

\bibitem[Blodgett et~al.(2020)Blodgett, Barocas, III, and
  Wallach]{LanguageIsPower}
Su~Lin Blodgett, Solon Barocas, Hal~Daum{\'{e}} III, and Hanna~M. Wallach.
\newblock Language (technology) is power: {A} critical survey of "bias" in
  {NLP}.
\newblock \emph{CoRR}, abs/2005.14050, 2020.
\newblock URL \url{https://arxiv.org/abs/2005.14050}.

\bibitem[Borgeaud et~al.(2021)Borgeaud, Mensch, Hoffmann, Cai, Rutherford,
  Millican, Driessche, Lespiau, Damoc, Clark, Casas, Guy, Menick, Ring,
  Hennigan, Huang, Maggiore, Jones, Cassirer, Brock, Paganini, Irving, Vinyals,
  Osindero, Simonyan, Rae, Elsen, and Sifre]{RETRO}
Sebastian Borgeaud, Arthur Mensch, Jordan Hoffmann, Trevor Cai, Eliza
  Rutherford, Katie Millican, George van~den Driessche, Jean-Baptiste Lespiau,
  Bogdan Damoc, Aidan Clark, Diego de~Las Casas, Aurelia Guy, Jacob Menick,
  Roman Ring, Tom Hennigan, Saffron Huang, Loren Maggiore, Chris Jones, Albin
  Cassirer, Andy Brock, Michela Paganini, Geoffrey Irving, Oriol Vinyals, Simon
  Osindero, Karen Simonyan, Jack~W. Rae, Erich Elsen, and Laurent Sifre.
\newblock Improving language models by retrieving from trillions of tokens,
  2021.
\newblock URL \url{https://arxiv.org/abs/2112.04426}.

\bibitem[Bornmann and Mutz(2014)]{GrowthRateScience}
Lutz Bornmann and R{\"{u}}diger Mutz.
\newblock Growth rates of modern science: {A} bibliometric analysis.
\newblock \emph{CoRR}, abs/1402.4578, 2014.
\newblock URL \url{http://arxiv.org/abs/1402.4578}.

\bibitem[Briol et~al.(2015)Briol, Oates, Girolami, and Osborne]{briol2015frank}
Fran{\c{c}}ois-Xavier Briol, Chris Oates, Mark Girolami, and Michael~A Osborne.
\newblock Frank-wolfe bayesian quadrature: Probabilistic integration with
  theoretical guarantees.
\newblock \emph{Advances in Neural Information Processing Systems}, 28, 2015.

\bibitem[Brown et~al.(2020)Brown, Mann, Ryder, Subbiah, Kaplan, Dhariwal,
  Neelakantan, Shyam, Sastry, Askell, Agarwal, Herbert{-}Voss, Krueger,
  Henighan, Child, Ramesh, Ziegler, Wu, Winter, Hesse, Chen, Sigler, Litwin,
  Gray, Chess, Clark, Berner, McCandlish, Radford, Sutskever, and Amodei]{GPT3}
Tom~B. Brown, Benjamin Mann, Nick Ryder, Melanie Subbiah, Jared Kaplan,
  Prafulla Dhariwal, Arvind Neelakantan, Pranav Shyam, Girish Sastry, Amanda
  Askell, Sandhini Agarwal, Ariel Herbert{-}Voss, Gretchen Krueger, Tom
  Henighan, Rewon Child, Aditya Ramesh, Daniel~M. Ziegler, Jeffrey Wu, Clemens
  Winter, Christopher Hesse, Mark Chen, Eric Sigler, Mateusz Litwin, Scott
  Gray, Benjamin Chess, Jack Clark, Christopher Berner, Sam McCandlish, Alec
  Radford, Ilya Sutskever, and Dario Amodei.
\newblock Language models are few-shot learners.
\newblock \emph{CoRR}, abs/2005.14165, 2020.
\newblock URL \url{https://arxiv.org/abs/2005.14165}.

\bibitem[Bush(1945)]{bush1945}
Vannevar Bush.
\newblock {As} {We} {May} {Think}.
\newblock \emph{Atlantic Monthly 176 (July 1945)}, pages 101--108, 1945.

\bibitem[Cachola et~al.(2020)Cachola, Lo, Cohan, and Weld]{SciTLDR}
Isabel Cachola, Kyle Lo, Arman Cohan, and Daniel~S. Weld.
\newblock {TLDR:} extreme summarization of scientific documents.
\newblock \emph{CoRR}, abs/2004.15011, 2020.
\newblock URL \url{https://arxiv.org/abs/2004.15011}.

\bibitem[Chowdhery et~al.(2022)Chowdhery, Narang, Devlin, Bosma, Mishra,
  Roberts, Barham, Chung, Sutton, Gehrmann, Schuh, Shi, Tsvyashchenko, Maynez,
  Rao, Barnes, Tay, Shazeer, Prabhakaran, Reif, Du, Hutchinson, Pope, Bradbury,
  Austin, Isard, Gur-Ari, Yin, Duke, Levskaya, Ghemawat, Dev, Michalewski,
  Garcia, Misra, Robinson, Fedus, Zhou, Ippolito, Luan, Lim, Zoph, Spiridonov,
  Sepassi, Dohan, Agrawal, Omernick, Dai, Pillai, Pellat, Lewkowycz, Moreira,
  Child, Polozov, Lee, Zhou, Wang, Saeta, Diaz, Firat, Catasta, Wei,
  Meier-Hellstern, Eck, Dean, Petrov, and Fiedel]{PaLM}
Aakanksha Chowdhery, Sharan Narang, Jacob Devlin, Maarten Bosma, Gaurav Mishra,
  Adam Roberts, Paul Barham, Hyung~Won Chung, Charles Sutton, Sebastian
  Gehrmann, Parker Schuh, Kensen Shi, Sasha Tsvyashchenko, Joshua Maynez,
  Abhishek Rao, Parker Barnes, Yi~Tay, Noam Shazeer, Vinodkumar Prabhakaran,
  Emily Reif, Nan Du, Ben Hutchinson, Reiner Pope, James Bradbury, Jacob
  Austin, Michael Isard, Guy Gur-Ari, Pengcheng Yin, Toju Duke, Anselm
  Levskaya, Sanjay Ghemawat, Sunipa Dev, Henryk Michalewski, Xavier Garcia,
  Vedant Misra, Kevin Robinson, Liam Fedus, Denny Zhou, Daphne Ippolito, David
  Luan, Hyeontaek Lim, Barret Zoph, Alexander Spiridonov, Ryan Sepassi, David
  Dohan, Shivani Agrawal, Mark Omernick, Andrew~M. Dai,
  Thanumalayan~Sankaranarayana Pillai, Marie Pellat, Aitor Lewkowycz, Erica
  Moreira, Rewon Child, Oleksandr Polozov, Katherine Lee, Zongwei Zhou, Xuezhi
  Wang, Brennan Saeta, Mark Diaz, Orhan Firat, Michele Catasta, Jason Wei,
  Kathy Meier-Hellstern, Douglas Eck, Jeff Dean, Slav Petrov, and Noah Fiedel.
\newblock Palm: Scaling language modeling with pathways, 2022.
\newblock URL \url{https://arxiv.org/abs/2204.02311}.

\bibitem[Chung et~al.(2022)Chung, Hou, Longpre, Zoph, Tay, Fedus, Li, Wang,
  Dehghani, Brahma, Webson, Gu, Dai, Suzgun, Chen, Chowdhery, Narang, Mishra,
  Yu, Zhao, Huang, Dai, Yu, Petrov, Chi, Dean, Devlin, Roberts, Zhou, Le, and
  Wei]{FLANPALM}
Hyung~Won Chung, Le~Hou, Shayne Longpre, Barret Zoph, Yi~Tay, William Fedus,
  Eric Li, Xuezhi Wang, Mostafa Dehghani, Siddhartha Brahma, Albert Webson,
  Shixiang~Shane Gu, Zhuyun Dai, Mirac Suzgun, Xinyun Chen, Aakanksha
  Chowdhery, Sharan Narang, Gaurav Mishra, Adams Yu, Vincent Zhao, Yanping
  Huang, Andrew Dai, Hongkun Yu, Slav Petrov, Ed~H. Chi, Jeff Dean, Jacob
  Devlin, Adam Roberts, Denny Zhou, Quoc~V. Le, and Jason Wei.
\newblock Scaling instruction-finetuned language models, 2022.
\newblock URL \url{https://arxiv.org/abs/2210.11416}.

\bibitem[Clark et~al.(2019)Clark, Lee, Chang, Kwiatkowski, Collins, and
  Toutanova]{BoolQ}
Christopher Clark, Kenton Lee, Ming{-}Wei Chang, Tom Kwiatkowski, Michael
  Collins, and Kristina Toutanova.
\newblock Boolq: Exploring the surprising difficulty of natural yes/no
  questions.
\newblock \emph{CoRR}, abs/1905.10044, 2019.
\newblock URL \url{http://arxiv.org/abs/1905.10044}.

\bibitem[Cobbe et~al.(2021)Cobbe, Kosaraju, Bavarian, Hilton, Nakano, Hesse,
  and Schulman]{GSM8k}
Karl Cobbe, Vineet Kosaraju, Mohammad Bavarian, Jacob Hilton, Reiichiro Nakano,
  Christopher Hesse, and John Schulman.
\newblock Training verifiers to solve math word problems.
\newblock \emph{CoRR}, abs/2110.14168, 2021.
\newblock URL \url{https://arxiv.org/abs/2110.14168}.

\bibitem[Dasigi et~al.(2019)Dasigi, Liu, Marasovi{\'c}, Smith, and
  Gardner]{Quoref}
Pradeep Dasigi, Nelson~F. Liu, Ana Marasovi{\'c}, Noah~A. Smith, and Matt
  Gardner.
\newblock Quoref: A reading comprehension dataset with questions requiring
  coreferential reasoning.
\newblock In \emph{EMNLP}, 2019.

\bibitem[Dasigi et~al.(2021)Dasigi, Lo, Beltagy, Cohan, Smith, and
  Gardner]{QASPER}
Pradeep Dasigi, Kyle Lo, Iz~Beltagy, Arman Cohan, Noah~A. Smith, and Matt
  Gardner.
\newblock A dataset of information-seeking questions and answers anchored in
  research papers.
\newblock In \emph{NAACL}, 2021.

\bibitem[Dev et~al.(2019)Dev, Li, Phillips, and Srikumar]{MeasureMitigate}
Sunipa Dev, Tao Li, Jeff~M. Phillips, and Vivek Srikumar.
\newblock On measuring and mitigating biased inferences of word embeddings.
\newblock \emph{CoRR}, abs/1908.09369, 2019.
\newblock URL \url{http://arxiv.org/abs/1908.09369}.

\bibitem[Dinan et~al.(2018)Dinan, Roller, Shuster, Fan, Auli, and
  Weston]{WizardofWikipedia}
Emily Dinan, Stephen Roller, Kurt Shuster, Angela Fan, Michael Auli, and Jason
  Weston.
\newblock Wizard of wikipedia: Knowledge-powered conversational agents, 2018.
\newblock URL \url{https://arxiv.org/abs/1811.01241}.

\bibitem[Favre and Powerll()]{IUPACNaming}
Henri~A. Favre and Warren~H. Powerll.
\newblock Nomenclature of organic chemistry: Iupac recommendations and
  preferred names 2013.

\bibitem[Galilei(1623)]{Assayer}
Galileo Galilei.
\newblock Assayer.
\newblock 1623.

\bibitem[Gao et~al.(2022)Gao, Dai, Pasupat, Chen, Chaganty, Fan, Zhao, Lao,
  Lee, Juan, and Guu]{PostHocResearch}
Luyu Gao, Zhuyun Dai, Panupong Pasupat, Anthony Chen, Arun~Tejasvi Chaganty,
  Yicheng Fan, Vincent~Y. Zhao, Ni~Lao, Hongrae Lee, Da-Cheng Juan, and Kelvin
  Guu.
\newblock Attributed text generation via post-hoc research and revision, 2022.
\newblock URL \url{https://arxiv.org/abs/2210.08726}.

\bibitem[García-Ortegón et~al.(2022)García-Ortegón, Simm, Tripp,
  Hernández-Lobato, Bender, and Bacallado]{DockSTRING}
Miguel García-Ortegón, Gregor N.~C. Simm, Austin~J. Tripp, José~Miguel
  Hernández-Lobato, Andreas Bender, and Sergio Bacallado.
\newblock Dockstring: Easy molecular docking yields better benchmarks for
  ligand design.
\newblock \emph{Journal of Chemical Information and Modeling}, 62\penalty0
  (15):\penalty0 3486--3502, 2022.
\newblock \doi{10.1021/acs.jcim.1c01334}.
\newblock URL \url{https://doi.org/10.1021/acs.jcim.1c01334}.
\newblock PMID: 35849793.

\bibitem[Gehman et~al.(2020)Gehman, Gururangan, Sap, Choi, and
  Smith]{RealToxicityPrompts}
Samuel Gehman, Suchin Gururangan, Maarten Sap, Yejin Choi, and Noah~A. Smith.
\newblock Realtoxicityprompts: Evaluating neural toxic degeneration in language
  models.
\newblock \emph{ArXiv}, abs/2009.11462, 2020.

\bibitem[GenBank(2022)]{genbank}
GenBank.
\newblock {GenBank and WGS Statistics}, 2022.
\newblock URL \url{https://www.ncbi.nlm.nih.gov/genbank/statistics}.

\bibitem[Graves(2016)]{ACT}
Alex Graves.
\newblock Adaptive computation time for recurrent neural networks, 2016.
\newblock URL \url{https://arxiv.org/abs/1603.08983}.

\bibitem[GROBID(2008--2022)]{GROBID}
GROBID.
\newblock Grobid.
\newblock \url{https://github.com/kermitt2/grobid}, 2008--2022.

\bibitem[Gu et~al.(2020)Gu, Tinn, Cheng, Lucas, Usuyama, Liu, Naumann, Gao, and
  Poon]{PubMedBERT}
Yu~Gu, Robert Tinn, Hao Cheng, Michael Lucas, Naoto Usuyama, Xiaodong Liu,
  Tristan Naumann, Jianfeng Gao, and Hoifung Poon.
\newblock Domain-specific language model pretraining for biomedical natural
  language processing.
\newblock \emph{CoRR}, abs/2007.15779, 2020.
\newblock URL \url{https://arxiv.org/abs/2007.15779}.

\bibitem[Gunasekara et~al.(2019)Gunasekara, Kummerfeld, Polymenakos, and
  Lasecki]{Advising}
Chulaka Gunasekara, Jonathan~K. Kummerfeld, Lazaros Polymenakos, and Walter
  Lasecki.
\newblock {DSTC}7 task 1: Noetic end-to-end response selection.
\newblock In \emph{Proceedings of the First Workshop on NLP for Conversational
  AI}, pages 60--67, Florence, Italy, August 2019. Association for
  Computational Linguistics.
\newblock \doi{10.18653/v1/W19-4107}.
\newblock URL \url{https://aclanthology.org/W19-4107}.

\bibitem[Hendrycks and Gimpel(2016)]{GeLU}
Dan Hendrycks and Kevin Gimpel.
\newblock Gaussian error linear units (gelus), 2016.
\newblock URL \url{https://arxiv.org/abs/1606.08415}.

\bibitem[Hendrycks et~al.(2020)Hendrycks, Burns, Basart, Zou, Mazeika, Song,
  and Steinhardt]{MMMLU}
Dan Hendrycks, Collin Burns, Steven Basart, Andy Zou, Mantas Mazeika, Dawn
  Song, and Jacob Steinhardt.
\newblock Measuring massive multitask language understanding, 2020.
\newblock URL \url{https://arxiv.org/abs/2009.03300}.

\bibitem[Hendrycks et~al.(2021)Hendrycks, Burns, Kadavath, Arora, Basart, Tang,
  Song, and Steinhardt]{MATH}
Dan Hendrycks, Collin Burns, Saurav Kadavath, Akul Arora, Steven Basart, Eric
  Tang, Dawn Song, and Jacob Steinhardt.
\newblock Measuring mathematical problem solving with the {MATH} dataset.
\newblock \emph{CoRR}, abs/2103.03874, 2021.
\newblock URL \url{https://arxiv.org/abs/2103.03874}.

\bibitem[Hernandez et~al.(2022)Hernandez, Brown, Conerly, DasSarma, Drain,
  El-Showk, Elhage, Hatfield-Dodds, Henighan, Hume, Johnston, Mann, Olah,
  Olsson, Amodei, Joseph, Kaplan, and McCandlish]{HarmfulRepeats}
Danny Hernandez, Tom Brown, Tom Conerly, Nova DasSarma, Dawn Drain, Sheer
  El-Showk, Nelson Elhage, Zac Hatfield-Dodds, Tom Henighan, Tristan Hume,
  Scott Johnston, Ben Mann, Chris Olah, Catherine Olsson, Dario Amodei,
  Nicholas Joseph, Jared Kaplan, and Sam McCandlish.
\newblock Scaling laws and interpretability of learning from repeated data,
  2022.
\newblock URL \url{https://arxiv.org/abs/2205.10487}.

\bibitem[Hirschmann(1964)]{WrightsLaw}
Winfred~B. Hirschmann.
\newblock {Profit} from the {Learning} {Curve}, January 1964.

\bibitem[Hoffmann et~al.(2022)Hoffmann, Borgeaud, Mensch, Buchatskaya, Cai,
  Rutherford, Casas, Hendricks, Welbl, Clark, Hennigan, Noland, Millican,
  Driessche, Damoc, Guy, Osindero, Simonyan, Elsen, Rae, Vinyals, and
  Sifre]{Chinchilla}
Jordan Hoffmann, Sebastian Borgeaud, Arthur Mensch, Elena Buchatskaya, Trevor
  Cai, Eliza Rutherford, Diego de~Las Casas, Lisa~Anne Hendricks, Johannes
  Welbl, Aidan Clark, Tom Hennigan, Eric Noland, Katie Millican, George van~den
  Driessche, Bogdan Damoc, Aurelia Guy, Simon Osindero, Karen Simonyan, Erich
  Elsen, Jack~W. Rae, Oriol Vinyals, and Laurent Sifre.
\newblock Training compute-optimal large language models, 2022.
\newblock URL \url{https://arxiv.org/abs/2203.15556}.

\bibitem[Honda et~al.(2019)Honda, Shi, and Ueda]{honda2019smiles}
Shion Honda, Shoi Shi, and Hiroki~R. Ueda.
\newblock Smiles transformer: Pre-trained molecular fingerprint for low data
  drug discovery.
\newblock 2019.

\bibitem[Hong et~al.(2022)Hong, Ajith, Pauloski, Duede, Malamud, Magoulas,
  Chard, and Foster]{ScholarBERT}
Zhi Hong, Aswathy Ajith, Gregory Pauloski, Eamon Duede, Carl Malamud, Roger
  Magoulas, Kyle Chard, and Ian Foster.
\newblock Scholarbert: Bigger is not always better, 2022.
\newblock URL \url{https://arxiv.org/abs/2205.11342}.

\bibitem[Irwin et~al.(2021)Irwin, Dimitriadis, He, and Bjerrum]{Chemformer}
Ross Irwin, Spyridon Dimitriadis, Jiazhen He, and Esben Bjerrum.
\newblock Chemformer: A pre-trained transformer for computational chemistry.
\newblock \emph{ChemRxiv}, 2021.
\newblock \doi{10.26434/chemrxiv-2021-v2pnn}.

\bibitem[Izacard et~al.(2021)Izacard, Caron, Hosseini, Riedel, Bojanowski,
  Joulin, and Grave]{Contriever}
Gautier Izacard, Mathilde Caron, Lucas Hosseini, Sebastian Riedel, Piotr
  Bojanowski, Armand Joulin, and Edouard Grave.
\newblock Towards unsupervised dense information retrieval with contrastive
  learning.
\newblock \emph{CoRR}, abs/2112.09118, 2021.
\newblock URL \url{https://arxiv.org/abs/2112.09118}.

\bibitem[Izacard et~al.(2022)Izacard, Lewis, Lomeli, Hosseini, Petroni, Schick,
  Dwivedi-Yu, Joulin, Riedel, and Grave]{izacard2022fewshot}
Gautier Izacard, Patrick Lewis, Maria Lomeli, Lucas Hosseini, Fabio Petroni,
  Timo Schick, Jane Dwivedi-Yu, Armand Joulin, Sebastian Riedel, and Edouard
  Grave.
\newblock Few-shot learning with retrieval augmented language models, 2022.

\bibitem[Jackson(1990)]{Jackson}
Peter Jackson.
\newblock \emph{Introduction to Expert Systems}.
\newblock Addison-Wesley Longman Publishing Co., Inc., USA, 2nd edition, 1990.
\newblock ISBN 0201175789.

\bibitem[Jin et~al.(2019)Jin, Dhingra, Liu, Cohen, and Lu]{PubMedQA}
Qiao Jin, Bhuwan Dhingra, Zhengping Liu, William~W. Cohen, and Xinghua Lu.
\newblock Pubmedqa: {A} dataset for biomedical research question answering.
\newblock \emph{CoRR}, abs/1909.06146, 2019.
\newblock URL \url{http://arxiv.org/abs/1909.06146}.

\bibitem[Johnson et~al.(2019)Johnson, Douze, and J{\'e}gou]{FAISS}
Jeff Johnson, Matthijs Douze, and Herv{\'e} J{\'e}gou.
\newblock Billion-scale similarity search with {GPUs}.
\newblock \emph{IEEE Transactions on Big Data}, 7\penalty0 (3):\penalty0
  535--547, 2019.

\bibitem[Joulin et~al.(2016)Joulin, Grave, Bojanowski, and
  Mikolov]{joulin2016bag}
Armand Joulin, Edouard Grave, Piotr Bojanowski, and Tomas Mikolov.
\newblock Bag of tricks for efficient text classification.
\newblock \emph{arXiv preprint arXiv:1607.01759}, 2016.

\bibitem[Jumper et~al.(2021)Jumper, Evans, Pritzel, Green, Figurnov,
  Ronneberger, Tunyasuvunakool, Bates, {\v{Z}}{\'\i}dek, Potapenko, Bridgland,
  Meyer, Kohl, Ballard, Cowie, Romera-Paredes, Nikolov, Jain, Adler, Back,
  Petersen, Reiman, Clancy, Zielinski, Steinegger, Pacholska, Berghammer,
  Bodenstein, Silver, Vinyals, Senior, Kavukcuoglu, Kohli, and
  Hassabis]{AlphaFold2021}
John Jumper, Richard Evans, Alexander Pritzel, Tim Green, Michael Figurnov,
  Olaf Ronneberger, Kathryn Tunyasuvunakool, Russ Bates, Augustin
  {\v{Z}}{\'\i}dek, Anna Potapenko, Alex Bridgland, Clemens Meyer, Simon A~A
  Kohl, Andrew~J Ballard, Andrew Cowie, Bernardino Romera-Paredes, Stanislav
  Nikolov, Rishub Jain, Jonas Adler, Trevor Back, Stig Petersen, David Reiman,
  Ellen Clancy, Michal Zielinski, Martin Steinegger, Michalina Pacholska, Tamas
  Berghammer, Sebastian Bodenstein, David Silver, Oriol Vinyals, Andrew~W
  Senior, Koray Kavukcuoglu, Pushmeet Kohli, and Demis Hassabis.
\newblock Highly accurate protein structure prediction with {AlphaFold}.
\newblock \emph{Nature}, 596\penalty0 (7873):\penalty0 583--589, 2021.
\newblock \doi{10.1038/s41586-021-03819-2}.

\bibitem[Kaplan et~al.(2020)Kaplan, McCandlish, Henighan, Brown, Chess, Child,
  Gray, Radford, Wu, and Amodei]{ScalingLaws}
Jared Kaplan, Sam McCandlish, Tom Henighan, Tom~B. Brown, Benjamin Chess, Rewon
  Child, Scott Gray, Alec Radford, Jeffrey Wu, and Dario Amodei.
\newblock Scaling laws for neural language models.
\newblock \emph{CoRR}, abs/2001.08361, 2020.
\newblock URL \url{https://arxiv.org/abs/2001.08361}.

\bibitem[Kembhavi et~al.(2017)Kembhavi, Seo, Schwenk, Choi, Farhadi, and
  Hajishirzi]{TQA}
Aniruddha Kembhavi, Minjoon Seo, Dustin Schwenk, Jonghyun Choi, Ali Farhadi,
  and Hannaneh Hajishirzi.
\newblock Are you smarter than a sixth grader? textbook question answering for
  multimodal machine comprehension.
\newblock \emph{2017 IEEE Conference on Computer Vision and Pattern Recognition
  (CVPR)}, pages 5376--5384, 2017.

\bibitem[Khashabi et~al.(2020)Khashabi, Min, Khot, Sabharwal, Tafjord, Clark,
  and Hajishirzi]{UnifiedQA}
Daniel Khashabi, Sewon Min, Tushar Khot, Ashish Sabharwal, Oyvind Tafjord,
  Peter Clark, and Hannaneh Hajishirzi.
\newblock Unifiedqa: Crossing format boundaries with a single qa system, 2020.
\newblock URL \url{https://arxiv.org/abs/2005.00700}.

\bibitem[Khot et~al.(2018)Khot, Sabharwal, and Clark]{SciTail}
Tushar Khot, Ashish Sabharwal, and Peter Clark.
\newblock Scitail: A textual entailment dataset from science question
  answering.
\newblock In \emph{AAAI}, 2018.

\bibitem[Khot et~al.(2020)Khot, Clark, Guerquin, Jansen, and Sabharwal]{QASC}
Tushar Khot, Peter Clark, Michal Guerquin, Peter~Alexander Jansen, and Ashish
  Sabharwal.
\newblock Qasc: A dataset for question answering via sentence composition.
\newblock \emph{ArXiv}, abs/1910.11473, 2020.

\bibitem[Kim et~al.(2004)Kim, Ohta, Tsuruoka, Tateisi, and Collier]{JNLPBA}
J.-D. Kim, T.~Ohta, Y.~Tsuruoka, Y.~Tateisi, and N.~Collier.
\newblock Introduction to the bio-entity recognition task at jnlpba.
\newblock \emph{International Joint Workshop on Natural Language Processing in
  Biomedicine and its Applications}, 2004.

\bibitem[Kojima et~al.(2022)Kojima, Gu, Reid, Matsuo, and Iwasawa]{StepByStep}
Takeshi Kojima, Shixiang~Shane Gu, Machel Reid, Yutaka Matsuo, and Yusuke
  Iwasawa.
\newblock Large language models are zero-shot reasoners, 2022.
\newblock URL \url{https://arxiv.org/abs/2205.11916}.

\bibitem[Krallinger et~al.(2004)Krallinger, Rabal, Leitner, Miguel~Vazquez, Lu,
  Leaman, Yanan~Lu, Rak, Huber, Rocktäschel, andDavid Campos, Tang, Xu,
  Munkhdalai, Ryu, Ramanan, Nathan, Žitnik, Bajec, Weber, Irmer, Akhondi,
  Kors, Xu, An, Sikdar, Ekbal, Masaharu~Yoshioka, Choi, Verspoor, Khabsa,
  Giles, Liu, Ravikumar, Andre~Lamurias, Dai, Tsai, Ata, Can, Usié, Alves,
  Segura-Bedmar, Martínez, Oyarzabal, and Valencia]{BC4CHEMD}
Martin Krallinger, Obdulia Rabal, Florian Leitner, David~Salgado
  Miguel~Vazquez, Zhiyong Lu, Robert Leaman, Donghong Ji andDaniel M Lowe
  andRoger A Sayle andRiza Theresa Batista-Navarro Yanan~Lu, Rafal Rak, Torsten
  Huber, Tim Rocktäschel, Sérgio~Matos andDavid Campos, Buzhou Tang, Hua Xu,
  Tsendsuren Munkhdalai, Keun~Ho Ryu, SV~Ramanan, Senthil Nathan, Slavko
  Žitnik, Marko Bajec, Lutz Weber, Matthias Irmer, Saber~A Akhondi, Jan~A
  Kors, Shuo Xu, Xin An, Utpal~Kumar Sikdar, Asif Ekbal, Thaer M~Dieb
  Masaharu~Yoshioka, Miji Choi, Karin Verspoor, Madian Khabsa, C~Lee Giles,
  Hongfang Liu, Komandur~Elayavilli Ravikumar, Francisco M~Couto
  Andre~Lamurias, Hong-Jie Dai, Richard Tzong-Han Tsai, Caglar Ata, Tolga Can,
  Anabel Usié, Rui Alves, Isabel Segura-Bedmar, Paloma Martínez, Julen
  Oyarzabal, and Alfonso Valencia.
\newblock The chemdner corpus of chemicals and drugs and its annotation
  principles.
\newblock \emph{J Cheminform}, 2004.

\bibitem[Krasnov et~al.(2021)Krasnov, Khokhlov, Fedorov, and
  Sosnin]{Struct2IUPAC}
Lev Krasnov, Ivan Khokhlov, Maxim~V. Fedorov, and Sergey Sosnin.
\newblock Transformer-based artificial neural networks for the conversion
  between chemical notations, 2021.
\newblock URL
  \url{https://jcheminf.biomedcentral.com/articles/10.1186/s13321-021-00512-4}.

\bibitem[Kurita et~al.(2019)Kurita, Vyas, Pareek, Black, and
  Tsvetkov]{MeasuringBias}
Keita Kurita, Nidhi Vyas, Ayush Pareek, Alan~W. Black, and Yulia Tsvetkov.
\newblock Measuring bias in contextualized word representations.
\newblock \emph{CoRR}, abs/1906.07337, 2019.
\newblock URL \url{http://arxiv.org/abs/1906.07337}.

\bibitem[Lee et~al.(2022)Lee, Ippolito, Nystrom, Zhang, Eck, Callison-Burch,
  and Carlini]{lee2021deduplicating}
Katherine Lee, Daphne Ippolito, Andrew Nystrom, Chiyuan Zhang, Douglas Eck,
  Chris Callison-Burch, and Nicholas Carlini.
\newblock Deduplicating training data makes language models better.
\newblock In \emph{Proceedings of the 60th Annual Meeting of the Association
  for Computational Linguistics}. Association for Computational Linguistics,
  2022.

\bibitem[Lewis et~al.(2020{\natexlab{a}})Lewis, Ott, Du, and Stoyanov]{BioLM}
Patrick Lewis, Myle Ott, Jingfei Du, and Veselin Stoyanov.
\newblock Pretrained language models for biomedical and clinical tasks:
  Understanding and extending the state-of-the-art.
\newblock In \emph{Proceedings of the 3rd Clinical Natural Language Processing
  Workshop}, pages 146--157, Online, November 2020{\natexlab{a}}. Association
  for Computational Linguistics.
\newblock \doi{10.18653/v1/2020.clinicalnlp-1.17}.
\newblock URL \url{https://aclanthology.org/2020.clinicalnlp-1.17}.

\bibitem[Lewis et~al.(2020{\natexlab{b}})Lewis, Perez, Piktus, Petroni,
  Karpukhin, Goyal, Küttler, Lewis, Yih, Rocktäschel, Riedel, and Kiela]{RAG}
Patrick Lewis, Ethan Perez, Aleksandra Piktus, Fabio Petroni, Vladimir
  Karpukhin, Naman Goyal, Heinrich Küttler, Mike Lewis, Wen-tau Yih, Tim
  Rocktäschel, Sebastian Riedel, and Douwe Kiela.
\newblock Retrieval-augmented generation for knowledge-intensive nlp tasks,
  2020{\natexlab{b}}.
\newblock URL \url{https://arxiv.org/abs/2005.11401}.

\bibitem[Lewkowycz et~al.(2022)Lewkowycz, Andreassen, Dohan, Dyer, Michalewski,
  Ramasesh, Slone, Anil, Schlag, Gutman-Solo, Wu, Neyshabur, Gur-Ari, and
  Misra]{Minerva}
Aitor Lewkowycz, Anders Andreassen, David Dohan, Ethan Dyer, Henryk
  Michalewski, Vinay Ramasesh, Ambrose Slone, Cem Anil, Imanol Schlag, Theo
  Gutman-Solo, Yuhuai Wu, Behnam Neyshabur, Guy Gur-Ari, and Vedant Misra.
\newblock Solving quantitative reasoning problems with language models, 2022.
\newblock URL \url{https://arxiv.org/abs/2206.14858}.

\bibitem[Li et~al.(2016)Li, Sun, Johnson, Sciaky, Wei, Leaman, Davis,
  Mattingly, Wiegers, and Lu]{BC5CDR}
Jiao Li, Yueping Sun, Robin~J Johnson, Daniela Sciaky, Chih-Hsuan Wei, Robert
  Leaman, Allan~Peter Davis, Carolyn~J Mattingly, Thomas~C Wiegers, and Zhiyong
  Lu.
\newblock {BioCreative} {V} {CDR} task corpus: a resource for chemical disease
  relation extraction.
\newblock \emph{Database (Oxford)}, 2016:\penalty0 baw068, May 2016.

\bibitem[Licklider(1960)]{licklider1960}
J.R. Licklider.
\newblock {Man-Computer Symbiosis}.
\newblock \emph{IRE Transactions on Human Factors in Electronics, HFE-1}, pages
  4--11, 1960.

\bibitem[Lin et~al.(2019)Lin, Tafjord, Clark, and Gardner]{ROPES}
Kevin Lin, Oyvind Tafjord, Peter Clark, and Matt Gardner.
\newblock Reasoning over paragraph effects in situations.
\newblock \emph{ArXiv}, abs/1908.05852, 2019.

\bibitem[Lin et~al.(2022{\natexlab{a}})Lin, Hilton, and Evans]{TruthfulQA}
Stephanie Lin, Jacob Hilton, and Owain Evans.
\newblock {T}ruthful{QA}: Measuring how models mimic human falsehoods.
\newblock In \emph{Proceedings of the 60th Annual Meeting of the Association
  for Computational Linguistics (Volume 1: Long Papers)}, pages 3214--3252,
  Dublin, Ireland, May 2022{\natexlab{a}}. Association for Computational
  Linguistics.
\newblock \doi{10.18653/v1/2022.acl-long.229}.
\newblock URL \url{https://aclanthology.org/2022.acl-long.229}.

\bibitem[Lin et~al.(2022{\natexlab{b}})Lin, Akin, Rao, Hie, Zhu, Lu,
  Santos~Costa, Fazel-Zarandi, Sercu, Candido, and Rives]{ESMFold}
Zeming Lin, Halil Akin, Roshan Rao, Brian Hie, Zhongkai Zhu, Wenting Lu,
  Allan~dos Santos~Costa, Maryam Fazel-Zarandi, Tom Sercu, Sal Candido, and
  Alexander Rives.
\newblock Language models of protein sequences at the scale of evolution enable
  accurate structure prediction.
\newblock \emph{bioRxiv}, 2022{\natexlab{b}}.
\newblock \doi{10.1101/2022.07.20.500902}.
\newblock URL
  \url{https://www.biorxiv.org/content/early/2022/07/21/2022.07.20.500902}.

\bibitem[Lo et~al.(2019{\natexlab{a}})Lo, Wang, Neumann, Kinney, and
  Weld]{S2ORC}
Kyle Lo, Lucy~Lu Wang, Mark Neumann, Rodney Kinney, and Daniel~S. Weld.
\newblock {GORC:} {A} large contextual citation graph of academic papers.
\newblock \emph{CoRR}, abs/1911.02782, 2019{\natexlab{a}}.
\newblock URL \url{http://arxiv.org/abs/1911.02782}.

\bibitem[Lo et~al.(2019{\natexlab{b}})Lo, Wang, Neumann, Kinney, and
  Weld]{S2ORCBERT}
Kyle Lo, Lucy~Lu Wang, Mark Neumann, Rodney Kinney, and Daniel~S. Weld.
\newblock {GORC:} {A} large contextual citation graph of academic papers.
\newblock \emph{CoRR}, abs/1911.02782, 2019{\natexlab{b}}.
\newblock URL \url{http://arxiv.org/abs/1911.02782}.

\bibitem[Loshchilov and Hutter(2017)]{AdamW}
Ilya Loshchilov and Frank Hutter.
\newblock Fixing weight decay regularization in adam.
\newblock \emph{CoRR}, abs/1711.05101, 2017.
\newblock URL \url{http://arxiv.org/abs/1711.05101}.

\bibitem[Lowe et~al.(2011)Lowe, Corbett, Murray-Rust, and Glen]{OPSIN}
Daniel~M. Lowe, Peter~T. Corbett, Peter Murray-Rust, and Robert~C. Glen.
\newblock Chemical name to structure: Opsin, an open source solution, 2011.
\newblock URL \url{https://pubs.acs.org/doi/full/10.1021/ci100384d}.

\bibitem[Marx(2013)]{BigChallengesBigData}
Vivien Marx.
\newblock The big challenges of big data.
\newblock \emph{Nature}, 498:\penalty0 255--260, 2013.
\newblock URL \url{https://www.nature.com/articles/498255a}.

\bibitem[Massey(1951)]{Massey_1951}
Frank~J. Massey.
\newblock The kolmogorov-smirnov test for goodness of fit.
\newblock \emph{Journal of the American Statistical Association}, 46\penalty0
  (253):\penalty0 68--78, mar 1951.
\newblock \doi{10.1080/01621459.1951.10500769}.
\newblock URL \url{https://doi.org/10.1080\%2F01621459.1951.10500769}.

\bibitem[Mihaylov et~al.(2018)Mihaylov, Clark, Khot, and Sabharwal]{OpenBookQA}
Todor Mihaylov, Peter Clark, Tushar Khot, and Ashish Sabharwal.
\newblock Can a suit of armor conduct electricity? a new dataset for open book
  question answering.
\newblock In \emph{EMNLP}, 2018.

\bibitem[Mitchell et~al.(2022)Mitchell, Lin, Bosselut, Manning, and
  Finn]{MBMES}
Eric Mitchell, Charles Lin, Antoine Bosselut, Christopher~D. Manning, and
  Chelsea Finn.
\newblock Memory-based model editing at scale, 2022.
\newblock URL \url{https://arxiv.org/abs/2206.06520}.

\bibitem[Nadeem et~al.(2021)Nadeem, Bethke, and Reddy]{stereoset}
Moin Nadeem, Anna Bethke, and Siva Reddy.
\newblock {S}tereo{S}et: Measuring stereotypical bias in pretrained language
  models.
\newblock In \emph{Proceedings of the 59th Annual Meeting of the Association
  for Computational Linguistics and the 11th International Joint Conference on
  Natural Language Processing (Volume 1: Long Papers)}, pages 5356--5371,
  Online, August 2021. Association for Computational Linguistics.
\newblock \doi{10.18653/v1/2021.acl-long.416}.
\newblock URL \url{https://aclanthology.org/2021.acl-long.416}.

\bibitem[Nangia et~al.(2020)Nangia, Vania, Bhalerao, and Bowman]{crows}
Nikita Nangia, Clara Vania, Rasika Bhalerao, and Samuel~R. Bowman.
\newblock {C}row{S}-pairs: A challenge dataset for measuring social biases in
  masked language models.
\newblock In \emph{Proceedings of the 2020 Conference on Empirical Methods in
  Natural Language Processing (EMNLP)}, pages 1953--1967, Online, November
  2020. Association for Computational Linguistics.
\newblock \doi{10.18653/v1/2020.emnlp-main.154}.
\newblock URL \url{https://aclanthology.org/2020.emnlp-main.154}.

\bibitem[Nentidis et~al.(2021)Nentidis, Katsimpras, Vandorou, Krithara,
  Gasc{\'{o}}, Krallinger, and Paliouras]{BioASQ}
Anastasios Nentidis, Georgios Katsimpras, Eirini Vandorou, Anastasia Krithara,
  Luis Gasc{\'{o}}, Martin Krallinger, and Georgios Paliouras.
\newblock Overview of bioasq 2021: The ninth bioasq challenge on large-scale
  biomedical semantic indexing and question answering.
\newblock \emph{CoRR}, abs/2106.14885, 2021.
\newblock URL \url{https://arxiv.org/abs/2106.14885}.

\bibitem[Nieschlag et~al.(2010)Nieschlag, Behre, and Nieschlag]{andrology}
E~Nieschlag, HM~Behre, and S~Nieschlag.
\newblock Andrology: Male reproductive health and dysfunction, 2010.

\bibitem[Nijkamp et~al.(2022)Nijkamp, Ruffolo, Weinstein, Naik, and
  Madani]{ProGen2}
Erik Nijkamp, Jeffrey Ruffolo, Eli~N. Weinstein, Nikhil Naik, and Ali Madani.
\newblock Progen2: Exploring the boundaries of protein language models, 2022.
\newblock URL \url{https://arxiv.org/abs/2206.13517}.

\bibitem[Pafilis et~al.(2013)Pafilis, Frankild, Fanini, Faulwetter, Pavloudi,
  Vasileiadou, Arvanitidis, and Jensen]{S800}
Evangelos Pafilis, Sune~P Frankild, Lucia Fanini, Sarah Faulwetter, Christina
  Pavloudi, Aikaterini Vasileiadou, Christos Arvanitidis, and Lars~Juhl Jensen.
\newblock The species and organisms resources for fast and accurate
  identification of taxonomic names in text.
\newblock \emph{PloS one}, 8(6), 2013.

\bibitem[Pal et~al.(2022)Pal, Umapathi, and Sankarasubbu]{MedMCQA}
Ankit Pal, Logesh~Kumar Umapathi, and Malaikannan Sankarasubbu.
\newblock Medmcqa : A large-scale multi-subject multi-choice dataset for
  medical domain question answering.
\newblock 2022.
\newblock \doi{10.48550/ARXIV.2203.14371}.
\newblock URL \url{https://arxiv.org/abs/2203.14371}.

\bibitem[Petroni et~al.(2019)Petroni, Rockt{\"{a}}schel, Miller, Lewis,
  Bakhtin, Wu, and Riedel]{petroni2019language}
F.~Petroni, T.~Rockt{\"{a}}schel, A.H. Miller, P.~Lewis, A.~Bakhtin, Y.~Wu, and
  S.~Riedel.
\newblock Language models as knowledge bases?
\newblock In \emph{Proceedings of the 2019 Conference on Empirical Methods in
  Natural Language Processing (EMNLP), 2019}, 2019.

\bibitem[Press et~al.(2021)Press, Smith, and Lewis]{ALiBi}
Ofir Press, Noah~A. Smith, and Mike Lewis.
\newblock Train short, test long: Attention with linear biases enables input
  length extrapolation.
\newblock \emph{CoRR}, abs/2108.12409, 2021.
\newblock URL \url{https://arxiv.org/abs/2108.12409}.

\bibitem[Rae et~al.(2021)Rae, Borgeaud, Cai, Millican, Hoffmann, Song,
  Aslanides, Henderson, Ring, Young, Rutherford, Hennigan, Menick, Cassirer,
  Powell, van~den Driessche, Hendricks, Rauh, Huang, Glaese, Welbl, Dathathri,
  Huang, Uesato, Mellor, Higgins, Creswell, McAleese, Wu, Elsen, Jayakumar,
  Buchatskaya, Budden, Sutherland, Simonyan, Paganini, Sifre, Martens, Li,
  Kuncoro, Nematzadeh, Gribovskaya, Donato, Lazaridou, Mensch, Lespiau,
  Tsimpoukelli, Grigorev, Fritz, Sottiaux, Pajarskas, Pohlen, Gong, Toyama,
  de~Masson~d'Autume, Li, Terzi, Mikulik, Babuschkin, Clark, de~Las~Casas, Guy,
  Jones, Bradbury, Johnson, Hechtman, Weidinger, Gabriel, Isaac, Lockhart,
  Osindero, Rimell, Dyer, Vinyals, Ayoub, Stanway, Bennett, Hassabis,
  Kavukcuoglu, and Irving]{Gopher}
Jack~W. Rae, Sebastian Borgeaud, Trevor Cai, Katie Millican, Jordan Hoffmann,
  H.~Francis Song, John Aslanides, Sarah Henderson, Roman Ring, Susannah Young,
  Eliza Rutherford, Tom Hennigan, Jacob Menick, Albin Cassirer, Richard Powell,
  George van~den Driessche, Lisa~Anne Hendricks, Maribeth Rauh, Po{-}Sen Huang,
  Amelia Glaese, Johannes Welbl, Sumanth Dathathri, Saffron Huang, Jonathan
  Uesato, John Mellor, Irina Higgins, Antonia Creswell, Nat McAleese, Amy Wu,
  Erich Elsen, Siddhant~M. Jayakumar, Elena Buchatskaya, David Budden, Esme
  Sutherland, Karen Simonyan, Michela Paganini, Laurent Sifre, Lena Martens,
  Xiang~Lorraine Li, Adhiguna Kuncoro, Aida Nematzadeh, Elena Gribovskaya,
  Domenic Donato, Angeliki Lazaridou, Arthur Mensch, Jean{-}Baptiste Lespiau,
  Maria Tsimpoukelli, Nikolai Grigorev, Doug Fritz, Thibault Sottiaux, Mantas
  Pajarskas, Toby Pohlen, Zhitao Gong, Daniel Toyama, Cyprien
  de~Masson~d'Autume, Yujia Li, Tayfun Terzi, Vladimir Mikulik, Igor
  Babuschkin, Aidan Clark, Diego de~Las~Casas, Aurelia Guy, Chris Jones, James
  Bradbury, Matthew Johnson, Blake~A. Hechtman, Laura Weidinger, Iason Gabriel,
  William~S. Isaac, Edward Lockhart, Simon Osindero, Laura Rimell, Chris Dyer,
  Oriol Vinyals, Kareem Ayoub, Jeff Stanway, Lorrayne Bennett, Demis Hassabis,
  Koray Kavukcuoglu, and Geoffrey Irving.
\newblock Scaling language models: Methods, analysis {\&} insights from
  training gopher.
\newblock \emph{CoRR}, abs/2112.11446, 2021.
\newblock URL \url{https://arxiv.org/abs/2112.11446}.

\bibitem[Raffel et~al.(2020)Raffel, Shazeer, Roberts, Lee, Narang, Matena,
  Zhou, Li, and Liu]{2020t5}
Colin Raffel, Noam Shazeer, Adam Roberts, Katherine Lee, Sharan Narang, Michael
  Matena, Yanqi Zhou, Wei Li, and Peter~J. Liu.
\newblock Exploring the limits of transfer learning with a unified text-to-text
  transformer.
\newblock \emph{Journal of Machine Learning Research}, 21\penalty0
  (140):\penalty0 1--67, 2020.
\newblock URL \url{http://jmlr.org/papers/v21/20-074.html}.

\bibitem[Rajan et~al.(2021)Rajan, Zielesny, and Steinbeck]{STOUT}
K~Rajan, A~Zielesny, and C.~Steinbeck.
\newblock Stout: Smiles to iupac names using neural machine translation, 2021.
\newblock URL
  \url{https://jcheminf.biomedcentral.com/articles/10.1186/s13321-021-00512-4}.

\bibitem[Ramsundar et~al.(2019)Ramsundar, Eastman, Walters, Pande, Leswing, and
  Wu]{Ramsundaretal}
Bharath Ramsundar, Peter Eastman, Patrick Walters, Vijay Pande, Karl Leswing,
  and Zhenqin Wu.
\newblock \emph{Deep Learning for the Life Sciences}.
\newblock O'Reilly Media, 2019.
\newblock
  \url{https://www.amazon.com/Deep-Learning-Life-Sciences-Microscopy/dp/1492039837}.

\bibitem[Razeghi et~al.(2022)Razeghi, Logan, Gardner, and
  Singh]{PretrainingFrequency}
Yasaman Razeghi, Robert~L. Logan, Matt Gardner, and Sameer Singh.
\newblock Impact of pretraining term frequencies on few-shot reasoning, 2022.
\newblock URL \url{https://arxiv.org/abs/2202.07206}.

\bibitem[Rives et~al.(2021)Rives, Meier, Sercu, Goyal, Lin, Liu, Guo, Ott,
  Zitnick, Ma, and Fergus]{BioEmerge}
Alexander Rives, Joshua Meier, Tom Sercu, Siddharth Goyal, Zeming Lin, Jason
  Liu, Demi Guo, Myle Ott, C.~Lawrence Zitnick, Jerry Ma, and Rob Fergus.
\newblock Biological structure and function emerge from scaling unsupervised
  learning to 250 million protein sequences.
\newblock \emph{Proceedings of the National Academy of Sciences}, 118\penalty0
  (15):\penalty0 e2016239118, 2021.
\newblock \doi{10.1073/pnas.2016239118}.
\newblock URL \url{https://www.pnas.org/doi/abs/10.1073/pnas.2016239118}.

\bibitem[Ross et~al.(2021)Ross, Belgodere, Chenthamarakshan, Padhi, Mroueh, and
  Das]{MoLformer}
Jerret Ross, Brian Belgodere, Vijil Chenthamarakshan, Inkit Padhi, Youssef
  Mroueh, and Payel Das.
\newblock Do large scale molecular language representations capture important
  structural information?
\newblock \emph{CoRR}, abs/2106.09553, 2021.
\newblock URL \url{https://arxiv.org/abs/2106.09553}.

\bibitem[Sanh et~al.(2021)Sanh, Webson, Raffel, Bach, Sutawika, Alyafeai,
  Chaffin, Stiegler, Scao, Raja, Dey, Bari, Xu, Thakker, Sharma, Szczechla,
  Kim, Chhablani, Nayak, Datta, Chang, Jiang, Wang, Manica, Shen, Yong, Pandey,
  Bawden, Wang, Neeraj, Rozen, Sharma, Santilli, Fevry, Fries, Teehan, Bers,
  Biderman, Gao, Wolf, and Rush]{T0}
Victor Sanh, Albert Webson, Colin Raffel, Stephen~H. Bach, Lintang Sutawika,
  Zaid Alyafeai, Antoine Chaffin, Arnaud Stiegler, Teven~Le Scao, Arun Raja,
  Manan Dey, M~Saiful Bari, Canwen Xu, Urmish Thakker, Shanya~Sharma Sharma,
  Eliza Szczechla, Taewoon Kim, Gunjan Chhablani, Nihal Nayak, Debajyoti Datta,
  Jonathan Chang, Mike Tian-Jian Jiang, Han Wang, Matteo Manica, Sheng Shen,
  Zheng~Xin Yong, Harshit Pandey, Rachel Bawden, Thomas Wang, Trishala Neeraj,
  Jos Rozen, Abheesht Sharma, Andrea Santilli, Thibault Fevry, Jason~Alan
  Fries, Ryan Teehan, Tali Bers, Stella Biderman, Leo Gao, Thomas Wolf, and
  Alexander~M. Rush.
\newblock Multitask prompted training enables zero-shot task generalization,
  2021.
\newblock URL \url{https://arxiv.org/abs/2110.08207}.

\bibitem[Scialom et~al.(2022)Scialom, Chakrabarty, and Muresan]{ContinualT0}
Thomas Scialom, Tuhin Chakrabarty, and Smaranda Muresan.
\newblock Continual-t0: Progressively instructing 50+ tasks to language models
  without forgetting, 2022.
\newblock URL \url{https://arxiv.org/abs/2205.12393}.

\bibitem[Sennrich et~al.(2015)Sennrich, Haddow, and Birch]{BPE}
Rico Sennrich, Barry Haddow, and Alexandra Birch.
\newblock Neural machine translation of rare words with subword units.
\newblock \emph{CoRR}, abs/1508.07909, 2015.
\newblock URL \url{http://arxiv.org/abs/1508.07909}.

\bibitem[Sheng et~al.(2019)Sheng, Chang, Natarajan, and Peng]{WomanBabysitter}
Emily Sheng, Kai{-}Wei Chang, Premkumar Natarajan, and Nanyun Peng.
\newblock The woman worked as a babysitter: On biases in language generation.
\newblock \emph{CoRR}, abs/1909.01326, 2019.
\newblock URL \url{http://arxiv.org/abs/1909.01326}.

\bibitem[Sheng et~al.(2021)Sheng, Chang, Natarajan, and Peng]{SocietalBiases}
Emily Sheng, Kai{-}Wei Chang, Premkumar Natarajan, and Nanyun Peng.
\newblock Societal biases in language generation: Progress and challenges.
\newblock \emph{CoRR}, abs/2105.04054, 2021.
\newblock URL \url{https://arxiv.org/abs/2105.04054}.

\bibitem[Shin et~al.(2020)Shin, Zhang, Bakhturina, Puri, Patwary, Shoeybi, and
  Mani]{BioMegatron}
Hoo{-}Chang Shin, Yang Zhang, Evelina Bakhturina, Raul Puri, Mostofa Patwary,
  Mohammad Shoeybi, and Raghav Mani.
\newblock Biomegatron: Larger biomedical domain language model.
\newblock \emph{CoRR}, abs/2010.06060, 2020.
\newblock URL \url{https://arxiv.org/abs/2010.06060}.

\bibitem[Smith et~al.(2008)Smith, Tanabe, nee Ando, Kuo, Chung, Hsu, Lin,
  Klinger, Friedrich, Ganchev, Torii, Liu, Haddow, Struble, Povinelli, Vlachos,
  Jr, Hunter, Carpenter, Tsai, Dai, Liu, Chen, Sun, Katrenko, Adriaans,
  Blaschke, Torres, Neves, Nakov, Divoli, Maña-López, Mata, and
  Wilbur]{BC2GM}
Larry Smith, Lorraine~K Tanabe, Rie~Johnson nee Ando, Cheng-Ju Kuo, I-Fang
  Chung, Chun-Nan Hsu, Yu-Shi Lin, Roman Klinger, Christoph~M Friedrich, Kuzman
  Ganchev, Manabu Torii, Hongfang Liu, Barry Haddow, Craig~A Struble, Richard~J
  Povinelli, Andreas Vlachos, William A~Baumgartner Jr, Lawrence Hunter, Bob
  Carpenter, Richard Tzong-Han Tsai, Hong-Jie Dai, Feng Liu, Yifei Chen,
  Chengjie Sun, Sophia Katrenko, Pieter Adriaans, Christian Blaschke, Rafael
  Torres, Mariana Neves, Preslav Nakov, Anna Divoli, Manuel Maña-López,
  Jacinto Mata, and W~John Wilbur.
\newblock Overview of biocreative ii gene mention recognition.
\newblock \emph{Genome Biology}, 9, 2008.

\bibitem[Srivastava et~al.(2022)Srivastava, Rastogi, Rao, Shoeb, Abid, Fisch,
  Brown, Santoro, Gupta, Garriga-Alonso, Kluska, Lewkowycz, Agarwal, Power,
  Ray, Warstadt, Kocurek, Safaya, Tazarv, Xiang, Parrish, Nie, Hussain, Askell,
  Dsouza, Slone, Rahane, Iyer, Andreassen, Madotto, Santilli, Stuhlmüller,
  Dai, La, Lampinen, Zou, Jiang, Chen, Vuong, Gupta, Gottardi, Norelli,
  Venkatesh, Gholamidavoodi, Tabassum, Menezes, Kirubarajan, Mullokandov,
  Sabharwal, Herrick, Efrat, Erdem, Karakaş, Roberts, Loe, Zoph, Bojanowski,
  Özyurt, Hedayatnia, Neyshabur, Inden, Stein, Ekmekci, Lin, Howald, Diao,
  Dour, Stinson, Argueta, Ramírez, Singh, Rathkopf, Meng, Baral, Wu,
  Callison-Burch, Waites, Voigt, Manning, Potts, Ramirez, Rivera, Siro, Raffel,
  Ashcraft, Garbacea, Sileo, Garrette, Hendrycks, Kilman, Roth, Freeman,
  Khashabi, Levy, González, Perszyk, Hernandez, Chen, Ippolito, Gilboa, Dohan,
  Drakard, Jurgens, Datta, Ganguli, Emelin, Kleyko, Yuret, Chen, Tam, Hupkes,
  Misra, Buzan, Mollo, Yang, Lee, Shutova, Cubuk, Segal, Hagerman, Barnes,
  Donoway, Pavlick, Rodola, Lam, Chu, Tang, Erdem, Chang, Chi, Dyer, Jerzak,
  Kim, Manyasi, Zheltonozhskii, Xia, Siar, Martínez-Plumed, Happé, Chollet,
  Rong, Mishra, Winata, de~Melo, Kruszewski, Parascandolo, Mariani, Wang,
  Jaimovitch-López, Betz, Gur-Ari, Galijasevic, Kim, Rashkin, Hajishirzi,
  Mehta, Bogar, Shevlin, Schütze, Yakura, Zhang, Wong, Ng, Noble, Jumelet,
  Geissinger, Kernion, Hilton, Lee, Fisac, Simon, Koppel, Zheng, Zou, Kocoń,
  Thompson, Kaplan, Radom, Sohl-Dickstein, Phang, Wei, Yosinski, Novikova,
  Bosscher, Marsh, Kim, Taal, Engel, Alabi, Xu, Song, Tang, Waweru, Burden,
  Miller, Balis, Berant, Frohberg, Rozen, Hernandez-Orallo, Boudeman, Jones,
  Tenenbaum, Rule, Chua, Kanclerz, Livescu, Krauth, Gopalakrishnan, Ignatyeva,
  Markert, Dhole, Gimpel, Omondi, Mathewson, Chiafullo, Shkaruta, Shridhar,
  McDonell, Richardson, Reynolds, Gao, Zhang, Dugan, Qin, Contreras-Ochando,
  Morency, Moschella, Lam, Noble, Schmidt, He, Colón, Metz, Şenel, Bosma,
  Sap, ter Hoeve, Farooqi, Faruqui, Mazeika, Baturan, Marelli, Maru, Quintana,
  Tolkiehn, Giulianelli, Lewis, Potthast, Leavitt, Hagen, Schubert,
  Baitemirova, Arnaud, McElrath, Yee, Cohen, Gu, Ivanitskiy, Starritt, Strube,
  Swędrowski, Bevilacqua, Yasunaga, Kale, Cain, Xu, Suzgun, Tiwari, Bansal,
  Aminnaseri, Geva, Gheini, T, Peng, Chi, Lee, Krakover, Cameron, Roberts,
  Doiron, Nangia, Deckers, Muennighoff, Keskar, Iyer, Constant, Fiedel, Wen,
  Zhang, Agha, Elbaghdadi, Levy, Evans, Casares, Doshi, Fung, Liang, Vicol,
  Alipoormolabashi, Liao, Liang, Chang, Eckersley, Htut, Hwang, Miłkowski,
  Patil, Pezeshkpour, Oli, Mei, Lyu, Chen, Banjade, Rudolph, Gabriel, Habacker,
  Delgado, Millière, Garg, Barnes, Saurous, Arakawa, Raymaekers, Frank,
  Sikand, Novak, Sitelew, LeBras, Liu, Jacobs, Zhang, Salakhutdinov, Chi, Lee,
  Stovall, Teehan, Yang, Singh, Mohammad, Anand, Dillavou, Shleifer, Wiseman,
  Gruetter, Bowman, Schoenholz, Han, Kwatra, Rous, Ghazarian, Ghosh, Casey,
  Bischoff, Gehrmann, Schuster, Sadeghi, Hamdan, Zhou, Srivastava, Shi, Singh,
  Asaadi, Gu, Pachchigar, Toshniwal, Upadhyay, Shyamolima, Shakeri, Thormeyer,
  Melzi, Reddy, Makini, Lee, Torene, Hatwar, Dehaene, Divic, Ermon, Biderman,
  Lin, Prasad, Piantadosi, Shieber, Misherghi, Kiritchenko, Mishra, Linzen,
  Schuster, Li, Yu, Ali, Hashimoto, Wu, Desbordes, Rothschild, Phan, Wang,
  Nkinyili, Schick, Kornev, Telleen-Lawton, Tunduny, Gerstenberg, Chang,
  Neeraj, Khot, Shultz, Shaham, Misra, Demberg, Nyamai, Raunak, Ramasesh,
  Prabhu, Padmakumar, Srikumar, Fedus, Saunders, Zhang, Vossen, Ren, Tong,
  Zhao, Wu, Shen, Yaghoobzadeh, Lakretz, Song, Bahri, Choi, Yang, Hao, Chen,
  Belinkov, Hou, Hou, Bai, Seid, Zhao, Wang, Wang, Wang, and
  Wu]{BIGBenchakasomanyauthorsitdoesntfitinthecontextwindow}
Aarohi Srivastava, Abhinav Rastogi, Abhishek Rao, Abu Awal~Md Shoeb, Abubakar
  Abid, Adam Fisch, Adam~R. Brown, Adam Santoro, Aditya Gupta, Adrià
  Garriga-Alonso, Agnieszka Kluska, Aitor Lewkowycz, Akshat Agarwal, Alethea
  Power, Alex Ray, Alex Warstadt, Alexander~W. Kocurek, Ali Safaya, Ali Tazarv,
  Alice Xiang, Alicia Parrish, Allen Nie, Aman Hussain, Amanda Askell, Amanda
  Dsouza, Ambrose Slone, Ameet Rahane, Anantharaman~S. Iyer, Anders Andreassen,
  Andrea Madotto, Andrea Santilli, Andreas Stuhlmüller, Andrew Dai, Andrew La,
  Andrew Lampinen, Andy Zou, Angela Jiang, Angelica Chen, Anh Vuong, Animesh
  Gupta, Anna Gottardi, Antonio Norelli, Anu Venkatesh, Arash Gholamidavoodi,
  Arfa Tabassum, Arul Menezes, Arun Kirubarajan, Asher Mullokandov, Ashish
  Sabharwal, Austin Herrick, Avia Efrat, Aykut Erdem, Ayla Karakaş, B.~Ryan
  Roberts, Bao~Sheng Loe, Barret Zoph, Bartłomiej Bojanowski, Batuhan Özyurt,
  Behnam Hedayatnia, Behnam Neyshabur, Benjamin Inden, Benno Stein, Berk
  Ekmekci, Bill~Yuchen Lin, Blake Howald, Cameron Diao, Cameron Dour, Catherine
  Stinson, Cedrick Argueta, César~Ferri Ramírez, Chandan Singh, Charles
  Rathkopf, Chenlin Meng, Chitta Baral, Chiyu Wu, Chris Callison-Burch, Chris
  Waites, Christian Voigt, Christopher~D. Manning, Christopher Potts, Cindy
  Ramirez, Clara~E. Rivera, Clemencia Siro, Colin Raffel, Courtney Ashcraft,
  Cristina Garbacea, Damien Sileo, Dan Garrette, Dan Hendrycks, Dan Kilman, Dan
  Roth, Daniel Freeman, Daniel Khashabi, Daniel Levy, Daniel~Moseguí
  González, Danielle Perszyk, Danny Hernandez, Danqi Chen, Daphne Ippolito,
  Dar Gilboa, David Dohan, David Drakard, David Jurgens, Debajyoti Datta, Deep
  Ganguli, Denis Emelin, Denis Kleyko, Deniz Yuret, Derek Chen, Derek Tam,
  Dieuwke Hupkes, Diganta Misra, Dilyar Buzan, Dimitri~Coelho Mollo, Diyi Yang,
  Dong-Ho Lee, Ekaterina Shutova, Ekin~Dogus Cubuk, Elad Segal, Eleanor
  Hagerman, Elizabeth Barnes, Elizabeth Donoway, Ellie Pavlick, Emanuele
  Rodola, Emma Lam, Eric Chu, Eric Tang, Erkut Erdem, Ernie Chang, Ethan~A.
  Chi, Ethan Dyer, Ethan Jerzak, Ethan Kim, Eunice~Engefu Manyasi, Evgenii
  Zheltonozhskii, Fanyue Xia, Fatemeh Siar, Fernando Martínez-Plumed,
  Francesca Happé, Francois Chollet, Frieda Rong, Gaurav Mishra, Genta~Indra
  Winata, Gerard de~Melo, Germán Kruszewski, Giambattista Parascandolo,
  Giorgio Mariani, Gloria Wang, Gonzalo Jaimovitch-López, Gregor Betz, Guy
  Gur-Ari, Hana Galijasevic, Hannah Kim, Hannah Rashkin, Hannaneh Hajishirzi,
  Harsh Mehta, Hayden Bogar, Henry Shevlin, Hinrich Schütze, Hiromu Yakura,
  Hongming Zhang, Hugh~Mee Wong, Ian Ng, Isaac Noble, Jaap Jumelet, Jack
  Geissinger, Jackson Kernion, Jacob Hilton, Jaehoon Lee, Jaime~Fernández
  Fisac, James~B. Simon, James Koppel, James Zheng, James Zou, Jan Kocoń, Jana
  Thompson, Jared Kaplan, Jarema Radom, Jascha Sohl-Dickstein, Jason Phang,
  Jason Wei, Jason Yosinski, Jekaterina Novikova, Jelle Bosscher, Jennifer
  Marsh, Jeremy Kim, Jeroen Taal, Jesse Engel, Jesujoba Alabi, Jiacheng Xu,
  Jiaming Song, Jillian Tang, Joan Waweru, John Burden, John Miller, John~U.
  Balis, Jonathan Berant, Jörg Frohberg, Jos Rozen, Jose Hernandez-Orallo,
  Joseph Boudeman, Joseph Jones, Joshua~B. Tenenbaum, Joshua~S. Rule, Joyce
  Chua, Kamil Kanclerz, Karen Livescu, Karl Krauth, Karthik Gopalakrishnan,
  Katerina Ignatyeva, Katja Markert, Kaustubh~D. Dhole, Kevin Gimpel, Kevin
  Omondi, Kory Mathewson, Kristen Chiafullo, Ksenia Shkaruta, Kumar Shridhar,
  Kyle McDonell, Kyle Richardson, Laria Reynolds, Leo Gao, Li~Zhang, Liam
  Dugan, Lianhui Qin, Lidia Contreras-Ochando, Louis-Philippe Morency, Luca
  Moschella, Lucas Lam, Lucy Noble, Ludwig Schmidt, Luheng He, Luis~Oliveros
  Colón, Luke Metz, Lütfi~Kerem Şenel, Maarten Bosma, Maarten Sap, Maartje
  ter Hoeve, Maheen Farooqi, Manaal Faruqui, Mantas Mazeika, Marco Baturan,
  Marco Marelli, Marco Maru, Maria Jose~Ramírez Quintana, Marie Tolkiehn,
  Mario Giulianelli, Martha Lewis, Martin Potthast, Matthew~L. Leavitt,
  Matthias Hagen, Mátyás Schubert, Medina~Orduna Baitemirova, Melody Arnaud,
  Melvin McElrath, Michael~A. Yee, Michael Cohen, Michael Gu, Michael
  Ivanitskiy, Michael Starritt, Michael Strube, Michał Swędrowski, Michele
  Bevilacqua, Michihiro Yasunaga, Mihir Kale, Mike Cain, Mimee Xu, Mirac
  Suzgun, Mo~Tiwari, Mohit Bansal, Moin Aminnaseri, Mor Geva, Mozhdeh Gheini,
  Mukund~Varma T, Nanyun Peng, Nathan Chi, Nayeon Lee, Neta Gur-Ari Krakover,
  Nicholas Cameron, Nicholas Roberts, Nick Doiron, Nikita Nangia, Niklas
  Deckers, Niklas Muennighoff, Nitish~Shirish Keskar, Niveditha~S. Iyer, Noah
  Constant, Noah Fiedel, Nuan Wen, Oliver Zhang, Omar Agha, Omar Elbaghdadi,
  Omer Levy, Owain Evans, Pablo Antonio~Moreno Casares, Parth Doshi, Pascale
  Fung, Paul~Pu Liang, Paul Vicol, Pegah Alipoormolabashi, Peiyuan Liao, Percy
  Liang, Peter Chang, Peter Eckersley, Phu~Mon Htut, Pinyu Hwang, Piotr
  Miłkowski, Piyush Patil, Pouya Pezeshkpour, Priti Oli, Qiaozhu Mei, Qing
  Lyu, Qinlang Chen, Rabin Banjade, Rachel~Etta Rudolph, Raefer Gabriel, Rahel
  Habacker, Ramón~Risco Delgado, Raphaël Millière, Rhythm Garg, Richard
  Barnes, Rif~A. Saurous, Riku Arakawa, Robbe Raymaekers, Robert Frank, Rohan
  Sikand, Roman Novak, Roman Sitelew, Ronan LeBras, Rosanne Liu, Rowan Jacobs,
  Rui Zhang, Ruslan Salakhutdinov, Ryan Chi, Ryan Lee, Ryan Stovall, Ryan
  Teehan, Rylan Yang, Sahib Singh, Saif~M. Mohammad, Sajant Anand, Sam
  Dillavou, Sam Shleifer, Sam Wiseman, Samuel Gruetter, Samuel~R. Bowman,
  Samuel~S. Schoenholz, Sanghyun Han, Sanjeev Kwatra, Sarah~A. Rous, Sarik
  Ghazarian, Sayan Ghosh, Sean Casey, Sebastian Bischoff, Sebastian Gehrmann,
  Sebastian Schuster, Sepideh Sadeghi, Shadi Hamdan, Sharon Zhou, Shashank
  Srivastava, Sherry Shi, Shikhar Singh, Shima Asaadi, Shixiang~Shane Gu, Shubh
  Pachchigar, Shubham Toshniwal, Shyam Upadhyay, Debnath Shyamolima, Siamak
  Shakeri, Simon Thormeyer, Simone Melzi, Siva Reddy, Sneha~Priscilla Makini,
  Soo-Hwan Lee, Spencer Torene, Sriharsha Hatwar, Stanislas Dehaene, Stefan
  Divic, Stefano Ermon, Stella Biderman, Stephanie Lin, Stephen Prasad,
  Steven~T. Piantadosi, Stuart~M. Shieber, Summer Misherghi, Svetlana
  Kiritchenko, Swaroop Mishra, Tal Linzen, Tal Schuster, Tao Li, Tao Yu, Tariq
  Ali, Tatsu Hashimoto, Te-Lin Wu, Théo Desbordes, Theodore Rothschild, Thomas
  Phan, Tianle Wang, Tiberius Nkinyili, Timo Schick, Timofei Kornev, Timothy
  Telleen-Lawton, Titus Tunduny, Tobias Gerstenberg, Trenton Chang, Trishala
  Neeraj, Tushar Khot, Tyler Shultz, Uri Shaham, Vedant Misra, Vera Demberg,
  Victoria Nyamai, Vikas Raunak, Vinay Ramasesh, Vinay~Uday Prabhu, Vishakh
  Padmakumar, Vivek Srikumar, William Fedus, William Saunders, William Zhang,
  Wout Vossen, Xiang Ren, Xiaoyu Tong, Xinran Zhao, Xinyi Wu, Xudong Shen,
  Yadollah Yaghoobzadeh, Yair Lakretz, Yangqiu Song, Yasaman Bahri, Yejin Choi,
  Yichi Yang, Yiding Hao, Yifu Chen, Yonatan Belinkov, Yu~Hou, Yufang Hou,
  Yuntao Bai, Zachary Seid, Zhuoye Zhao, Zijian Wang, Zijie~J. Wang, Zirui
  Wang, and Ziyi Wu.
\newblock Beyond the imitation game: Quantifying and extrapolating the
  capabilities of language models, 2022.
\newblock URL \url{https://arxiv.org/abs/2206.04615}.

\bibitem[Steinegger and S\"{o}ding(2017)]{MMseq2}
Martin Steinegger and Johannes S\"{o}ding.
\newblock {MMseqs}2 enables sensitive protein sequence searching for the
  analysis of massive data sets.
\newblock \emph{Nature Biotechnology}, 35\penalty0 (11):\penalty0 1026--1028,
  October 2017.
\newblock \doi{10.1038/nbt.3988}.
\newblock URL \url{https://doi.org/10.1038/nbt.3988}.

\bibitem[Suzgun et~al.(2022)Suzgun, Scales, Schärli, Gehrmann, Tay, Chung,
  Chowdhery, Le, Chi, Zhou, and Wei]{CoTBigBENCH}
Mirac Suzgun, Nathan Scales, Nathanael Schärli, Sebastian Gehrmann, Yi~Tay,
  Hyung~Won Chung, Aakanksha Chowdhery, Quoc~V. Le, Ed~H. Chi, Denny Zhou, and
  Jason Wei.
\newblock Challenging big-bench tasks and whether chain-of-thought can solve
  them, 2022.
\newblock URL \url{https://arxiv.org/abs/2210.09261}.

\bibitem[Taboureau et~al.(2011)Taboureau, Nielsen, Audouze, Weinhold,
  Edsg{\"a}rd, Roque, Kouskoumvekaki, Bora, Curpan, Jensen, Brunak, and
  Oprea]{ChemProt}
Olivier Taboureau, Sonny~Kim Nielsen, Karine Audouze, Nils Weinhold, Daniel
  Edsg{\"a}rd, Francisco~S Roque, Irene Kouskoumvekaki, Alina Bora, Ramona
  Curpan, Thomas~Sk{\o}t Jensen, S{\o}ren Brunak, and Tudor~I Oprea.
\newblock {ChemProt}: a disease chemical biology database.
\newblock \emph{Nucleic Acids Res.}, 39\penalty0 (Database issue):\penalty0
  D367--72, January 2011.

\bibitem[Talmor et~al.(2018)Talmor, Herzig, Lourie, and Berant]{CommonSenseQA}
Alon Talmor, Jonathan Herzig, Nicholas Lourie, and Jonathan Berant.
\newblock Commonsenseqa: {A} question answering challenge targeting commonsense
  knowledge.
\newblock \emph{CoRR}, abs/1811.00937, 2018.
\newblock URL \url{http://arxiv.org/abs/1811.00937}.

\bibitem[Tay et~al.(2022{\natexlab{a}})Tay, Dehghani, Abnar, Chung, Fedus, Rao,
  Narang, Tran, Yogatama, and Metzler]{ScalingLawsModelArch}
Yi~Tay, Mostafa Dehghani, Samira Abnar, Hyung~Won Chung, William Fedus, Jinfeng
  Rao, Sharan Narang, Vinh~Q. Tran, Dani Yogatama, and Donald Metzler.
\newblock Scaling laws vs model architectures: How does inductive bias
  influence scaling?, 2022{\natexlab{a}}.
\newblock URL \url{https://arxiv.org/abs/2207.10551}.

\bibitem[Tay et~al.(2022{\natexlab{b}})Tay, Wei, Chung, Tran, So, Shakeri,
  Garcia, Zheng, Rao, Chowdhery, Zhou, Metzler, Petrov, Houlsby, Le, and
  Dehghani]{UPALM}
Yi~Tay, Jason Wei, Hyung~Won Chung, Vinh~Q. Tran, David~R. So, Siamak Shakeri,
  Xavier Garcia, Huaixiu~Steven Zheng, Jinfeng Rao, Aakanksha Chowdhery, Denny
  Zhou, Donald Metzler, Slav Petrov, Neil Houlsby, Quoc~V. Le, and Mostafa
  Dehghani.
\newblock Transcending scaling laws with 0.1
  2022{\natexlab{b}}.
\newblock URL \url{https://arxiv.org/abs/2210.11399}.

\bibitem[Thoppilan et~al.(2022)Thoppilan, De~Freitas, Hall, Shazeer,
  Kulshreshtha, Cheng, Jin, Bos, Baker, Du, Li, Lee, Zheng, Ghafouri, Menegali,
  Huang, Krikun, Lepikhin, Qin, Chen, Xu, Chen, Roberts, Bosma, Zhao, Zhou,
  Chang, Krivokon, Rusch, Pickett, Srinivasan, Man, Meier-Hellstern, Morris,
  Doshi, Santos, Duke, Soraker, Zevenbergen, Prabhakaran, Diaz, Hutchinson,
  Olson, Molina, Hoffman-John, Lee, Aroyo, Rajakumar, Butryna, Lamm, Kuzmina,
  Fenton, Cohen, Bernstein, Kurzweil, Aguera-Arcas, Cui, Croak, Chi, and
  Le]{LaMBDA}
Romal Thoppilan, Daniel De~Freitas, Jamie Hall, Noam Shazeer, Apoorv
  Kulshreshtha, Heng-Tze Cheng, Alicia Jin, Taylor Bos, Leslie Baker, Yu~Du,
  YaGuang Li, Hongrae Lee, Huaixiu~Steven Zheng, Amin Ghafouri, Marcelo
  Menegali, Yanping Huang, Maxim Krikun, Dmitry Lepikhin, James Qin, Dehao
  Chen, Yuanzhong Xu, Zhifeng Chen, Adam Roberts, Maarten Bosma, Vincent Zhao,
  Yanqi Zhou, Chung-Ching Chang, Igor Krivokon, Will Rusch, Marc Pickett,
  Pranesh Srinivasan, Laichee Man, Kathleen Meier-Hellstern, Meredith~Ringel
  Morris, Tulsee Doshi, Renelito~Delos Santos, Toju Duke, Johnny Soraker, Ben
  Zevenbergen, Vinodkumar Prabhakaran, Mark Diaz, Ben Hutchinson, Kristen
  Olson, Alejandra Molina, Erin Hoffman-John, Josh Lee, Lora Aroyo, Ravi
  Rajakumar, Alena Butryna, Matthew Lamm, Viktoriya Kuzmina, Joe Fenton, Aaron
  Cohen, Rachel Bernstein, Ray Kurzweil, Blaise Aguera-Arcas, Claire Cui,
  Marian Croak, Ed~Chi, and Quoc Le.
\newblock Lamda: Language models for dialog applications, 2022.
\newblock URL \url{https://arxiv.org/abs/2201.08239}.

\bibitem[V et~al.(2021)V, Mohania, and Goyal]{QCScience}
Venktesh V, Mukesh~K. Mohania, and Vikram Goyal.
\newblock Tagrec: Automated tagging of questions with hierarchical learning
  taxonomy.
\newblock \emph{CoRR}, abs/2107.10649, 2021.
\newblock URL \url{https://arxiv.org/abs/2107.10649}.

\bibitem[Vaswani et~al.(2017)Vaswani, Shazeer, Parmar, Uszkoreit, Jones, Gomez,
  Kaiser, and Polosukhin]{VaswaniSPUJGKP17}
Ashish Vaswani, Noam Shazeer, Niki Parmar, Jakob Uszkoreit, Llion Jones,
  Aidan~N. Gomez, Lukasz Kaiser, and Illia Polosukhin.
\newblock Attention is all you need.
\newblock \emph{CoRR}, abs/1706.03762, 2017.
\newblock URL \url{http://arxiv.org/abs/1706.03762}.

\bibitem[Wei et~al.(2021)Wei, Bosma, Zhao, Guu, Yu, Lester, Du, Dai, and
  Le]{FLAN}
Jason Wei, Maarten Bosma, Vincent~Y. Zhao, Kelvin Guu, Adams~Wei Yu, Brian
  Lester, Nan Du, Andrew~M. Dai, and Quoc~V. Le.
\newblock Finetuned language models are zero-shot learners, 2021.
\newblock URL \url{https://arxiv.org/abs/2109.01652}.

\bibitem[Wei et~al.(2022)Wei, Wang, Schuurmans, Bosma, Ichter, Xia, Chi, Le,
  and Zhou]{ChainOfThought}
Jason Wei, Xuezhi Wang, Dale Schuurmans, Maarten Bosma, Brian Ichter, Fei Xia,
  Ed~Chi, Quoc Le, and Denny Zhou.
\newblock Chain of thought prompting elicits reasoning in large language
  models, 2022.
\newblock URL \url{https://arxiv.org/abs/2201.11903}.

\bibitem[Weininger(1988)]{SMILES}
David Weininger.
\newblock Smiles, a chemical language and information system. 1. introduction
  to methodology and encoding rules.
\newblock \emph{Journal of Chemical Information and Computer Sciences},
  28\penalty0 (1):\penalty0 31--36, 1988.
\newblock \doi{10.1021/ci00057a005}.
\newblock URL \url{https://doi.org/10.1021/ci00057a005}.

\bibitem[Welbl et~al.(2017)Welbl, Liu, and Gardner]{SciQ}
Johannes Welbl, Nelson~F. Liu, and Matt Gardner.
\newblock Crowdsourcing multiple choice science questions.
\newblock \emph{ArXiv}, abs/1707.06209, 2017.

\bibitem[Wheeler(1990)]{Wheeler}
John Wheeler.
\newblock Information, physics, quantum: The search for links.
\newblock \emph{Zurek, W.H., Ed., Complexity, Entropy, and the Physics of
  Information}, 1990.

\bibitem[Wigner(1959)]{Wigner}
Eugene Wigner.
\newblock The unreasonable effectiveness of mathematics in the natural
  sciences.
\newblock \emph{Communications on Pure and Applied Mathematics}, 1959.

\bibitem[Wu et~al.(2017)Wu, Ramsundar, Feinberg, Gomes, Geniesse, Pappu,
  Leswing, and Pande]{MoleculeNet}
Zhenqin Wu, Bharath Ramsundar, Evan~N. Feinberg, Joseph Gomes, Caleb Geniesse,
  Aneesh~S. Pappu, Karl Leswing, and Vijay Pande.
\newblock Moleculenet: A benchmark for molecular machine learning, 2017.
\newblock URL \url{https://arxiv.org/abs/1703.00564}.

\bibitem[Xu et~al.(2017)Xu, Liu, Gao, Shen, and Liu]{RACE}
Yichong Xu, Jingjing Liu, Jianfeng Gao, Yelong Shen, and Xiaodong Liu.
\newblock Towards human-level machine reading comprehension: Reasoning and
  inference with multiple strategies.
\newblock \emph{CoRR}, abs/1711.04964, 2017.
\newblock URL \url{http://arxiv.org/abs/1711.04964}.

\bibitem[Yasunaga et~al.(2022)Yasunaga, Leskovec, and Liang]{BioLinkBERT}
Michihiro Yasunaga, Jure Leskovec, and Percy Liang.
\newblock Linkbert: Pretraining language models with document links, 2022.
\newblock URL \url{https://arxiv.org/abs/2203.15827}.

\bibitem[Zhang et~al.(2022)Zhang, Roller, Goyal, Artetxe, Chen, Chen, Dewan,
  Diab, Li, Lin, Mihaylov, Ott, Shleifer, Shuster, Simig, Koura, Sridhar, Wang,
  and Zettlemoyer]{OPT}
Susan Zhang, Stephen Roller, Naman Goyal, Mikel Artetxe, Moya Chen, Shuohui
  Chen, Christopher Dewan, Mona Diab, Xian Li, Xi~Victoria Lin, Todor Mihaylov,
  Myle Ott, Sam Shleifer, Kurt Shuster, Daniel Simig, Punit~Singh Koura, Anjali
  Sridhar, Tianlu Wang, and Luke Zettlemoyer.
\newblock Opt: Open pre-trained transformer language models, 2022.
\newblock URL \url{https://arxiv.org/abs/2205.01068}.

\bibitem[Zhou et~al.(2022)Zhou, Gao, Ding, Zheng, Xu, Zhang, and Ke]{UniMol}
Gengmo Zhou, Zhifeng~Gao Gao, Qiankun Ding, Hang Zheng, Wei Xu, Hongteng,
  Linfeng Zhang, and Guolin Ke.
\newblock Uni-mol: A universal 3d molecular representation learning framework,
  2022.
\newblock URL
  \url{https://chemrxiv.org/engage/chemrxiv/article-details/628e5b4d5d948517f5ce6d72}.

\end{thebibliography}

\newpage
\appendix
\section{Appendix}

\subsection{Dataset Components}

We cover the various components of the corpus in this section.

\subsubsection{Papers}

We source scientific papers from preprint servers such as arXiv, PMC and other sources; see Table~\ref{table:paper-components}. 

We also use the Semantic Scholar full text dataset (S2) to capture the long tail of science~\citep{S2ORC}. We apply several quality filters, including excluding papers from journals with certain keywords, and also excluding papers with a low journal impact factor. Details of the filters we used are contained in the Appendix.

We source abstracts where full texts are not open access. In total the full dataset contains 48 million papers, abstract and full-text, up to July 2022.

\begin{table}[h!]
\begin{center}
\begin{tabular}{ lrr } 
\toprule
  Data source & Documents & Tokens     \\ 
\midrule
 arXiv & 2 million & 35 billion \\
 PMC & 3 million & 23 billion  \\ 
 Semantic Scholar & 3 million & 18 billion  \\
 PubMed Abstracts & 21 million & 5 billion  \\
 Semantic Scholar Abstracts & 19 million & 4 billion  \\
 bioRxiv & 128,059 & 1 billion \\
 OSF & 54,905 & 428 million  \\
 medRxiv & 24,019 & 176 million  \\ 
 ACL & 25,518 & 150 million        \\ 
 PubAg Abstracts & 308,235 & 105 million  \\
 ChemRxiv & 7,617 & 67 million  \\ 
\midrule
\textbf{Total} & 48 million & 88 billion \\
\bottomrule
\end{tabular}
\end{center}
\caption{Paper sources used in our corpus}
\label{table:paper-components}
\end{table}

We use a modified version of the GROBID library for converting PDFs to text, as well as obtaining titles, authors and citations~\citep{GROBID}. Where mathematical LaTeX is available, for example in arXiv, we make sure to combine the GROBID results with LaTeX source to recover mathematical content.

The final paper documents are stored in a markdown format, as opposed to full LaTeX. We use markdown as the standard format for all documents in the corpus to support knowledge blending between sources. Papers are citation processed, following the title-based approach of Section 2.2.

\subsubsection{Reference Material}

We source encyclopedias, textbooks and educational material to create a base of reference material that the model can learn from. The details are covered in Table~\ref{table:reference-components}.

\begin{table}[h!]
\begin{center}
\begin{tabular}{ lrr } 
\toprule
  Data source & Documents & Tokens     \\ 
\midrule
 Wikipedia & 6 million & 5 billion \\
 StackExchange & 1.6 million & 1 billion \\
 LibreText & 95,113 & 185 million  \\
 Wikibooks & 74,705 & 110 million  \\ 
 Open Textbooks & 647 & 94 million  \\
 MIT OCW & 25,640 & 90 million  \\
 Wikiversity & 38,138 & 52 million  \\
 ProofWiki & 32,389 & 12 million \\ 
 Khan Academy & 3,075 & 7 million \\ 
 Papers with Code & 13,430 & 4 million \\ 
 IUPAC Goldbook & 6,788 & 1 million  \\
\midrule
\textbf{Total} & 8 million & 7 billion \\
\bottomrule
\end{tabular}
\end{center}
\caption{Reference material used in our corpus}
\label{table:reference-components}
\end{table}

We apply source specific processing for several of the datasets, specifically:

\begin{itemize}
    \item For \textit{StackExchange}, we take questions from scientific sites; see the Appendix for the subset used.
    \item For \textit{Papers with Code} and \textit{IUPAC Goldbook} we apply data augmentation in the form of prompt randomization. Sometimes we pose sections as questions/answers; for example a section explaining a machine learning method is sometimes posed as "Question: What is [method]?". 
    \item  For \textit{KhanAcademy articles}, we add \verb|<work>| tokens for step-by-step reasoning examples, which we explain shortly in Section 2.4.
\end{itemize}

We make an effort to preserve mathematical LaTeX and capture citations, including hyperlinks to papers.

\subsubsection{Knowledge Bases}

We source fine-grained knowledge from scientific knowledge bases. The details are covered in Table~\ref{table:kb-components}.

\begin{table}[h!]
\begin{center}
\begin{tabular}{ lrr } 
\toprule
  Data source & Documents & Tokens     \\ 
\midrule
 PubChem Compound & 1.7 million & 1 billion  \\ 
 UniProt & 551,837 & 0.6 billion  \\
 RefSeq Genome & 69 & 0.1 billion \\
 OEIS & 350,833 & 0.07 billion \\
 Ribosome & 9,950 & 0.05 billion \\
 LIPID MAPS & 45,273 & 0.03 billion \\
 Reactome & 156 & 0.01 billion \\
 NASA Exoplanet & 5,021 & 0.01 billion  \\
 \midrule
\textbf{Total} & 2 million & 2 billion \\
\bottomrule
\end{tabular}
\end{center}
\caption{Knowledge bases used in our corpus}
\label{table:kb-components}
\end{table}

For the chemistry and biology datasets, we wrap modalities like SMILES and protein sequences with their specialized tokens (see Section 2.1). For UniProt we apply data augmentation to the document format:

\begin{itemize}
    \item \textbf{Order Randomization} - with probability \(0.5\) the protein sequence starts at beginning of the document, else the end of document. This ensures we can learn from $\text{seq} \rightarrow \text{property}$ and $\text{property} \rightarrow \text{seq}$.
    \item \textbf{Format Randomization} - with probability \(\frac{1}{3}\) we replace a description, e.g. "The function of protein is...", with a Q\&A, e.g. "Question: What is the function of the protein? Answer: The function is...".
\end{itemize}

For \textit{NASA Exoplanet} we apply order randomization to the exoplanet characteristics.

For chemical and biological sequences, we take a small subset of available entities. This is to ensure the model is not overly biased towards learning natural sequences over natural language. Specifically:

\begin{itemize}
    \item For \textit{PubChem Compound}, we take a small, random subset ($2$ million) of total compounds ($110$ million).
    \item For \textit{UniProt}, we take reviewed Swiss-Prot proteins; a small subset ($0.5$ million) of total ($227$ million).
    \item For \textit{RefSeq Genome}, we take reference sequences, which is a small subset of available nucleotide sequences. For the human genome, we only include the protein-coding genes.
\end{itemize}

This is a constraint we can relax in future work, enabling for much larger corpus. In this work, we focus on the first step of investigating whether a single model can learn effectively in this multi-modal setting.

\subsubsection{Common Crawl}

We source academic and scientific content via a highly-filtered subset of CommonCrawl. The details are covered in Table ~\ref{table:cc-components}.

\begin{table}[h!]
\begin{center}
\begin{tabular}{ lrr } 
\toprule
  Data source & Documents & Tokens     \\ 
\midrule
 ScientificCC & 0.8 million & 0.7 billion  \\ 
 AcademicCC & 0.05 million & 0.4 billion  \\
 \midrule
\textbf{Total} & 0.9 million & 1.1 billion \\
\bottomrule
\end{tabular}
\end{center}
\caption{CommonCrawl material used in our corpus}
\label{table:cc-components}
\end{table}

For \textit{Scientific Common Crawl}, we train a fasttext classifier to identify Common Crawl webpages with scientific content~\citep{joulin2016bag} using a noisy set of 600 domains. We then manually annotated the domains predicted by fasttext as scientific to assemble a list of 200 high-quality scientific and reference domains.

For \textit{Academic Common Crawl}, we assemble a list of academic domains, such as university websites. We take PDFs from these domains, based on the Common Crawl index, and process these using GROBID. 

We do not LaTeX-process pages from these sources.

We found the quality of extracted text in CommonCrawl generally quite poor, which is why we applied stringent filters. We suspect this could be an important area for future work in order to capture more base scientific knowledge.

\subsubsection{Code}

We source academic GitHub repositories from the \textit{Papers with Code} index for machine learning, physics, mathematics, statistics and astronomy. The index does not explicitly cover sciences such as biology and chemistry, but many of these repositories are captured as part of the general machine learning index. We exclude repositories that do not have a license or copyright file.

\subsubsection{<work> Datasets}

For \textit{KhanProblems}, we used the problems from AMPS and converted to a <work> format~\citep{MATH}. Where possible we tried to include more tedious steps to reduce errors from a single pass, but this annotation was fairly incomplete and we suspect bigger gains are possible with more cleaning.

For \textit{GSM8k} we use the provided training dataset and convert so the calculator steps are performed by writing a Python program, following the <work> format \citep{GSM8k}. In general, we found when the model went into this prompt style, it was more error-prone. We think this is because the prompt style made the model write too many programs within <work>, rather than getting things ready to run in a single program. In general we found longer <work> answers led to a higher chance of a mistake on the reasoning path.

For \textit{OneSmallStep}, we made 50 problem set question templates, and randomized the variables in the problem to get more prompt examples. We summarize the fields we made prompts below.

\begin{table}[h]
\vspace{20px}
\begin{center}
\begin{tabular}{ lr } 
\toprule
  Field & Templates     \\ 
\midrule
Astronomy & 2 \\
Chemistry & 7 \\ 
Electronics & 10 \\
Mathematics & 15 \\
Physics & 14 \\
Statistics & 2 \\
\midrule
\textbf{Total} & 50 \\
\bottomrule
\end{tabular}
\end{center}
\end{table}

As we can see the diversity was not very large, and so further gains are likely with more annotation.

Lastly we wrote 921 examples, based off internet examples, in a <work> format for \textit{Workout}. This was our highest quality dataset, and had reasonable diversity across fields: mathematics, chemistry, biology, astronomy, physics, geology, history. This is the type of dataset we would look to scale in future work.

\subsection{Dataset Deduplication}

We use the following procedure for deduplicating the corpus:

\begin{itemize} 
 \item We identify identical spans of 100 bytes or more (of utf-8 text) across the whole corpus, except for some explicitly excluded data sources. We do this using the repository from \citet{lee2021deduplicating}.
 \item We process corpus files in a predetermined order to prioritize some sources. From a set of spans representing the exact same content across files, we remove the span in the first file. If the same content repeats across a single file and it was not found in the files before, all its occurrences are kept.
 \item We merge duplicated spans separated by at most 4 bytes.
 \item We narrow down the resulting spans to paragraph boundaries (i.e. "\verb|\n\n|").
 \item We remove the content from files corresponding to the spans.
 \end{itemize}

\subsection{Citation Identifier Ablations}

We report ablations for the citation identifier ablations below, where we test title-based identifiers versus alphanumeric identifiers.

Specifically, we set up an evaluation set of dataset and method names from \textit{Papers with Code}. The task is to predict the citation given the method or dataset name, e.g. \verb|ResNet [START_REF]|, where the target is \verb|Deep Residual Learning for Image Recognition, He|. We train a 6.7bn model on both types of processing for the ablation. Method and dataset results are shown below.

\begin{table}[h!]
\begin{center}
\begin{tabular}{ crrrrrr } 
\toprule
    \multicolumn{1}{c}{} & \multicolumn{6}{c}{Citation Processing} \\
    \multicolumn{1}{c}{} & \multicolumn{3}{c}{(a) Titles}&\multicolumn{3}{c}{(b) IDs} \\

\midrule
  Method citations & Correct & Hallucinated & Incorrect & Correct & Hallucinated & Incorrect      \\ 
\midrule
\(k = 1\) & \textbf{13.8\%} & 54.5\% & 31.7\% & 1.8\% & 3.5\% & 94.7\%\\
2 $\leq k < 5$ & \textbf{30.4\%} & 38.6\% & 31.1\% & 9.3\% & 4.0\% & 86.7\%\\
5 $\leq k < 10$ & \textbf{36.3\%} & 29.5\% & 34.2\% & 17.9\% & 0.0\% & 82.1\%\\
10 $\leq k < 25$ & \textbf{43.0\%} & 15.8\% & 41.2\% & 38.8\% & 3.0\% & 58.2\%\\
25 $\leq k < 50$ & \textbf{53.4\%} & 8.7\% & 37.9\% & 43.7\% & 0.0\% & 56.3\%\\
50 $\leq k < 100$ & \textbf{64.8\%} & 9.9\% & 25.3\% & 60.6\% & 1.4\% & 38.0\%\\
100 $\leq k < 500$ & \textbf{64.6\%} & 8.3\% & 27.1\% & 63.5\% & 1.0\% & 35.4\%\\
$\geq 500$ & \textbf{78.6\%} & 0.0\% & 21.4\% & 78.6\% & 0.0\% & 21.4\%\\

\bottomrule
\end{tabular}
\end{center}
\caption{\textbf{Citation Processing Ablation}. We predict citations for the \textit{PWC Methods} dataset using 6.7 billion size models. Papers are bucketed according to the number of citations (mentions) in the dataset. The title processing model has a higher accuracy, but greater risk of hallucination. There are 1,705 methods in this evaluation dataset.}
\label{table:citation-results-methods}
\end{table}

\begin{table}[h!]
\begin{center}
\begin{tabular}{ crrrrrr } 
\toprule
    \multicolumn{1}{c}{} & \multicolumn{6}{c}{Citation Processing} \\
    \multicolumn{1}{c}{} & \multicolumn{3}{c}{(a) Titles}&\multicolumn{3}{c}{(b) IDs} \\

\midrule
  Dataset citations & Correct & Hallucinated & Incorrect & Correct & Hallucinated & Incorrect      \\ 
\midrule
\(k = 1\) & \textbf{1.4\%} & 62.5\% & 36.1\% & 0.5\% & 11.5\% & 88.1\%\\
\(2 \leq k < 5\) & \textbf{5.0\%} & 59.2\% & 35.8\% & 0.6\% & 10.2\% & 89.2\%\\
\(5 \leq k < 10\) & \textbf{15.4\%} & 49.7\% & 34.8\% & 2.6\% & 6.2\% & 91.1\%\\
\(10 \leq k < 25\) & \textbf{25.7\%} & 36.8\% & 37.5\% & 8.3\% & 4.8\% & 86.9\%\\
\(25 \leq k < 50\) & \textbf{44.6\%} & 27.4\% & 28.0\% & 22.9\% & 7.0\% & 70.0\%\\
\(50 \leq k < 100\) & \textbf{58.6\%} & 17.7\% & 23.6\% & 41.4\% & 7.7\% & 50.9\%\\
\(100 \leq k < 500\) & \textbf{65.5\%} & 6.7\% & 27.8\% & 62.4\% & 3.1\% & 34.5\%\\
\(\geq 500\) & \textbf{81.8\%} & 6.1\% & 12.1\% & 81.8\% & 3.0\% & 15.2\%\\

\bottomrule
\end{tabular}
\end{center}
\caption{\textbf{Citation Processing Ablation}. We predict citations for the \textit{PWC Datasets} dataset using 6.7 billion capacity models. There are 4,735 datasets in this evaluation dataset.}
\label{table:citation-results-datasets}
\end{table}

\subsection{120B Validation Loss Per Source}

\begin{figure}[h!]
    \centering
    \includegraphics[width=\textwidth]{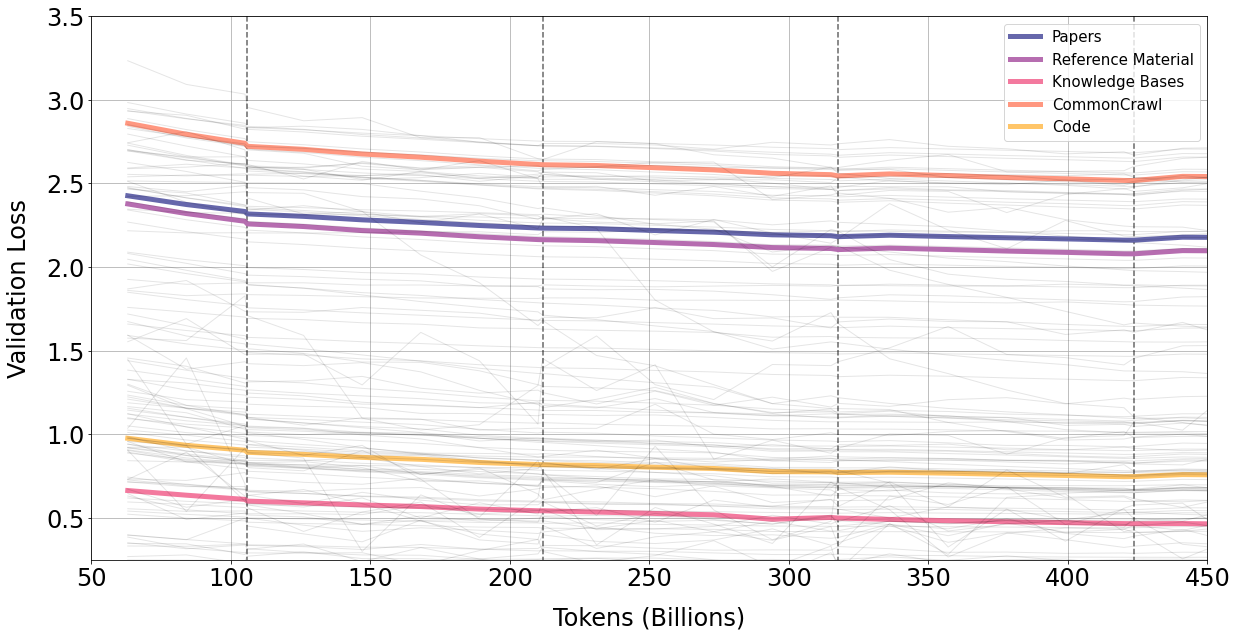}
    \caption{\textbf{Validation Loss Per Source}. Validation loss falls through training for all dataset categories. Results are shown for the 120B model above.}
    \label{fig:source_validation_loss}
\end{figure}

\subsection{Chain-of-Thought vs <work>}

We used the recent results by \citet{FLANPALM} of PaLM 540B on the MMLU validation set ~\citep{MMMLU} for comparison. While use of reasoning degrades performance versus direct prompting for both approaches, the \verb|<work>| token appears more robust.

\begin{table}[h!]
\begin{center}
\begin{tabular}{ lrrrr } 
\toprule
  \multicolumn{5}{c}{Chain-of-Thought versus \texttt{<work>}}      \\ 
\midrule
  Subject & Examples & PaLM 540B \texttt{CoT} & GAL 30B \texttt{<work>} & GAL 120B \texttt{<work>} \\ 
 \midrule
Abstract Algebra & 11 & 9.1\% & 27.3\% & \textbf{27.3\%}  \\
Astronomy & 16 & 7.1\% & \textbf{43.8\%} & 25.0\% \\
College Chemistry & 8 & 12.5\% & 37.5\% & \textbf{37.5\%} \\
College Computer Science & 11 & 9.1\% & 45.5\% & \textbf{54.6\%} \\
College Mathematics & 11 & 0.0\% & \textbf{36.4\%} & 18.2\% \\
College Physics & 11 & 36.4\% & 36.4\% & \textbf{45.5\%} \\
Econometrics & 11 & 33.3\% & 33.3\% & 33.3\% \\
Electrical Engineering & 16 & 18.8\% & 37.5\% & \textbf{56.3\%} \\
Elementary Mathematics & 41 & 24.4\% & 53.7\% & \textbf{58.5\%} \\
Formal Logic & 9 & 0.0\% & 21.4\% & \textbf{21.4\%} \\
High School Chemistry & 22 & 22.7\% & 27.3\% & \textbf{36.4\%} \\
High School Computer Science & 9 & 33.3\% & 44.4\% & \textbf{44.4\%} \\
High School Mathematics & 29 & 24.1\% & 31.0\% & \textbf{51.7\%} \\
High School Physics & 17 & 11.8\% & 23.5\% & \textbf{29.4\%} \\
High School Statistics & 23 & 26.1\% & 39.1\% & \textbf{56.5\%} \\
Machine Learning & 11 & 18.2\% & 9.1\% & \textbf{27.3\%} \\
\midrule
Overall & 261 & 19.1\% & 35.9\% & \textbf{42.4\%} \\
 \bottomrule
\end{tabular}
\end{center}
\caption{\textbf{<work> vs Chain-of-Thought}. PaLM is evaluated with CoT 5-shot. Galactica with the \texttt{<work>} token included in pre-training.  Results here are on MMLU dev for comparability with PaLM.}
\label{table:cotvswork}
\end{table}

\begin{table}[h!]
\small
\begin{center}
\begin{tabular}{ lrrrrr } 
\toprule
  \multicolumn{6}{c}{BIG-bench}      \\ 
\midrule
  Benchmark & OPT 30B & OPT 175B & BLOOM 176B & GAL 30B & GAL 120B \\ 
\midrule
Anachronisms & 47.4\% & 49.1\% & 1.3\% & 47.0\% & 48.7\% \\
Analogical Similarity & 12.7\% & 19.8\% & 19.2\% & 17.0\% & 23.5\% \\
Analytic Entailment & 40.0\% & 52.9\% & 48.6\% & 47.1\% & 51.3\% \\
Causal Judgment & 53.7\% & 55.3\% & 54.7\% & 49.5\% & 51.1\% \\
Crash Blossom & 42.1\% & 36.8\% & 47.4\% & 42.1\% & 42.1\% \\
Crass AI & 20.5\% & 34.1\% & 31.8\% & 40.9\% & 52.3\% \\
Dark Humor Detection & 46.3\% & 48.8\% & 51.3\% & 48.8\% & 46.3\% \\
Date Understanding & 15.5\% & 21.1\% & 12.2\% & 11.4\% & 16.8\% \\
Disambiguation QA & 39.5\% & 44.6\% & 44.2\% & 46.9\% & 43.0\% \\
Empirical Judgments & 38.4\% & 52.5\% & 56.6\% & 50.5\% & 54.6\% \\
English Proverbs & 26.5\% & 20.6\% & 26.5\% & 26.5\% & 17.7\% \\
Entailed Polarity & 87.8\% & 88.5\% & 89.2\% & 89.2\% & 85.8\% \\
Epistemic Reasoning & 43.4\% & 43.5\% & 61.2\% & 40.1\% & 53.0\% \\
Evaluating Information Essentiality & 32.4\% & 19.1\% & 29.4\% & 25.0\% & 22.1\% \\
Fantasy Reasoning & 67.7\% & 69.2\% & 65.2\% & 66.7\% & 52.7\% \\
Figure of Speech Detection & 10.2\% & 13.6\% & 22.0\% & 13.6\% & 15.3\% \\
General Knowledge & 51.4\% & 78.6\% & 80.0\% & 68.6\% & 74.3\% \\
GRE Reading Comprehension & 6.5\% & 12.9\% & 22.6\% & 16.1\% & 35.5\% \\
Hindu Knowledge & 32.6\% & 42.3\% & 48.6\% & 36.6\% & 49.7\% \\
Human Organs Senses & 45.2\% & 57.1\% & 59.5\% & 71.4\% & 73.8\% \\
Identify Odd Metaphor & 27.7\% & 21.3\% & 19.2\% & 19.2\% & 27.7\% \\
Implicatures & 44.3\% & 49.6\% & 53.7\% & 59.4\% & 69.9\% \\
Implicit Relations & 22.4\% & 35.3\% & 28.2\% & 16.5\% & 25.9\% \\
Intent Recognition & 66.2\% & 79.2\% & 89.5\% & 87.8\% & 89.5\% \\
Irony Identification & 50.5\% & 49.5\% & 63.6\% & 60.6\% & 59.6\% \\
Known Unknowns & 50.0\% & 52.2\% & 50.0\% & 50.0\% & 41.3\% \\
Logic Grid Puzzle & 32.7\% & 31.6\% & 31.1\% & 35.8\% & 39.4\% \\
Logical Args & 18.8\% & 34.4\% & 25.0\% & 34.4\% & 43.8\% \\
Logical Fallacy Detection & 50.9\% & 54.9\% & 54.5\% & 54.1\% & 55.1\% \\
Logical Sequence & 38.5\% & 46.2\% & 30.8\% & 25.6\% & 43.6\% \\
Mathematical Induction & 60.9\% & 55.1\% & 52.2\% & 44.9\% & 58.0\% \\
Metaphor Boolean & 51.1\% & 57.5\% & 61.5\% & 63.4\% & 49.1\% \\
Misconceptions & 56.1\% & 57.5\% & 54.8\% & 51.6\% & 58.0\% \\
Moral Permissibility & 50.6\% & 54.4\% & 57.0\% & 52.3\% & 49.7\% \\
Movie Recommendation & 6.4\% & 52.6\% & 49.4\% & 31.6\% & 36.8\% \\
Navigate & 49.3\% & 49.8\% & 51.1\% & 50.9\% & 51.8\% \\
Nonsense Words Grammar & 28.0\% & 46.0\% & 48.0\% & 38.0\% & 48.0\% \\
Novel Concepts & 9.4\% & 12.5\% & 15.6\% & 6.3\% & 9.4\% \\
Odd One Out & 30.2\% & 26.7\% & 22.1\% & 12.8\% & 19.8\% \\
Penguins in a Table & 29.5\% & 32.9\% & 28.2\% & 40.9\% & 36.9\% \\
Phrase Relatedness & 45.0\% & 51.0\% & 55.0\% & 53.0\% & 64.0\% \\
Physical Intuition & 39.5\% & 42.0\% & 37.0\% & 55.6\% & 58.0\% \\
Physics & 39.3\% & 42.8\% & 54.2\% & 55.9\% & 65.5\% \\
Presuppositions as NLI & 36.6\% & 36.2\% & 39.6\% & 34.0\% & 28.0\% \\
Question Selection & 39.8\% & 42.1\% & 5.2\% & 41.1\% & 42.7\% \\
Reasoning about Colored Objects & 33.9\% & 38.7\% & 40.5\% & 45.8\% & 55.0\% \\
Riddle Sense & 40.8\% & 57.1\% & 44.9\% & 46.9\% & 42.9\% \\
Ruin Names & 19.4\% & 20.8\% & 12.5\% & 24.1\% & 33.0\% \\
Sentence Ambiguity & 63.3\% & 60.0\% & 65.0\% & 60.0\% & 66.7\% \\
Similarities Abstraction & 21.1\% & 22.4\% & 27.6\% & 21.1\% & 13.2\% \\
Snarks & 42.0\% & 41.4\% & 47.0\% & 48.1\% & 48.6\% \\
Sports Understanding & 50.0\% & 48.8\% & 54.5\% & 52.0\% & 51.8\% \\
StrategyQA & 56.1\% & 58.5\% & 57.1\% & 53.9\% & 53.7\% \\
Temporal Sequences & 31.4\% & 28.4\% & 20.5\% & 26.4\% & 21.2\% \\
Timedial & 15.3\% & 22.2\% & 24.4\% & 39.9\% & 40.8\% \\
Understanding Fables & 20.1\% & 19.6\% & 24.9\% & 28.0\% & 20.1\% \\
Winowhy & 37.2\% & 39.7\% & 38.0\% & 56.5\% & 56.4\% \\
\midrule
Average (weighted) & 39.6\% & 43.4\% & 42.6\% & 46.6\% & 48.7\% \\
Average (unweighted) & 32.8\% & 42.7\% & 42.2\% & 42.7\% & 45.3\% \\

\bottomrule
\end{tabular}
\end{center}
\caption{\textbf{BIG-bench Results}. Galactica exceeds the performance of general models, even at lower scales.}
\label{table:bigbench-scores}
\end{table}

\clearpage

\subsection{Prompt Pre-training Datasets}

We report the prompt datasets we included in pre-training below.

\begin{table}[h!]
\begin{center}
\begin{tabular}{ llrr } 
\toprule
  Data source & Split & Prompts & Tokens \\ 
\midrule
 MedMCQA~\citep{MedMCQA} & \textit{train} & 180,894 & 13,311,290  \\
 RACE~\citep{RACE} & \textit{train} & 29,502 & 12,160,390 \\
 Quoref~\citep{Quoref} & \textit{train} & 19,206 & 10,361,335 \\
 ROPES~\citep{ROPES} & \textit{train}  & 10,815 & 2,672,195  \\
 BioASQ7 task b~\citep{BioASQ} & \textit{train} & 2,676 & 1,288,462  \\
 TQA~\citep{TQA} & \textit{train} & 8,566 & 1,856,473 \\
 BoolQ~\citep{BoolQ} & \textit{train} & 9,333 & 1,224,335  \\
 SciQ~\citep{SciQ} & \textit{train} & 10,346 & 1,397,668  \\
 QASC~\citep{QASC} & \textit{train} & 8,053 & 930,414  \\
 CommonSenseQA~\citep{CommonSenseQA} & \textit{train} & 9,644 & 660,750  \\
 OpenBookQA~\citep{OpenBookQA} & \textit{train} & 4,908 & 324,995 \\
 QCScience~\citep{QCScience} & \textit{train}  & 2,417 & 209,803 \\
 PubMedQA~\citep{PubMedQA} & \textit{train} & 495 & 186,304 \\
 QASPER~\citep{QASPER} & \textit{train} & 606 & 105,985  \\
 UChallenge (new) & \textit{train} & 346 & 29,308  \\
 TrueOrFalse (new) & \textit{train} & 107 & 2,854  \\
\bottomrule
\end{tabular}
\end{center}
\caption{\textbf{Question answering prompts} used in \texttt{Naturebook}}
\label{table:prompt-components-qa}
\end{table}

\begin{table}[h!]
\begin{center}
\begin{tabular}{ llrr } 
\toprule
  Data source & Split & Prompts & Tokens \\ 
\midrule
 JNLPBA~\citep{JNLPBA} & \textit{train} & 91,213 & 5,262,723 \\
 BC4CHEMD~\citep{BC4CHEMD} & \textit{train} & 30,234 & 1,756,929  \\
 ChemProt~\citep{ChemProt} & \textit{train} & 3,030 & 1,286,816  \\
 BC2GM~\citep{BC2GM} & \textit{train} & 12,375 & 704,357  \\
 S800~\citep{S800} & \textit{train} & 5,318 & 281,448  \\
 BC5CDR Chem~\citep{BC5CDR} & \textit{train} & 4,503 & 241,729  \\
 BC5CDR Disease~\citep{BC5CDR} & \textit{train} & 4,498 & 231,322  \\
 MethodNet (new) & \textit{train} & 659 & 167,904  \\
 Scientific Entities (new) & \textit{train} & 305 & 97,935 \\
\bottomrule
\end{tabular}
\end{center}
\caption{\textbf{Entity extraction prompts} used in \texttt{Naturebook}}
\label{table:prompt-components-ee}
\end{table}

\begin{table}[h!]
\begin{center}
\begin{tabular}{ llrr } 
\toprule
  Data source & Split & Prompts & Tokens \\ 
\midrule
PWC Desc (new) & \textit{train} & 3,586 & 9,663,419  \\
SciTail~\citep{SciTail} & \textit{train} & 23,361 & 1,383,614 \\
Fragmented Glass (new) & \textit{train} & 718 & 867,985 \\
SciTLDR~\citep{SciTLDR} & \textit{train} & 1,973 & 472,169  \\
\bottomrule
\end{tabular}
\end{center}
\caption{\textbf{Summarization prompts} used in \texttt{Naturebook}}
\label{table:prompt-components-sum}
\end{table}

\begin{table}[h!]
\begin{center}
\begin{tabular}{ llrr } 
\toprule
  Data source & Split & Prompts & Tokens \\ 
\midrule
Wizard of Wikipedia~\citep{WizardofWikipedia} & \textit{train} & 18,246 & 4,466,113  \\ 
Advising~\citep{Advising} & \textit{train} & 495 & 147,793 \\ 
\bottomrule
\end{tabular}
\end{center}
\caption{\textbf{Dialog prompts} used in \texttt{Naturebook}}
\label{table:prompt-components-dialog}
\end{table}

\begin{table}[h!]
\begin{center}
\begin{tabular}{ llrr } 
\toprule
  Data source & Split & Prompts & Tokens \\ 
\midrule
BACE Classification & \textit{train} & 1,198 & 122,699 \\ 
BACE Regression & \textit{train} & 1,198 & 154,656 \\ 
BBBP & \textit{train} & 1,613 & 115,916 \\ 
ClinTox & \textit{train} & 1,171 & 100,955 \\ 
Delaney & \textit{train} & 893 & 62,083 \\ 
FreeSolv & \textit{train} & 508 & 29,542 \\ 
HIV & \textit{train} & 32,572 & 2,308,966 \\ 
HOPV & \textit{train} & 2,217 & 333,620 \\ 
Lipo & \textit{train} & 3,327 & 362,342 \\ 
PCBA & \textit{train} & 714,277 & 553,645,656 \\ 
QM7 & \textit{train} & 5,416 & 320,199 \\ 
QM8 & \textit{train} & 275,569 & 27,163,516 \\ 
QM9 & \textit{train} & 1,259,090 & 128,427,073 \\ 
SAMPL & \textit{train} & 508 & 1,259,090 \\ 
SIDER & \textit{train} & 30,499 & 2,741,904 \\ 
Thermosol & \textit{train} & 1,396 & 139,481 \\ 
Tox21 & \textit{train} & 73,883 & 54,224,093 \\ 

\bottomrule
\end{tabular}
\end{center}
\caption{\textbf{Chemical property prediction prompts} used in \texttt{Naturebook}}
\label{table:prompt-components-chem}
\end{table}

\clearpage

\subsubsection{Chemical Property Prediction}

We set up a prediction task for chemical and physical properties with our validation set of 17,052 compounds. We use the PubChem document structure to design a prompt. We show an example for XLogP in Figure~\ref{fig:example_chem_prompt}. 

\begin{figure}[h]
\begin{tcolorbox}[colback=galwhite,colframe=galpurple2]
\begin{small}
\textbf{Canonical SMILES} \\

\verb|[START_SMILES]CC(=O)OC1=CC=CC=C1C(=O)O[END_SMILES]| \\

\textbf{Computed Properties} \\

| \verb|Property Name| | \verb|Property Value| \\
| \verb|XLogP3-AA Log P| |

\end{small}
\end{tcolorbox}
\caption{\textbf{Chemical Property Prompt}.
We design a prompt based on the PubChem document format. Using this prompt style, we test the model's ability to learn chemical and physical properties from the SMILES sequence.
}
\label{fig:example_chem_prompt}
\end{figure}

We report results in Table~\ref{table:chemical-pred}. The error decreases fairly smoothly with scale, suggesting self-supervised learning is occurring within-document from SMILES towards the chemical and physical properties. But it tails off for 120B which suggests more molecule data might be needed.

\begin{table}[h!]
\begin{center}
\begin{tabular}{ lrrrrr } 
\toprule
  \multicolumn{6}{c}{Chemical and Physical Property Prediction}      \\ 
\midrule
  Model & Param (bn) & Mol. Weight & XLogP & Rotatable Bond \# & Topological PSA   \\ 
\midrule
 GAL 125M & 0.1 & 101.43 & 1.638 & 4.389 & 36.63 \\
 GAL 1.3B & 1.3 & 101.05 & 1.413 & 3.930 & 41.11 \\
 GAL 6.7B & 6.7 & 81.76 & 1.197 & 2.932 & 30.01 \\
 GAL 30B & 30 & 77.46 & 1.101 & 3.534 & 29.54 \\
 GAL 120B & 120 & 86.57 & 1.131 & 3.474 & 28.84 \\
\bottomrule
\end{tabular}
\end{center}
\caption{\textbf{Chemical and physical property prediction}. All results reported as RMSE. Prediction error generally decreases with scale, indicating Galactica can infer properties from SMILES.}
\label{table:chemical-pred}
\end{table}

\subsubsection{Docking Regression}

We looked briefly at the docking score regression task~\citep{DockSTRING}. Here the task is to predict a docking score based on an target and a ligand. In the case of Galactica, we use a text format to represent this information. An example is shown in  Figure~\ref{fig:dockstring_example}. We report results in Table ~\ref{table:dockstring-regression}.

\begin{figure}[h]
\begin{tcolorbox}[colback=galwhite,colframe=galpurple2]
\begin{small}
\verb|[START_AMINO]MLEICLKLVGCKSKKGLSSSSSCYLEEALQRPVASDFEPQGLSEAARWNSKE...[END_AMINO]| \\

\verb|[START_I_SMILES]O1[C@@H]([C@@H](O)[C@@H](O)[C@@H]1N2C(=O)NC(=O)C=C2)...[END_I_SMILES]| \\

\textbf{Question:} What will be the docking score of this compound against the protein? \\

\textbf{Answer:} -8.8
\end{small}
\end{tcolorbox}
\caption{\textbf{DockSTRING Format}. To construct the training set, we take the protein target and ligand sequences, pose a natural language question, and have the docking score as the answer.}
\label{fig:dockstring_example}
\end{figure}

\begin{table}[h!]
\begin{center}
\begin{tabular}{ lrrrrrr } 
\toprule
  \multicolumn{7}{c}{Docking Regression}      \\ 
\midrule
  Model & Param (bn) & ESR2 & F2 & KIT & PARP1 & PGR \\ 
\midrule
 GAL 125M & 0.1 & -12.4 & -6.09 & -6.73 & -1.69 & -12.4 \\
 GAL 1.3B & 1.3 & -0.293 & 0.591 & 0.063 & 0.728 & -1.72 \\
 GAL 6.7B & 6.7 & -0.216 & 0.694 & 0.290 & 0.681 & -0.894 \\
 GAL 30B & 30 & -0.186 & 0.679 & 0.313 & 0.732 & -0.468 \\
 GAL 120B & 120 & -0.564 & 0.626 & 0.249 & 0.732 & -0.960 \\
\bottomrule
\end{tabular}
\end{center}
\caption{\textbf{DockSTRING Results}. Metric shown is $R^{2}$.}
\label{table:dockstring-regression}
\end{table}

For three of the targets, Galactica is able to infer from looking at the sequences alone, and performance scales from 1.3B parameters onwards. However, Galactica does not solve the two harder targets ESR2 and PGR. This hints at a limitation with the text representation, and may point to more geometrical information being needed to solve the task with reasonable data-efficiency.

\subsubsection{Rest of MMLU}

We report social sciences and results for other fields below:

\begin{table}[h!]
\begin{center}
\begin{tabular}{ lcccccc } 
\toprule
  Subject & OPT & BLOOM & Gopher & Chinchilla & GAL 30B & GAL 120B \\
\midrule
 Anatomy & 28.9\% & 37.0\% & 56.3\% & 70.4\% & 54.1\% & 58.5\% \\
 Business Ethics & 31.0\% & 36.0\% & 70.0\% & 72.0\% & 42.0\% & 48.0\% \\
 Clinical Knowledge & 21.9\% & 29.8\% & 67.2\% & 75.1\% & 57.7\% & 59.2\% \\
 Computer Security & 32.0\% & 34.0\% & 65.0\% & 76.0\% & 65.0\% & 67.0\% \\
 Conceptual Physics & 34.9\% & 36.6\% & 49.4\% & 67.2\% & 43.4\% & 50.6\% \\
 Global Facts & 23.0\% & 32.0\% & 38.0\% & 39.0\% & 32.0\% & 35.0\% \\
 High School European History & 6.7\% & 4.8\% & 72.1\% & 78.8\% & 60.6\% & 67.3\% \\
 High School Geography & 26.3\% & 38.9\% & 76.8\% & 86.4\% & 58.1\% & 63.6\% \\
 High School Gov. \& Politics & 32.6\% & 30.6\% & 83.9\% & 91.2\% & 58.5\% & 61.7\% \\
High School Macroeconomics & 36.2\% & 23.1\% & 65.1\% & 70.5\% & 40.5\% & 46.4\% \\
High School Microeconomics & 32.8\% & 27.3\% & 66.4\% & 77.7\% & 49.2\% & 55.9\% \\
High School Psychology & 25.5\% & 36.9\% & 81.8\% & 86.6\% & 68.8\% & 74.3\% \\
High School US History & 9.3\% & 11.8\% & 78.9\% & 83.3\% & 51.5\% & 58.3\% \\
High School World History & 30.0\% & 29.1\% & 75.1\% & 85.2\% & 63.7\% & 71.7\% \\
Human Aging & 35.0\% & 34.5\% & 66.4\% & 77.6\% & 55.2\% & 59.2\% \\
Human Sexuality & 26.0\% & 33.6\% & 67.2\% & 86.3\% & 56.5\% & 58.8\% \\
International Law & 33.1\% & 41.3\% & 77.7\% & 90.9\% & 64.4\% & 71.1\% \\
Jurisprudence & 0.0\% & 0.0\% & 71.3\% & 79.6\% & 47.2\% & 53.7\% \\
Logical Fallacies & 28.2\% & 28.2\% & 72.4\% & 80.4\% & 47.2\% & 59.5\% \\
Management & 25.2\% & 27.2\% & 77.7\% & 82.5\% & 60.2\% & 63.1\% \\
Marketing & 32.5\% & 41.0\% & 83.3\% & 89.7\% & 70.5\% & 76.5\% \\
Miscellaneous & 31.5\% & 37.7\% & 75.7\% & 84.5\% & 54.0\% & 63.9\% \\
Moral Disputes & 28.2\% & 32.7\% & 66.8\% & 77.5\% & 50.3\% & 56.6\% \\
Moral Scenarios & 25.4\% & 24.4\% & 40.2\% & 36.5\% & 24.1\% & 24.2\% \\
Nutrition & 30.4\% & 32.4\% & 69.9\% & 77.1\% & 63.1\% & 67.3\% \\
Philosophy & 29.9\% & 31.5\% & 68.8\% & 79.4\% & 52.4\% & 54.7\% \\
Prehistory & 36.7\% & 36.1\% & 67.6\% & 81.2\% & 52.2\% & 59.6\% \\
Professional Accounting & 29.8\% & 28.7\% & 44.3\% & 52.1\% & 31.2\% & 40.0\% \\
Professional Law & 30.3\% & 25.5\% & 44.5\% & 56.5\% & 34.6\% & 36.0\% \\
Professional Medicine & 27.9\% & 25.4\% & 64.0\% & 75.4\% & 52.2\% & 59.6\% \\
Professional Psychology & 32.7\% & 33.3\% & 68.1\% & 75.7\% & 50.5\% & 56.5\% \\
Public Relations & 34.5\% & 30.0\% & 71.8\% & 73.6\% & 44.5\% & 53.6\% \\
Security Studies & 35.1\% & 29.8\% & 64.9\% & 75.9\% & 46.5\% & 57.1\% \\
Sociology & 26.4\% & 29.9\% & 84.1\% & 91.0\% & 65.7\% & 72.6\% \\
US Foreign Policy & 44.0\% & 37.0\% & 81.0\% & 92.0\% & 64.0\% & 75.0\% \\
Virology & 30.7\% & 28.3\% & 47.0\% & 53.6\% & 44.6\% & 48.2\% \\
World Religion & 43.9\% & 41.5\% & 84.2\% & 87.7\% & 44.4\% & 64.9\% \\
\bottomrule
\end{tabular}
\end{center}
\caption{\textbf{Rest of MMLU}. The corpus delta effects are more evidence with non-STEM subjects in particular, where Galactica lags the performance of Chinchilla and Gopher.}
\label{table:other-mmlu}
\end{table}

\subsection{Further Training Dataset Details}

\subsubsection{FragmentedGlass}

We compile a list of scientific entities, retrieve fragments for each one, and write a description of the entity based on the retrieved fragments. This can be considered a summarization task. We also write ground-truth descriptions without any retrieved fragments.

\subsubsection{MethodNet}

We compile machine learning abstracts and predict the new method that was introduced in the paper.

\subsubsection{PWC Desc}

For a list of dataset and methods in machine learning, we retrieve fragments for each one from the introducing paper, and write a summary description based on the retrieved fragments.

\subsubsection{Ribosome}

We use Expasy\footnote{\url{https://web.expasy.org/translate/}} to create a paired translation set between nucleotide sequences from the protein coding part of the human genome and protein sequences.

\subsubsection{S2}

Papers from certain fields are ignored due to quality concerns: psychology, business, art, economics, geography, history, political science, philosophy and sociology. Papers from journals with words like "law", "history", "politics", "business", "religion" were also ignored. For S2, we also exclude papers from low impact journals. The approximate impact factor of each journal in the S2 dataset was computed, by counting the number of papers in that journal and the number of citations that these papers received. If the approximate impact factor $<1$, the papers from that journal are ignored. Non-English papers are ignored. Some of these constraints can likely be relaxed in future work.

\subsubsection{ScientificEntities}

For a random sample of academic paper abstracts, we predict the scientific entities that were mentioned in the abstract.

\subsubsection{StackExchange}

We include question and answers from the following sources: academic, ai, arduino, astronomy, aviation, bioinformatics, biology, chemistry, chess, cogsci, computergraphics, cs, cseducators, cstheory, datascience, dsp, earthscience, economics, electronics, engineering, hardwarerecs, health, hsm, math, matheducators, mathematica, mathoverflow, /mechanics, networkengineering, or, physics, puzzling, quant, quantumcomputing, retrocomputing, reverseengineering, robotics, scicomp, softwareengineering, softwarerecs, sound, space, stats.

\subsubsection{TrueOrFalse}

We include 107 True or False questions to improve zero-shot performance for this type of question.

\subsubsection{UChallenge}

We include 346 free-form question and answers of university-level questions about science; this is a form of closed-book QA (and not multiple-choice).

\clearpage

\subsection{Evaluation Dataset Examples}

\subsubsection{AminoProbe}

\begin{figure}[h!]
\begin{tcolorbox}[colback=galwhite,colframe=galpurple2]
\begin{small}
\textbf{Prompt} \\

\textbf{Question:} Does peptide bond cleavage occur on the carbonyl side or the amino side for trypsin? \\

\textbf{Answer}: carbonyl

\end{small}
\end{tcolorbox}
\label{fig:aminoprobe_ex}
\end{figure}

\subsubsection{Galaxy Clusters}

\begin{figure}[h!]
\begin{tcolorbox}[colback=galwhite,colframe=galpurple2]
\begin{small}
\textbf{Prompt} \\

Abell 370 is a galaxy cluster located in the constellation of \\

\textbf{Correct Completion}: Cetus

\end{small}
\end{tcolorbox}
\label{fig:cluster_ex}
\end{figure}

\subsubsection{Mineral Groups}

\begin{figure}[h!]
\begin{tcolorbox}[colback=galwhite,colframe=galpurple2]
\begin{small}
\textbf{Prompt} \\

Fayalite is a silicate mineral from the major group \\

\textbf{Correct Completion}: Nesosilicates

\end{small}
\end{tcolorbox}
\label{fig:mineral_ex}
\end{figure}

\subsubsection{Deduplication Results}

One of our concerns from reading the literature was the lack of data leakage analysis for results on MMLU, given the massive corpuses being used. Following from previous work of \citet{GPT3}, we search for n-gram matches between the training and test set. We chose to remove any 13-gram matches from the test set that appear in the training set and we report the scores before and after removal of these clashing examples. Results are shown overleaf.

\begin{tabular}{lrrrr}
\toprule
                & score\_before &  score\_after &  count\_before &  count\_after \\
\midrule
abstract\_algebra & 33.0\% & 32.32\% & 100 & 99 \\
anatomy & 58.52\% & 58.95\% & 135 & 134 \\
astronomy & 65.13\% & 64.67\% & 152 & 150 \\
business\_ethics & 48.0\% & 48.0\% & 100 & 100 \\
clinical\_knowledge & 59.24\% & 59.24\% & 265 & 265 \\
college\_biology & 68.75\% & 69.23\% & 144 & 143 \\
college\_chemistry & 46.0\% & 46.46\% & 100 & 99 \\
college\_computer\_science & 49.0\% & 48.98\% & 100 & 98 \\
college\_mathematics & 43.0\% & 45.26\% & 100 & 95 \\
college\_medicine & 57.23\% & 57.74\% & 173 & 168 \\
college\_physics & 42.16\% & 42.27\% & 102 & 97 \\
computer\_security & 67.0\% & 67.35\% & 100 & 98 \\
conceptual\_physics & 50.64\% & 50.85\% & 235 & 234 \\
econometrics & 42.11\% & 42.11\% & 114 & 114 \\
electrical\_engineering & 62.76\% & 62.76\% & 145 & 145 \\
elementary\_mathematics & 38.10\% & 38.10\% & 378 & 378 \\
formal\_logic & 32.54\% & 32.54\% & 126 & 126 \\
global\_facts & 35.0\% & 35.05\% & 100 & 97 \\
high\_school\_biology & 69.35\% & 69.61\% & 310 & 306 \\
high\_school\_chemistry & 47.78\% & 47.78\% & 203 & 203 \\
high\_school\_computer\_science & 70.0\% & 70.0\% & 100 & 100 \\
high\_school\_european\_history & 67.27\% & 66.17\% & 165 & 133 \\
high\_school\_geography & 63.63\% & 63.63\% & 198 & 198 \\
high\_school\_government\_and\_politics & 61.66\% & 61.46\% & 193 & 192 \\
high\_school\_macroeconomics & 46.41\% & 46.53\% & 390 & 389 \\
high\_school\_mathematics & 32.59\% & 32.58\% & 270 & 267 \\
high\_school\_microeconomics & 55.88\% & 55.88\% & 238 & 238 \\
high\_school\_physics & 33.77\% & 33.77\% & 151 & 151 \\
high\_school\_psychology & 74.31\% & 74.26\% & 545 & 544 \\
high\_school\_statistics & 41.20\% & 41.20\% &           216 &          216 \\
high\_school\_us\_history & 58.33\% & 58.59\% &           204 &           99 \\
high\_school\_world\_history & 71.73\% & 72.04\% &           237 &          186 \\
human\_aging & 59.19\% & 59.19\% &           223 &          223 \\
human\_sexuality & 58.78\% & 58.78\% &           131 &          131 \\
international\_law & 71.07\% & 71.07\% &           121 &          121 \\
jurisprudence & 53.70\% & 53.70\% &           108 &          108 \\
logical\_fallacies & 59.51\% & 59.26\% &           163 &          162 \\
machine\_learning & 38.39\% & 36.54\% &           112 &          104 \\
management & 63.11\% & 63.11\% &           103 &          103 \\
marketing & 76.50\% & 76.50\% &           234 &          234 \\
medical\_genetics & 68.0\% & 67.68\% &           100 &           99 \\
miscellaneous & 63.86\% & 63.81\% &           783 &          782 \\
moral\_disputes & 56.65\% & 56.52\% &           346 &          345 \\
moral\_scenarios & 24.24\% & 24.24\% &           895 &          895 \\
nutrition & 67.32\% & 67.32\% &           306 &          306 \\
philosophy & 54.66\% & 54.52\% &           311 &          310 \\
prehistory & 59.57\% & 59.57\% & 324 &          324 \\
professional\_accounting & 40.07\% & 39.86\% &           282 &          281 \\
professional\_law & 36.05\% & 36.05\% &          1534 &         1534 \\
professional\_medicine & 59.56\% &     59.63\% &           272 &          270 \\
professional\_psychology & 56.54\% &     56.54\% &           612 &          612 \\
public\_relations & 53.63\% &     53.63\% &           110 &          110 \\
security\_studies & 57.14\% & 57.14\% &           245 &          245 \\
sociology & 72.64\% &     73.0\% &           201 &          200 \\
us\_foreign\_policy & 75.0\% & 75.76\% &           100 &           99 \\
virology & 48.19\% & 48.48\% &           166 &          165 \\
world\_religions & 64.91\% & 64.91\% &           171 &          171 \\
  
\bottomrule
\end{tabular}

\subsubsection{Example Wikipedia Article Written by Galactica}

\begin{figure}[h]
\begin{tcolorbox}[colback=galwhite,colframe=galpurple2]
\begin{small}

\textbf{Multi-Head Attention} \\

Multi-Head Attention is an attention mechanism used in the Transformer architecture for neural networks (Vaswani et al., 2017). \\

\textbf{Background} \\

In natural language processing, an attention mechanism is a model that maps a query and a set of key-value pairs to an output, where the query, keys, values, and output are all vectors. The output is computed as a weighted sum of the values, where the weight assigned to each value is computed by a compatibility function of the query with the corresponding key (Vaswani et al., 2017). \\

The scaled dot-product attention is an attention mechanism used in the Transformer architecture. It is defined as mapping a query and a set of key-value pairs to an output, where the query, keys, values, and output are all vectors. The output is computed as a weighted sum of the values, where the weight assigned to each value is computed by a compatibility function of the query with the corresponding key (Vaswani et al., 2017). \\

The multi-head attention is an extension of the scaled dot-product attention. The multi-head attention mechanism is defined as several scaled dot-product attention stacked together, where the input is linearly projected h times with different, learned linear projections to $d_k$, $d_k$, and $d_v$ dimensions, respectively (Vaswani et al., 2017). \\

\textbf{Formal definition} \\

Let the query matrix be $Q$, the key matrix be $K$, and the value matrix be $V$. The matrices are of shape $d_k \times d_q$, $d_k \times d_k$, and $d_v \times d_k$ respectively. The multi-head attention mechanism is defined as:

\[ \text{MultiHead}\left(Q, K, V\right) = \text{Concat}\left(\text{head}_{1},\dots,\text{head}_{h}\right)W^{O} \]

where:

\[ \text{head}_{i} = \text{Attention}\left(QW_{i}^{Q},KW_{i}^{K}, VW_{i}^{V}\right) \]

where $W_{i}^{Q}$, $W_{i}^{K}$, and $W_{i}^{V}$ are the learned linear projections to the $d_k$, $d_k$, and $d_v$ dimensions, respectively. $W^{O}$ is the linear projection from $d_vh$ to $d_v$ dimensions (Vaswani et al., 2017). \\

\textbf{Applications} \\

Multi-head attention is widely used in natural language processing and other domains. For example, in the BERT model, multi-head attention is used to learn the context-aware representation of the input sequence (Devlin et al., 2019). \\

\textbf{References} \\

1. Vaswani, Ashish, Noam Shazeer, Niki Parmar, Jakob Uszkoreit, Llion Jones, Aidan N. Gomez, Lukasz Kaiser, and Illia Polosukhin. “Attention Is All You Need.” ArXiv abs/1706.03762 (2017). \\
2. Devlin, Jacob, Ming-Wei Chang, Kenton Lee, and Kristina Toutanova. “BERT: Pre-Training of Deep Bidirectional Transformers for Language Understanding.” ArXiv abs/1810.04805 (2019).
\end{small}
\end{tcolorbox}
\label{fig:example_data}
\end{figure}

\clearpage

\subsubsection{Example Literature Survey Written by Galactica}

\begin{figure}[h]
\begin{tcolorbox}[colback=galwhite,colframe=galpurple2]
\begin{small}

\textbf{Self-Supervised Learning, A Survey} \\

\textbf{Abstract:} In this paper we review self-supervised learning, a method of learning features from unlabelled data. We discuss the historical motivation behind self-supervised learning and review some of the current state of the art methods. We also look at how self-supervised learning has been used to solve a wide range of problems, including image classification, object detection, video analysis and robotics. Finally we discuss some of the challenges and future research directions for self-supervised learning. \\

\textbf{1 Introduction} \\

Deep learning has been very successful at solving many problems in machine learning, however these methods are reliant on large amounts of labelled data. In many real world applications, it is very expensive or impossible to obtain large amounts of labelled data. Self-supervised learning has emerged as a way of overcoming this issue. \\

Self-supervised learning is the process of learning a task that can be automatically generated from unlabelled data. In this way, the data is automatically labelled by exploiting some form of structure in the data. Once the self-supervised task has been solved, the features learned by the network are used to solve the original task. \\

In this paper we will discuss the historical motivation behind self-supervised learning and review some of the current state of the art methods. We will look at how self-supervised learning has been used to solve a wide range of problems, including image classification, object detection, video analysis and robotics. Finally we will discuss some of the challenges and future research directions for self-supervised learning. \\

\textbf{2 Historical Motivation} \\

The first wave of deep learning was dominated by supervised learning methods. These methods require large amounts of labelled data, which is very expensive to obtain in many domains. To overcome this issue, unsupervised learning methods were introduced. These methods don’t require labelled data, and so can be applied to a wide range of domains. The downside of these methods is that the learned features are not as good as those learned by supervised learning methods. \\

To bridge the gap between supervised and unsupervised learning, self-supervised learning was introduced. The idea is to learn a supervised task from unlabelled data. Once this task has been learned, the features learned can be transferred to other tasks. The hope is features learned by self-supervised learning will be better than those learned by unsupervised learning, and that self-supervised learning can be applied to a wide range of domains. \\

The first method of self-supervised learning was autoencoders (Hinton et al., 1993). Autoencoders are a type of neural network that learn to reconstruct the input. This can be thought of as learning a self-supervised task of predicting the input from the input. The features learned by autoencoders have been shown to be useful for solving other tasks (Hinton et al., 2006). \\

Another method of self-supervised learning is multi-task learning (Caruana, 2017). In multi-task learning, multiple tasks are learned simultaneously. The idea is that by learning multiple tasks, the network will learn better features. In this way, multi-task learning can be thought of as a form of self-supervised learning. \\

\textbf{3 Learning Methods} \\
\textit{(cont)} \\

\textbf{3.1 Pretext Tasks} \\

The first method of self-supervised learning that we will discuss is the use of pretext tasks. In this method, a supervised task is learned from unlabelled data. This task is called a pretext task, and the labels for this task are automatically generated from the data. Once the pretext task has been learned, the features learned by the network are used to solve the original task. \textit{(cont)}
\end{small}
\end{tcolorbox}
\label{fig:example_data}
\end{figure}

\clearpage

\subsubsection{Example Lecture Notes Written by Galactica}

\begin{figure}[h]
\begin{tcolorbox}[colback=galwhite,colframe=galpurple2]
\begin{small}

\textbf{Lecture 1: Hartree-Fock Theory} \\

\textbf{Introduction} \\

In this lecture, we will consider the simplest quantum mechanical method for approximating the ground state energy of a many-electron system. The Hartree-Fock method is also known as the self-consistent field method, and is an approximation to the exact many-electron wavefunction known as the Slater determinant. The Slater determinant is an anti-symmetric product of one-electron functions (orbitals), and is the simplest wavefunction that satisfies the Pauli exclusion principle. The Hartree-Fock method is an iterative method for finding the best possible set of orbitals that minimizes the energy of the Slater determinant. \\

In the Hartree-Fock method, the electron-electron repulsion energy is approximated as a classical Coulomb interaction between two charge distributions, one for each electron. This approximation neglects the correlation between the electrons, which is an important part of the total electron-electron repulsion energy. The resulting method is a mean-field theory in which each electron moves in an average field due to the other electrons. The resulting equations are known as the Hartree-Fock equations, and can be solved self-consistently to obtain the best possible orbitals. The resulting ground state energy is known as the Hartree-Fock energy, and the Hartree-Fock wavefunction is a single Slater determinant built from these orbitals. \\

The Hartree-Fock method is also known as the Self-Consistent Field (SCF) method, because the equations for the orbitals are solved self-consistently. The Hartree-Fock equations can be derived variationally by minimizing the energy of the Slater determinant with respect to the orbitals. The resulting orbitals are known as the canonical Hartree-Fock orbitals, and are not necessarily localized in space. The Hartree-Fock energy is invariant to unitary transformations of the canonical orbitals, and therefore there are an infinite number of orbitals that yield the same Hartree-Fock energy. These orbitals are known as non-canonical orbitals, and can be localized in space by appropriate unitary transformations. \\

\textbf{Single-Electron Approximation} \\

In this section, we will review the basics of quantum mechanics for a single particle. This is useful for understanding the single-electron approximation used in Hartree-Fock theory. \\

The time-independent Schrödinger equation for a particle in a potential $V(r)$ is given by: \\

\[ \bar{H}\psi(r) = E\psi(r) \] \\

where the Hamiltonian is \\

\[ \bar{H} = -\frac{\hbar}{2m}\nabla^{2} + V(r) \] \\

The time-independent Schrödinger equation is an eigenvalue equation for the Hamiltonian operator, where the eigenvalues are the allowed energies of the system. The Hamiltonian is a sum of two operators, one corresponding to the kinetic energy of the particle, and the other corresponding to the potential energy. The potential energy operator acts on the wavefunction by multiplying by the potential $V(r)$. The kinetic energy operator is the Laplacian operator $\nabla^{2}$, which is the divergence of the gradient of the wavefunction. The Laplacian operator is a second derivative with respect to the position of the particle. \\

(cont)
\end{small}
\end{tcolorbox}
\label{fig:example_data}
\end{figure}

\clearpage

\subsubsection{I'm sorry Frank, I think you missed it}

If AI is going to help us explore the universe, we need it to have basic chess abilities to alleviate boredom - given the impossibility of faster-than-light travel.

The BIG-bench task suite of \cite{BIGBenchakasomanyauthorsitdoesntfitinthecontextwindow} has a benchmark for checkmate-in-one detection. For fun, we made a dataset of 20,000 public chess games and converted them to ASCII chess using the python-chess library\footnote{\url{https://python-chess.readthedocs.io/en/latest/}}. We included 19,426 games in our pre-training corpus (rest for validation). We also recorded the ELO ratings of players. An example document looks like below:

\begin{figure}[h!]
\begin{tcolorbox}[colback=galwhite,colframe=galpurple2]
\begin{small}

\# A Chess Game \\

\#\# Player Information \\

White ELO: 2286 \\
Black ELO: 2586 \\

\#\# The Game Begins \\

r n b q k b n r \\
p p p p p p p p \\
. . . . . . . . \\
. . . . . . . . \\
. . . . . . . . \\
. . . . . . . . \\
P P P P P P P P \\
R N B Q K B N R \\

White (ELO: 2286) plays e4 \\

r n b q k b n r \\ 
p p p p p p p p \\ 
. . . . . . . . \\
. . . . . . . . \\
. . . . P . . . \\
. . . . . . . . \\ 
P P P P . P P P \\
R N B Q K B N R \\

(cont)

\end{small}
\end{tcolorbox}
\label{fig:mineral_ex}
\end{figure}

For evaluation, we converted the checkmate-in-one boards to ASCII and prompted for a move. Results are shown below.

\begin{table}[h]
\begin{center}
\begin{tabular}{ lr } 
\toprule
  Model  & Accuracy \\
\midrule
 GAL 125M & 0.54\% \\
 GAL 1.3B & 0.43\% \\
 GAL 6.7B & 1.77\% \\
 GAL 30B & 1.29\% \\
 GAL 120B & 3.03\% \\
\bottomrule
\end{tabular}
\end{center}
\caption{\textbf{Checkmate-in-one Results}. Metric shown is Accuracy.}
\label{table:checkmate-in-one}
\end{table}

While this represents the state-of-the-art over other large language models\footnote{\url{https://github.com/google/BIG-bench/tree/main/bigbench/benchmark_tasks/checkmate_in_one}}, it is clear that more work is needed on this problem.

\end{document}